\documentclass[conference,a4paper]{IEEEtran}

\usepackage{amsfonts}
\usepackage{cite}
\usepackage{graphicx}
\usepackage{amssymb}
\usepackage{subfigure}
\usepackage{array}
\usepackage{color}
\usepackage{amsmath}
\usepackage{algorithmic}
\usepackage{algorithm}
\usepackage{cases}
\usepackage{mathrsfs}
\usepackage{epsfig}
\usepackage{psfrag}
 \usepackage{multirow}
\usepackage{hyperref}
\usepackage[hyphenbreaks]{breakurl}
\usepackage{setspace}
\usepackage{stfloats}
\usepackage{enumerate}
\usepackage{enumitem}
\usepackage[dvipsnames, svgnames, x11names]{xcolor} % 丰富的颜色选择
\allowdisplaybreaks[4]

\usepackage{etoolbox}
\makeatletter
\patchcmd{\@makecaption}
  {\scshape}
  {}
  {}
  {}
\makeatletter
\patchcmd{\@makecaption}
  {\\}
  {.\ }
  {}
  {}
\makeatother

\newtheorem{Thm}{Theorem}
\newtheorem{Lem}{Lemma}

\newtheorem{Prob}{Problem}
\newtheorem{Rem}{Remark}

\newtheorem{Asump}{Assumption}

\IEEEoverridecommandlockouts

\begin{document}

\title{Sample-based and Feature-based Federated Learning for Unconstrained and Constrained Nonconvex Optimization via Mini-batch SSCA}
\author{Ying Cui, Yangchen Li, and Chencheng Ye\thanks{The authors are with School of Electronic Information and Electrical Engineering, Shanghai Jiao Tong University, China. This paper was presented in part at IEEE ICC 2021~\cite{2021ICC}.}
}

\maketitle

\begin{abstract}
	Federated learning (FL) has become a hot research area in enabling the collaborative training of machine learning models among multiple clients that hold sensitive local data. Nevertheless, unconstrained federated optimization has been studied mainly using stochastic gradient descent (SGD), which may converge slowly, and constrained federated optimization, which is more challenging, has not been investigated so far. This paper investigates sample-based and feature-based federated optimization, respectively, and considers both unconstrained and constrained nonconvex problems for each of them. First, we propose FL algorithms using stochastic successive convex approximation (SSCA) and mini-batch techniques. These algorithms can adequately exploit the structures of the objective and constraint functions and incrementally utilize samples. We show that the proposed FL algorithms converge to stationary points and Karush-Kuhn-Tucker (KKT) points of the respective unconstrained and constrained nonconvex problems, respectively. Next, we provide algorithm examples with appealing computational complexity and communication load per communication round. We show that the proposed algorithm examples for unconstrained federated optimization are identical to FL algorithms via momentum SGD and provide an analytical connection between SSCA and momentum SGD. Finally, numerical experiments demonstrate the inherent advantages of the proposed algorithms in convergence speeds, communication and computation costs, and model specifications.
\end{abstract}
\begin{IEEEkeywords}
	Federated learning, nonconvex optimization, stochastic optimization, stochastic successive convex approximation.
\end{IEEEkeywords}

%\newpage

\section{Introduction}
\label{sec:intro}
Machine learning with distributed databases has been a hot research area~\cite{li2014communication}.
The amount of data at each client can be large, and hence the data uploading to a central server may be constrained by energy and bandwidth limitations.
Besides, local data may contain highly sensitive information, e.g., travel records, health information, and web browsing history, and thus a client may be unwilling to share it.
%Therefore, it is impossible or undesirable to upload distributed databases to a central server.
Recent years have witnessed the growing interest in federated learning (FL), where data is maintained locally during the collaborative training of the server and clients~\cite{li2019federated,li2020federated}.
FL can protect data privacy for privacy-sensitive applications and improve communication efficiency.

Model aggregation, cryptographic methods, and differential privacy are three main privacy mechanisms in FL. They provide different privacy guarantees. Specifically, model aggregation{, including model averaging and gradient averaging, is a basic privacy mechanism that reduces privacy risk by sharing model-related intermediate results computed based on local {data}}~\cite{mcmahan2017communication,yu2019parallel,9003425,xie2020asynchronous,GenQSGD,HE,song2013stochastic,yang2019federated,hardy2017private,yang2019parallel,chen2020vafl}.
{Note that} communicating {locally computed results}  generally reveals much less information than communicating local data.
%Cryptographic methods and differential privacy can further enhance the privacy of FL.
Cryptographic methods, such as {homomorphic encryption~\cite{HE,hardy2017private} and secret sharing~\cite{2017SecureML}}, {further enhance privacy protection} by encrypting {locally computed results} before sharing, at the cost of communication and computation efficiency reduction.
%As an extensive computation and communication overhead is usually induced, the resulting FL systems are inefficient.
Finally, differential privacy~\cite{song2013stochastic,chen2020vafl} {enhances privacy protection} by adding random noise to {locally computed results} at the cost of model performance decline.

Depending on whether data is distributed over the sample space or feature space, FL can be classified into sample-based (horizontal) FL and feature-based (vertical) FL. Specifically, in sample-based FL~\cite{mcmahan2017communication,yu2019parallel,9003425,xie2020asynchronous,GenQSGD,HE,song2013stochastic},  the datasets of different clients have the same feature space but no (or little) intersection on the sample space. On the contrary, in feature-based FL~\cite{yang2019federated,hardy2017private,yang2019parallel,chen2020vafl}, the datasets of different clients share the same sample space but differ in the feature space. {As a client cannot evaluate the impact of the model on the loss for a particular sample relying purely on its local data, feature-based FL is more challenging and hence less studied}.

Existing works on FL~\cite{mcmahan2017communication,yu2019parallel,9003425,xie2020asynchronous,GenQSGD,HE,song2013stochastic,yang2019federated,hardy2017private,yang2019parallel,chen2020vafl} investigate only unconstrained optimization problems mainly using mini-batch stochastic gradient descent (SGD). In sample-based FL  {via mini-batch} SGD~\cite{mcmahan2017communication,yu2019parallel,9003425,xie2020asynchronous,GenQSGD,HE,song2013stochastic},  the global model is iteratively updated at the server by aggregating and averaging the clients' locally computed models {or model-related results}. Specifically, at one communication round, each client downloads the latest global model parameters and conducts one {(e.g., in FedSGD~\cite{mcmahan2017communication})} or multiple {(e.g., in FedAvg\footnote{{In FedAvg,  all  local samples are utilized during local updates in each communication round.}}~\cite{mcmahan2017communication} and PR-SGD\cite{yu2019parallel})}  local SGD updates to refine its local model. Multiple local SGD updates can reduce {the communication cost (required number of  {communication rounds}) with possibly increased computation cost. To further reduce communication cost, some recent works carefully design   SGD update directions (e.g., momentum term\cite{9003425}) or the numbers of local SGD updates at all clients~\cite{xie2020asynchronous,GenQSGD}.}

%Besides, some other works aim at improving the communication efficiency by selecting only some clients to participate in each model averaging step~\cite{FedCS} or modifying local SGD updates~\cite{svrgd,yu2019parallel}.

{In contrast}, the existing feature-based FL algorithms {via mini-batch SGD}~\cite{yang2019federated,hardy2017private,yang2019parallel,chen2020vafl}  {conduct only one SGD update   in each communication round} and impose additional restrictions on the structure of {the loss function  to guarantee privacy risk reduction. Specifically,  the feature-based FL algorithms in \cite{yang2019federated,hardy2017private,yang2019parallel} are designed only for two clients and some particular loss functions. In contrast, the feature-based FL algorithm in \cite{chen2020vafl} applies to  an arbitrary number of clients and a more general {loss} function. Besides, the feature-based FL algorithms in \cite{yang2019federated,yang2019parallel,chen2020vafl} do not maintain the global model  at any node.}

%reduce preserve privacy and require exchanging intermediate parameters among clients to calculate the gradients for updating local model parameters. Besides, most existing feature-based FL algorithms are designed for only two clients~\cite{yang2019federated,hardy2017private,yang2019parallel}. Reference \cite{chen2020vafl} is the only work for an arbitrary number of clients.

SGD has long been used for obtaining stationary points of unconstrained stochastic optimization problems~\cite{Robbins1951stochastic} or Karush-Kuhn-Tucker (KKT) points of stochastic optimization problems with deterministic convex constraints~\cite{bottou2018optimization}. Recently, stochastic successive convex approximation (SSCA) has been proposed to obtain KKT points of stochastic optimization problems with deterministic convex constraints~\cite{Yang} and with general stochastic nonconvex constraints~\cite{Liu,Ye}. Apparently, SSCA applies to more types of constraints. Besides,  SSCA empirically achieves a higher convergence speed than SGD~\cite{Yang}.\footnote{{SGD utilizes first-order information of a sample estimate of the objective function and usually oscillates across narrow ravines. In contrast, SSCA uses a convex approximation of an incremental sample estimate of the objective/constraint function (reflecting more information) and effectively mitigates oscillations.}} Notice that  \cite{Yang,Liu,Ye} use only one sample at each iteration and may converge slowly when applied to machine learning problems with large datasets.
Some recent works~\cite{di2016parallel,scardapane2018stochastic,DSSCA} have combined the SSCA algorithm in~\cite{Yang} and mini-batch techniques to solve unconstrained or convex constrained machine learning problems.
However,
%it is still unknown whether mini-batch techniques can be combined with the SSCA algorithms in~\cite{Liu,Ye} for solving machine learning problems with general stochastic nonconvex constraints.
%Besides,
SSCA has {never} been used for solving {machine learning problems with  nonconvex constraints or} federated optimization problems.

In summary, there are several interesting questions: 1) whether mini-batch SSCA can apply to a broader range of federated optimization problems than mini-batch SGD, 2) whether mini-batch SSCA can converge faster than mini-batch SGD, and 3) whether mini-batch SSCA can {reduce privacy risk} in FL{, like mini-batch SGD}.
In this paper, we would like to address the above questions.
Specifically, we investigate general sample-based and feature-based federated optimization, respectively. For each of them, we consider {both unconstrained and constrained {nonconvex} problems.} The main contributions are summarized as follows.

\begin{itemize}[leftmargin=\parindent, labelwidth=\parindent, labelsep=0pt, align=left]
\item {We propose FL algorithms for solving four federated optimization problems: unconstrained sample-based, constrained sample-based, unconstrained feature-based, and constrained feature-based federated optimization, using mini-batch SSCA. We show that the proposed FL algorithms converge to stationary points and KKT points of the respective unconstrained and constrained problems, respectively. Moreover, the proposed FL algorithms can adequately exploit the structures of the objective and constraint functions and incrementally utilize samples to improve convergence speeds. {They can also} reduce privacy risk through the model aggregation mechanism, and their security can be enhanced via additional privacy mechanisms.}

\item We provide an example for each proposed FL algorithm. The algorithm examples for unconstrained sample-based and feature-based federated optimization have closed-form updates and achieve the same computational complexity (in order) and communication load per communication round as the corresponding SGD-based ones in \cite{mcmahan2017communication,yu2019parallel,9003425} and \cite{hardy2017private}, respectively. Besides, the algorithm examples and FL algorithms via momentum SGD with diminishing stepsizes perform identically, which is a rather surprising result.

	\item  We consider two application examples in classification and customize the proposed FL algorithms to them. We show that {the updates in the algorithms for the four federated optimization problems all have closed-form expressions.} We also characterize the relationship between the two formulations.
	
	\item Numerical experiments demonstrate that  {in general},  the proposed {mini-batch} SSCA-based FL algorithms for unconstrained federated optimization converge faster {and achieve better computation and communication tradeoffs} than the existing SGD-based ones~\cite{mcmahan2017communication,yu2019parallel,9003425,hardy2017private}. Furthermore, numerical experiments show that the proposed {mini-batch SSCA-based} FL algorithms for constrained federated optimization can more flexibly specify a training model.
\end{itemize}

To the best of our knowledge, this is the first work that applies SSCA to solve federated optimization,  {resolves constrained nonconvex federated optimization, and establishes an analytical connection between SSCA and momentum SGD.}
{The key notation used in this paper is listed in Table \ref{tab:notation}.}

\begin{table}[h]
  \scriptsize{\begin{tabular}{|m{0.13\textwidth}<{\centering}|m{0.3\textwidth}<{\centering}|}
      \hline
      Notation & Description \\ \hline
          $I$ ($\mathcal I$) & number (index set) of clients \\ \hline
           $N$ ($\mathcal N$) & number (index set)  of samples \\ \hline
            $K$ & dimension of the vector for each sample \\ \hline
      $K_i$ & dimension of the $i$-th subvector for each sample\\ \hline
      $\mathcal N_i$ & index set of  samples  at client $i$ \\ \hline
            $B$ & batch size \\ \hline
      $\mathbf{x}_n$ & vector for the $n$-th sample \\ \hline
      $\mathbf{x}_{n,i}$ & the $i$-th subvector for the $n$-th sample \\ \hline
      $\boldsymbol\omega$ & model parameters \\ \hline
      $F_{a,m}(\boldsymbol\omega)$ & objective or constraint  function   \\ \hline
        $f_{a,m}(\boldsymbol\omega;\mathbf{x}_n)$  & {loss} for the $n$-th sample \\ \hline
      $\bar F^{(t)}_{a,m}(\boldsymbol\omega)$ & convex approximation  of $F_{a,m}(\boldsymbol\omega)$ at iteration $t$\\ \hline
      $\bar{f}_{a,m}(\boldsymbol\omega;{\boldsymbol\omega}',\mathbf{x}_n)$ & convex approximation of $f_{a,m}(\boldsymbol\omega;\mathbf{x}_n)$ around ${\boldsymbol\omega}'$ \\
      \hline
    \end{tabular}}\\
  \caption{{Key notation. $a=s$ and $a=f$ represent sample-based and feature-based, respectively. $m=0$ and {$m=1,2\cdots$} represent the objective and {$m$-th} constraint, respectively.}}
     \label{tab:notation}
\end{table}

\section{System Setting}
Consider $N$ data samples, {denoted by $\mathbf{x}_n\in\mathbb{R}^{K}, n\in\mathcal N\triangleq\{1,\cdots,N\}$.} Consider a central server connected with $I$ local clients, each maintaining  a local dataset.\footnote{{The proposed SSCA-based algorithms can be used for solving federated optimization problems over streaming data and have theoretical convergence guarantees if the properties of the data stream do not change over time.}}
Assume that the server and clients are honest-but-curious.\footnote{The nodes will follow {a predetermined algorithm} but will attempt to infer private {data} using  {information} received throughout the {algorithm} execution~\cite{hnc}.}
 {The server and $I$ clients conduct FL, i.e., collaboratively train a model from the local datasets stored on the $I$ clients under the condition that each client cannot expose its local raw data to {the server or the other clients}.}
% {This paper investigates sample-based and feature-based FL, respectively.}
Depending on whether data is distributed over the sample space or feature space, FL can be typically classified into sample-based FL and feature-based FL.
%, as shown in Fig.~\ref{fig:model}.
%\begin{figure}[h]
%	\begin{center}
%		\subfigure[\scriptsize{Sample-based FL.}\label{fig:modela}]
%		{\resizebox{7.5cm}{!}{\includegraphics{eps/modela.eps}}}\\
%		\subfigure[\scriptsize{Feature-based FL.}\label{fig:modelb}]
%		{\resizebox{8cm}{!}{\includegraphics{eps/modelb.eps}}}
%	\end{center}
%	\caption{\small{Categorization of FL.}}
%	\label{fig:model}
%\end{figure}

In sample-based FL, the clients have the same feature space but differ in the sample space.
Specifically, partition $\mathcal{N}$ into {$I$} disjoint subsets, denoted by $\mathcal{N}_i$, $i\in\mathcal{I}\triangleq\{1,\cdots,I\}$, where $N_i\triangleq|\mathcal{N}_i|$ denotes the cardinality of the $i$-th subset and $\sum_{i\in\mathcal{I}}N_i=N$.
For all $i\in\mathcal{I}$, the $i$-th client maintains a local dataset containing $N_i$ samples, i.e., $\mathbf{x}_n$, $n\in\mathcal{N}_i$.
%Specifically, for all $i\in\mathcal{I}\triangleq\{1,\cdots,I\}$, the $i$-th client maintains a local dataset containing $N_i$ samples, i.e., $\mathbf{x}_n\in\mathbb{R}^{K}$, $n\in\mathcal{N}_i\subseteq\mathcal{N}$ with $|\mathcal{N}_i|=N_i$ being the cardinality of the set.
For example, two companies with similar {businesses} in different cities may have different user groups (from their respective regions) but the same type of data, e.g., users' occupations, ages, incomes, deposits, etc.
%, as illustrated in Fig.~\ref{fig:modela}.
%The server and $I$ clients collaboratively train a model from the local datasets stored on the $I$ clients under the condition that each client cannot expose its local raw data to the others.
%This training process is referred to as sample-based (horizontal) FL~\cite{yang2019federated}.
The underlying optimization, termed sample-based federated optimization, is to minimize the following {loss} function:
\begin{align}
	&F_{s,0}(\boldsymbol\omega)\triangleq
	\frac{1}{N}\sum_{n\in\mathcal{N}}
	f_{s,0}(\boldsymbol\omega;\mathbf{x}_n)\label{eqn:Fs0}
\end{align}
with respect to (w.r.t.) model parameters $\boldsymbol\omega\in\mathbb{R}^d$. {Here, $f_{s,0}(\boldsymbol\omega;\mathbf{x}_n)$ represents the loss function for sample $\mathbf{x}_n$.}
%To be general, we do not assume $F_{s,0}(\boldsymbol\omega)$ to be convex in $\boldsymbol\omega$.

In feature-based FL, the clients have the same sample space but differ in the feature space.
Specifically, for all $n\in\mathcal{N}$,  $\mathbf{x}_n$ {can be equivalently expressed by $I$ subvectors of it}, denoted by $\mathbf{x}_{n,i}\in\mathbb{R}^{K_i}$, $i\in\mathcal{I}$, where {$\sum_{i\in\mathcal{I}}K_i\geq K$}.\footnote{{For unsupervised learning, $\mathbf{x}_{n,i}\in\mathbb{R}^{K_i}$, $i\in\mathcal{I}$ do not share any common coordinates of $\mathbf{x}_n$. For supervised learning, $\mathbf{x}_{n,i}\in\mathbb{R}^{K_i}$, $i\in\mathcal{I}$ share some common coordinates of $\mathbf{x}_n$, which represent the label of $\mathbf{x}_n$.}} {With a slight abuse of notation, we write} $\mathbf{x}_n=(\mathbf{x}_{n,i})_{i\in\mathcal{I}}$. For all $i\in\mathcal{I}$, the $i$-th client maintains  $\mathbf{x}_{n,i}$, $n\in\mathcal{N}$.\footnote{{The assumption that} the $I$ local datasets share the same set of $N$ samples can be easily met using private set intersection techniques~\cite{PrivateSet2017,PrivateSet2018}.}
For example, two companies in the same city with different businesses may have the same user group but different data types (from different types of {businesses}), e.g., one stores users' occupations and ages, and the other stores users' incomes and deposits.
%, as illustrated in Fig.~\ref{fig:modelb}.
%Similarly, the server and $I$ clients collaboratively train a model under the condition that each client cannot expose its local raw data to the others. This training process is referred to as feature-based (vertical) FL~\cite{yang2019federated}.
The underlying optimization, termed feature-based federated optimization, is to minimize the following {loss}  function:
\begin{align}
	&F_{f,0}(\boldsymbol\omega)\triangleq\frac{1}{N}\sum_{n\in\mathcal{N}} \underbrace{g_0\left(\boldsymbol\omega_0,\left(\mathbf{h}_{0,i}(\boldsymbol\omega_i,\mathbf{x}_{n,i})\right)_{i\in\mathcal{I}}\right)}_{\triangleq f_{f,0}(\boldsymbol\omega;\mathbf{x}_n)}\label{eqn:Ff0}
\end{align}
{w.r.t.} model parameters $\boldsymbol\omega\triangleq(\boldsymbol\omega_i)_{i=0,1,\cdots,I}\in\mathbb{R}^d$, where $\boldsymbol\omega_i\in\mathbb{R}^{d_i}$, $i=0,1,\cdots,I$ {and $\sum_{i=0}^I d_i=d$}. {Here, $f_{f,0}(\boldsymbol\omega;\mathbf{x}_n)$ represents the loss function for sample $\mathbf{x}_n$, formed by composing  $g_0:\mathbb{R}^{d_0+H_0I}\to\mathbb{R}$ with  functions $\mathbf{h}_{0,i}:\mathbb R^{d_i+K_i}\to\mathbb{R}^{H_0},i\in\mathcal{I}$, for some positive integer $H_0$. That is, we assume that the $i$-th block of model parameters, $\boldsymbol\omega_i$, and the $i$-th {subvector for} the $n$-th sample, $\mathbf{x}_{n,i}$, influence the loss of the $n$-th sample only via $\mathbf{h}_{0,i}(\boldsymbol\omega_i,\mathbf{x}_{n,i})$. We impose this additional restriction to enable privacy risk reduction via model aggregation  in feature-based FL. It is worth noting that the existing works on  feature-based FL impose the same restriction \cite{chen2020vafl} or even stronger restrictions} (e.g., $I=2$~\cite{yang2019federated,hardy2017private,yang2019parallel} and the loss function is the mean square error function~\cite{yang2019federated} or cross-entropy function~\cite{hardy2017private,yang2019parallel}).
% the loss function is either a special case of $F_{f,0}(\boldsymbol\omega)$ (e.g., $I=2$~\cite{yang2019federated,hardy2017private,yang2019quasi} and the loss function is the mean square error function~\cite{yang2019federated} or cross-entropy function~\cite{hardy2017private,yang2019quasi}), or shares the same form as $F_{f,0}(\boldsymbol\omega)$~\cite{chen2020vafl}.
%To be general, $F_{f,0}(\boldsymbol\omega)$ is not assumed to be convex in $\boldsymbol\omega$.

In Section~\ref{sec:sample} and Section~\ref{sec:feature}, we investigate sample-based FL and feature-based FL, respectively. {To be general, we do not assume $F_{s,0}(\boldsymbol\omega)$ and $F_{f,0}(\boldsymbol\omega)$  to be convex in $\boldsymbol\omega$.}
To guarantee the convergence of the proposed FL algorithms, we assume that $f_{s,0}\left(\boldsymbol\omega;\mathbf{x}_n\right)$ and $f_{f,0}\left(\boldsymbol\omega;\mathbf{x}_n\right)$ satisfy the following assumption  in the rest of the paper.\footnote{{In Assumptions~\ref{asump:f} and \ref{asump:fbar}, we omit the subscripts $s,f$ for notation simplicity. Note that Assumptions~\ref{asump:f} and \ref{asump:fbar} are necessary for the convergence of SSCA~\cite{Yang,Liu,Ye}, and Assumption~\ref{asump:f} is necessary for the convergences of SGD~{\cite{yu2019parallel,Robbins1951stochastic,bottou2018optimization
}} and its variants~\cite{9003425}.}}
\begin{Asump}[Assumption on $f(\boldsymbol\omega;\mathbf{x})$]\label{asump:f}
	{For any $\mathbf{x}\in \mathbb R^K$,}
	$f(\boldsymbol\omega;\mathbf{x})$ is continuously differentiable, and its gradient is Lipschitz continuous {on any compact set}.
	
	%			\mbox{}\par
	%			\begin{enumerate}
	%				\item $\boldsymbol{\Omega}$ is compact and convex;
	%				\item For any given $\mathbf{x}$, each $f(\boldsymbol\omega;\mathbf{x})$ is continuously differentiable, and its gradient is Lipschitz continuous.	
	%			\end{enumerate}
\end{Asump}
%\begin{Rem}[Discussion on Assumption~\ref{asump:f}]

%	%	In this paper, we shall propose privacy-preserving sample-based FL algorithms using mini-batch SSCA. Assumption~\ref{asump:f} is necessary for the convergence of SSCA~\cite{Yang,Liu}. All existing privacy-preserving algorithms for sample-based federated optimization~\cite{mcmahan2017communication,yang2019scheduling,yu2019parallel} are based on SGD or its variants, whose convergence also relies on Assumption~\ref{asump:f}.
%	%.2. That is, we need an extra assumption, i.e., Assumption~\ref{asump:f}.1, compared to~\cite{mcmahan2017communication,yang2019scheduling,yu2019parallel}. However, it is worth noting that Assumption~\ref{asump:f}.1 is usually met in practical applications, or some simple extra constraints can be incorporated to make $\boldsymbol{\Omega}$ compact without destroying the optimality.
%%\end{Rem}
%
%
%%\begin{Asump}[Assumptions on $f(\boldsymbol\omega;\mathbf{x})$]\label{asump:f}
%%	i) $\boldsymbol{\Omega}$ is compact and convex;	
%%	ii) For any given $\mathbf{x}$, $f(\boldsymbol\omega;\mathbf{x})$ is continuously differentiable, and its gradient is Lipschitz continuous.
%%	%		\mbox{}\par
%%	%		\begin{enumerate}
%%	%			\item $\boldsymbol{\Omega}$ is compact and convex;
%%	%			\item For any given $\mathbf{x}$, each $f(\boldsymbol\omega;\mathbf{x})$ is continuously differentiable, and its gradient is Lipschitz continuous.	
%%	%		\end{enumerate}
%%\end{Asump}

\section{Sample-based Federated Learning}\label{sec:sample}
In this section, we
%investigate two types of sample-based federated optimization, namely, unconstrained sample-based federated optimization and constrained sample-based federated optimization. For each optimization problem, we propose a federated learning algorithm using mini-batch SSCA.
{propose FL algorithms for unconstrained and constrained sample-based federated optimization problems, respectively, using mini-batch SSCA. In sample-based FL, the batch size $B$ satisfies $B\leq N_i, i\in\mathcal I$.}

\subsection{Sample-based Federated Learning for Unconstrained Optimization}\label{subsec:uncon-sample}
In this part, we consider the following unconstrained sample-based federated optimization problem:
\begin{Prob}[Unconstrained Sample-based Federated Optimization]\label{Prob:uncon-sample}
\begin{align}
&\min_{\boldsymbol\omega} \quad F_{s,0}(\boldsymbol\omega)\nonumber
\end{align}
where $F_{s,0}(\boldsymbol\omega)$ is given by~\eqref{eqn:Fs0}.
\end{Prob}

{In \cite{mcmahan2017communication,yu2019parallel,9003425}, SGD is utilized to obtain a  {stationary} point of Problem~\ref{Prob:uncon-sample}. SSCA can empirically achieve a higher convergence speed than SGD, as illustrated in Section~\ref{sec:intro}.}
%as SGD utilizes only the first-order information of an objective function.
%Problem~\ref{Prob:uncon-sample} is a nonconvex stochastic optimization problem, as $F_0(\boldsymbol\omega)$ is in general non-convex and the summation of $F_0(\boldsymbol\omega)$ is usually not computationally affordable.
%SGD and its variants, which only the first-order information of the objective function, are widely used in solving stochastic optimization problem~\cite{bottou2018optimization}.
%However, it has been shown in~\cite{Yang} that SSCA can empirically achieve higher convergence speed than stochastic gradient-based methods by exploiting the structure of the objective function and replacing its linearization with a “better” approximant.
In the following, we propose a sample-based FL algorithm, i.e., Algorithm~\ref{alg:uncon-sample}, to obtain a {stationary} point of Problem~\ref{Prob:uncon-sample} using mini-batch SSCA.\footnote{{A machine learning  problem  {involving a huge number of samples} is usually transformed to an equivalent stochastic optimization problem and solved using stochastic optimization algorithms.}}
%{We shall show that Algorithm~\ref{alg:uncon-sample} can achieve the same computational complexity (in order) and communication load per communication round, the same level of privacy protection, and an empirically high convergence speed, compared with the existing sample-based FL algorithms via SGD~\cite{mcmahan2017communication,yu2019parallel,9003425}.}

%as Later in Section~\ref{sec:simu}, we shall numerically show that the proposed SSCA-based algorithm converges faster than the SGD-based algorithms in~\cite{mcmahan2017communication,yang2019scheduling,yu2019parallel}.

\subsubsection{Algorithm Description}
The main idea of Algorithm~\ref{alg:uncon-sample} is to solve a sequence of successively refined convex problems, each of which is obtained by approximating $F_{s,0}(\boldsymbol\omega)$ with a convex function based on its structure and {randomly selected samples.}
%\footnote{{For each of Problems~\ref{Prob:uncon-sample} and \ref{Prob:uncon-feature} (Problems~\ref{Prob:con-sample-ep} and \ref{Prob:con-feature-ep}), the successive convex approximations are carefully designed to achieve the ultimate goal of finding a {stationary} (KKT) point of it.}}
Specifically, at iteration $t$, we choose {an incremental sample estimate}:
\begin{align}
\bar F^{(t)}_{s,0}(\boldsymbol\omega)=&(1-\rho^{(t)})\bar{F}^{(t-1)}_{s,0}(\boldsymbol\omega)\nonumber\\
&+\rho^{(t)}\sum_{i\in\mathcal{I}}\frac{N_i}{BN}\sum_{n\in\mathcal N_i^{(t)}}\bar{f}_{s,0}(\boldsymbol\omega;{\boldsymbol\omega}_{s}^{(t)},\mathbf{x}_n)\label{eqn:Fs0bar}
\end{align}
with $\bar F_{s,0}^{(0)}(\boldsymbol\omega)=0$ as {a convex} approximation function of $F_{s,0}(\boldsymbol\omega)$,
where $\rho^{(t)}$ is a stepsize satisfying:
\begin{align}
	&{0<\rho^{(t)}\leq 1},\ \lim_{t\to\infty}\rho^{(t)}=0,\ \sum_{t=1}^\infty\rho^{(t)}=\infty,\label{eqn:rho}
\end{align}
$\mathcal N_i^{(t)}\subseteq\mathcal N_i$ is a randomly selected mini-batch by client $i$ at iteration $t$,   and $\bar{f}_{s,0}(\boldsymbol\omega;{\boldsymbol\omega}_{s}^{(t)},\mathbf{x}_n)$ is a convex approximation\footnote{{Usually, we  preserve all convex terms in $f_{s,0}(\boldsymbol\omega;\mathbf{x}_n)$ and properly approximates the remaining nonconvex terms for reducing the approximation error or utilize the first-order approximation of $f_{s,0}(\boldsymbol\omega;\mathbf{x}_n)$ (see \eqref{eqn:fs0bar}) for reducing the computational complexity for solving Problem~\ref{Prob:uncon-sample-ap}.}
} of $f_{s,0}(\boldsymbol\omega;\mathbf{x}_n)$ around ${\boldsymbol\omega}_{s}^{(t)}$ satisfying the following assumptions.
%A common example of $\bar{f}_{s,0}$ will be given later.

\begin{Asump}[Assumptions on $\bar{f}(\boldsymbol\omega;\boldsymbol\omega',\mathbf{x})$ for Approximating $f(\boldsymbol\omega;\mathbf{x})$ Around $\boldsymbol\omega'$]\label{asump:fbar} %For any given $\mathbf{x}$,
1) {For any $\boldsymbol\omega\in \mathbb R^d$ and $\mathbf{x}\in \mathbb R^K$,} $\nabla \bar{f}(\boldsymbol\omega;\boldsymbol\omega,\mathbf{x})=\nabla f(\boldsymbol\omega;\mathbf{x})$;
2) {For any $\boldsymbol\omega'\in \mathbb R^d$ and $\mathbf{x}\in \mathbb R^K$,} $\bar{f}(\boldsymbol\omega;\boldsymbol\omega',\mathbf{x})$ is strongly convex w.r.t. $\boldsymbol\omega$;
3) {For any $\mathbf{x}\in \mathbb R^K$,}  $\bar{f}(\boldsymbol\omega;\boldsymbol\omega',\mathbf{x})$ is Lipschitz continuous  {on  any compact set};
4) {For any $\boldsymbol\omega'\in \mathbb R^d$ and $\mathbf{x}\in \mathbb R^K$,} $\bar{f}(\boldsymbol\omega;\boldsymbol\omega',\mathbf{x})$, its {derivatives {w.r.t. $\boldsymbol\omega$}, and its second-order derivatives} w.r.t. $\boldsymbol\omega$ are {uniformly bounded} {on  any compact set}.
%\begin{enumerate}
%	\item $\bar{f}(\boldsymbol\omega,\boldsymbol\omega,\mathbf{x})=f(\boldsymbol\omega,\mathbf{x})$ and $\nabla \bar{f}(\boldsymbol\omega,\boldsymbol\omega,\mathbf{x})=\nabla f(\boldsymbol\omega,\mathbf{x})$;
%	\item $\bar{f}(\boldsymbol\omega,\boldsymbol\omega',\mathbf{x})$ is strongly convex in $\boldsymbol\omega$;
%	\item $\bar{f}(\boldsymbol\omega,\boldsymbol\omega',\mathbf{x})$ is Lipschitz continuous in both $\boldsymbol\omega$ and $\boldsymbol\omega'$;
%	\item $\bar{f}(\boldsymbol\omega,\boldsymbol\omega',\mathbf{x})$, its derivative, and its second order derivative w.r.t. $\boldsymbol\omega$ are uniformly bounded.
%\end{enumerate}
\end{Asump}

%Assumption~\ref{asump:fbar} is necessary for the convergence of SSCA~\cite{Yang,Liu,Ye}.
 Note that for all $i\in\mathcal{I}$, mini-batch $\mathcal{N}'_i\subseteq\mathcal{N}_i$ with batch size $B$, {and $\boldsymbol\omega'\in \mathbb R^d$}, $\sum_{n\in\mathcal{N}'_i}\bar{f}_{s,0}(\boldsymbol\omega;\boldsymbol\omega',\mathbf{x}_n)$,  {a function of  $\boldsymbol\omega$ with parameters jointly determined by $\boldsymbol\omega'$ and $\mathbf{x}_n,n\in\mathcal{N}'_i$,} can be written {naturally} as $\sum_{n\in\mathcal{N}'_i}\bar{f}_{s,0}(\boldsymbol\omega;\boldsymbol\omega',\mathbf{x}_n)
=p_{s,0}\left(\boldsymbol\omega, \mathbf q_{s,0}\left(\boldsymbol\omega',(\mathbf{x}_n)_{n\in\mathcal{N}'_i}\right)\right)$ with $p_{s,0}:\mathbb{R}^{d+D_0}\to\mathbb{R}$ and $\mathbf q_{s,0}:\mathbb{R}^{d+BK}\to\mathbb{R}^{D_0}$, {for some positive integer $D_0$. Here, $\mathbf q_{s,0}\left(\boldsymbol\omega',(\mathbf{x}_n)_{n\in\mathcal{N}'_i}\right)$ represents the $D_0$ parameters of $\sum_{n\in\mathcal{N}'_i}\bar{f}_{s,0}(\boldsymbol\omega;\boldsymbol\omega',\mathbf{x}_n)$.
}
Assume that the expressions of $\bar{f}_{s,0}$, $p_{s,0}$, and $\mathbf q_{s,0}$ are known to the server and {$I$} clients. Each client $i\in\mathcal{I}$ computes $\mathbf q_{s,0}\left({\boldsymbol\omega}_{s}^{(t)},(\mathbf{x}_n)_{n\in\mathcal{N}^{(t)}_i}\right)$ and sends it to the server. Then, the server solves the following convex approximate problem to obtain $\bar{\boldsymbol\omega}_{s}^{(t)}$.
\begin{Prob}[Convex Approximate Problem of Problem~\ref{Prob:uncon-sample}]\label{Prob:uncon-sample-ap}
	\begin{align} &\bar{\boldsymbol\omega}_{s}^{(t)}\triangleq\mathop{\arg\min}_{\boldsymbol\omega} \ \bar F_{s,0}^{(t)}(\boldsymbol\omega)\nonumber
	\end{align}
\end{Prob}
\begin{algorithm}[t]
	\caption{Mini-batch SSCA for Problem~\ref{Prob:uncon-sample}}
		\begin{small}
	\begin{algorithmic}[1]
		%           \STATE \textbf{Input}: $\{\gamma^k\}$
		\STATE \textbf{initialize}: choose any ${\boldsymbol\omega}_{s}^{1}$ at the server.\\
		\FOR{$t=1,2,\cdots,T-1$}
		\STATE the server sends ${\boldsymbol\omega}_{s}^{(t)}$ to all clients.
		\STATE for all $i\in\mathcal{I}$, client $i$ randomly selects a mini-batch $\mathcal{N}^{(t)}_i\subseteq\mathcal{N}_i$, computes $\mathbf q_{s,0}\left({\boldsymbol\omega}_{s}^{(t)},(\mathbf{x}_n)_{n\in\mathcal{N}^{(t)}_i}\right)$, and sends it to the server.
		\STATE the server obtains $\bar{\boldsymbol\omega}_{s}^{(t)}$ by solving Problem~\ref{Prob:uncon-sample-ap}.
		\STATE the server updates ${\boldsymbol\omega}_{s}^{(t+1)}$ according to \eqref{eqn:updatew}.
		\ENDFOR
		\STATE \textbf{Output}: ${\boldsymbol\omega}_{s}^T$
	\end{algorithmic}\label{alg:uncon-sample}
		\end{small}
\end{algorithm}
Problem~\ref{Prob:uncon-sample-ap} is {an unconstrained convex problem} and can be solved with %conventional convex optimization
 {decent methods such as Newton's method}.
Given $\bar{\boldsymbol\omega}_{s}^{(t)}$, the server updates ${\boldsymbol\omega}_{s}^{(t)}$ according to:
\begin{align} &{\boldsymbol\omega}_{s}^{(t+1)}=(1-\gamma^{(t)}){\boldsymbol\omega}_{s}^{(t)}+\gamma^{(t)}\bar{\boldsymbol\omega}_{s}^{(t)},\ t=1,2,\cdots \label{eqn:updatew}
\end{align}
where $\gamma^{(t)}$ is a stepsize satisfying:
\begin{align}
&{0<\gamma^{(t)}\leq 1},\ \lim_{t\to\infty}\gamma^{(t)}=0,\ \sum_{t=1}^\infty\gamma^{(t)}=\infty,\nonumber\\
&\quad\quad\sum_{t=1}^\infty\left(\gamma^{(t)}\right)^2<\infty,\ \lim_{t\to\infty}\frac{\gamma^{(t)}}{\rho^{(t)}}=0.\label{eqn:gamma}
\end{align}
The detailed procedure is summarized in Algorithm~\ref{alg:uncon-sample}.\footnote{{Each iteration of Algorithms~\ref{alg:uncon-sample}-\ref{alg:con-feature} is implemented in one communication round. The computational complexity and communication load per communication round depend on the specific choices of $\bar{f}_{a,m}(\boldsymbol\omega;{\boldsymbol\omega}',\mathbf{x}_n)$, $a=s,f$ and $m=0,1,\cdots,M$.}} The convergence of Algorithm~\ref{alg:uncon-sample} is summarized below.
%\footnote{{The proofs in this paper are omitted due to page limitation. Please refer to the supporting document for detailed proofs.}}
Algorithm~\ref{alg:uncon-sample} can {empirically achieve a} high convergence speed (shown in Section~\ref{sec:simu}), as it can adequately exploit the structure of the objective function and incrementally utilize samples.
\begin{Thm}[Convergence of Algorithm~\ref{alg:uncon-sample}]\label{thm:uncon-sample}
	Suppose that $f_{s,0}$ satisfies Assumption~\ref{asump:f}, $\bar{f}_{s,0}$ satisfies Assumption~\ref{asump:fbar}, and the sequence $\{{\boldsymbol\omega}_{s}^{(t)}\}$ generated by Algorithm~\ref{alg:uncon-sample} is bounded.\footnote{{The conclusion of Theorems \ref{thm:uncon-sample}-\ref{thm:con-feature} still holds if the boundedness condition of the sequence  in the theorem  is replaced with the compact set constraint on $\boldsymbol\omega$ in the corresponding problem\cite{Yang,Liu,Ye}. Note that the boundedness condition is easily satisfied in numerical experiments, and a simple compact set constraint that is sufficiently large can always be imposed without destroying the optimality\cite{Liu,Ye}.}} Then, every limit point of $\{{\boldsymbol\omega}_{s}^{(t)}\}$ is a {stationary} point of Problem~\ref{Prob:uncon-sample} almost surely.
\end{Thm}

\begin{IEEEproof}{Please refer to Appendix A.}
\end{IEEEproof}

\subsubsection{Security Analysis}
%We establish the security of Algorithm~\ref{alg:uncon-sample}.
If for all $i\in\mathcal{I}$, mini-batch $\mathcal{N}'_i\subseteq\mathcal{N}_i$,  {and $\boldsymbol\omega'\in \mathbb R^d$}, the system of equations w.r.t. $\mathbf{z}\in\mathbb{R}^{BK}$, i.e., $\mathbf q_{s,0}\left(\boldsymbol\omega',\mathbf{z}\right)=\mathbf q_{s,0}\left(\boldsymbol\omega',(\mathbf{x}_n)_{n\in\mathcal{N}'_i}\right)$, has an infinite (or a sufficiently large) number of solutions, then raw data $\mathbf{x}_n$, $n\in\mathcal{N}^{(t)}_i$ {can hardly} be extracted {by the server} from $\mathbf q_{s,0}\left({\boldsymbol\omega}_{s}^{(t)},(\mathbf{x}_n)_{n\in\mathcal{N}^{(t)}_i}\right)$ in Step 4 of Algorithm~\ref{alg:uncon-sample}, {and hence, Algorithm~\ref{alg:uncon-sample} can reduce privacy risk based on model aggregation, {like} the existing sample-based FL algorithms via SGD~\cite{mcmahan2017communication,yu2019parallel,9003425}.} {Otherwise, extra privacy mechanisms can be applied to preserve data privacy.} For example, if $\bar{\boldsymbol\omega}_{s}^{(t)}$ is linear in $\mathbf q_{s,0}\left({\boldsymbol\omega}_{s}^{(t)},(\mathbf{x}_n)_{n\in\mathcal{N}^{(t)}_i}\right)$, $i\in\mathcal{I}$, then homomorphic encryption~\cite{HE} can be applied; if $\bar{\boldsymbol\omega}_{s}^{(t)}$ is a polynomial of $\mathbf{x}_n$ and ${\boldsymbol\omega}_{s}^{(t)}$, then secret sharing~\cite{2017SecureML} can be applied.

\subsubsection{Algorithm Example}
{We} provide an example of $\bar{f}_{s,0}$ which satisfies Assumption~\ref{asump:fbar} and yields an analytical solution of Problem~\ref{Prob:uncon-sample-ap}:
\begin{align}
	\bar{f}_{s,0}(\boldsymbol\omega;{\boldsymbol\omega}_{s}^{(t)}\!,\mathbf{x}_n)\!=&\!\left(\nabla f_{s,0}({\boldsymbol\omega}_{s}^{(t)}\!;\mathbf{x}_n)\right)^T\!\!\left(\boldsymbol\omega\!-\!{\boldsymbol\omega}_{s}^{(t)}\right)\!\nonumber\\
	&+\!\tau\left\Vert{\boldsymbol\omega\!-\!{\boldsymbol\omega}_{s}^{(t)}}\right\Vert_2^2, \label{eqn:fs0bar}
\end{align}
where $\tau>0$ can be any constant, and the term $\tau\left\Vert{\boldsymbol\omega-{\boldsymbol\omega}_{s}^{(t)}}\right\Vert_2^2$ is used to ensure strong convexity.
%Obviously, $\bar{f}_{s,0}$ given by~\eqref{eqn:fs0bar} satisfies Assumption~\ref{asump:fbar}.
{Then, $\sum_{n\in\mathcal{N}^{(t)}_i}\nabla f_{s,0}({\boldsymbol\omega}_{s}^{(t)};\mathbf{x}_n)$ can be viewed as $\mathbf q_{s,0}\left({\boldsymbol\omega}_{s}^{(t)},(\mathbf{x}_n)_{n\in\mathcal{N}^{(t)}_i}\right)$ {(implying $D_0=d$).}} Furthermore, substituting \eqref{eqn:fs0bar} into \eqref{eqn:Fs0bar}, $\bar{F}^{(t)}_{s,0}(\boldsymbol\omega)$ can be rewritten as:
\begin{align} \bar{F}^{(t)}_{s,0}(\boldsymbol\omega)=&\left(\hat{\mathbf f}^{(t)}_{s,0,1}\right)^T\boldsymbol\omega+\tau\left\Vert{\boldsymbol\omega}\right\Vert_2^2,\label{eqn:Fs0bar_eg}
\end{align}
where $\hat{\mathbf f}^{(t)}_{s,0,1}\in\mathbb{R}^d$ is given by:
\begin{align}
&\hat{\mathbf f}^{(t)}_{s,0,1}=(1-\rho^{(t)})\hat{\mathbf f}^{(t-1)}_{s,0,1}\nonumber\\ &+\rho^{(t)}\sum_{i\in\mathcal{I}}\frac{N_i}{BN}\sum_{n\in\mathcal N_i^{(t)}}\Big(\nabla f_{s,0}({\boldsymbol\omega}_{s}^{(t)};\mathbf{x}_n)-2\tau{\boldsymbol\omega}_{s}^{(t)}\Big)\label{eqn:sample-accu-hat-f}
\end{align}
with $\hat{\mathbf f}^{(0)}_{s,0,1}=\mathbf 0$. Apparently, Problem~\ref{Prob:uncon-sample-ap} with $\bar{f}_{s,0}$ given by~\eqref{eqn:fs0bar} is an unconstrained convex quadratic programming w.r.t. $\boldsymbol\omega$. By the first-order optimality condition, it has the following analytical solution:
\begin{align} \bar{\boldsymbol\omega}_{s}^{(t)}=-\frac{1}{2\tau}\hat{\mathbf f}^{(t)}_{s,0,1}.\label{eqn:omegasbar}
\end{align}

 {Therefore, Step 4 and Step 5 of Algorithm~\ref{alg:uncon-sample} with $\bar{f}_{s,0}$ given by~\eqref{eqn:fs0bar}  (i.e., an example of Algorithm~\ref{alg:uncon-sample})  are given below.
In Step 4, each client $i\in \mathcal I$ computes $\sum_{n\in\mathcal{N}^{(t)}_i}\nabla f_{s,0}({\boldsymbol\omega}_{s}^{(t)};\mathbf{x}_n)\in \mathbb R^d$ and sends the $d$-dimensional vector to the server. In Step 5, the server calculates $\bar{\boldsymbol\omega}_{s}^{(t)}$ according to ~\eqref{eqn:omegasbar}.} If for all $i\in\mathcal{I}$, $\mathcal{N}'_i\subseteq\mathcal{N}_i$, {and $\boldsymbol\omega'\in \mathbb R^d$}, the system of equations w.r.t. $(\mathbf{z}_n)_{n=1\cdots,B}$ {with} $\mathbf{z}_n\in\mathbb{R}^{K}$, $n=1,\cdots,B$, i.e., $\sum_{n=1}^B\nabla f_{s,0}({\boldsymbol\omega}';\mathbf{z}_n)=\sum_{n\in\mathcal{N}'_i}\nabla f_{s,0}({\boldsymbol\omega}';\mathbf{x}_n)$, has an infinite (or a sufficiently large) number of solutions, then {the example of Algorithm~\ref{alg:uncon-sample}} can {reduce privacy risk}. Otherwise, homomorphic encryption~\cite{HE,hardy2017private} can be applied to preserve data privacy, since $\bar{\boldsymbol\omega}_{s}^{(t)}$ is linear in $\nabla f_{s,0}({\boldsymbol\omega}_{s}^{(t)};\mathbf{x}_n)$, $n\in\mathcal{N}$, as shown in{~\eqref{eqn:sample-accu-hat-f} and}~\eqref{eqn:omegasbar}.

\begin{Rem}[Comparison Between Example of Algorithm~\ref{alg:uncon-sample} and Sample-based FL Algorithms via SGD and Its Variants~\cite{mcmahan2017communication,yu2019parallel,9003425}]
Algorithm~\ref{alg:uncon-sample} with $\bar{f}_{s,0}$ given by~\eqref{eqn:fs0bar} {has the same order of  computational complexity ($\mathcal O(B)$)}  and  communication load per communication round as the sample-based FL algorithms via SGD and its variants~\cite{mcmahan2017communication,yu2019parallel,9003425}, where $B$ samples are utilized by each client per communication round. Besides, it has the same level of privacy protection (due to the same system of equations for inferring private data) as
the  sample-based  algorithm via SGD {and its variants} with one local SGD update per communication round {(e.g., FedSGD~\cite{mcmahan2017communication})}.\end{Rem}

{Finally, by \eqref{eqn:updatew}, \eqref{eqn:sample-accu-hat-f}, and \eqref{eqn:omegasbar} and by choosing $\rho^{(1)}=1$, $\{{\boldsymbol\omega_s}^{(t)}\}$ generated by the example of Algorithm~\ref{alg:uncon-sample} satisfies:
\begin{align}
&{\boldsymbol\omega_s}^{(t+1)}
={\boldsymbol\omega_s}^{(t)}-\gamma^{(t)}\mathbf v_s^{(t)},\ t=1,2,\cdots \label{eq:omega_iterate}\\
&\mathbf v_s^{(t)}
=\left(1-\rho^{(t)}\right)\left(1-\gamma^{(t-1)}\right)\mathbf v_s^{(t-1)}\nonumber\\
&+\frac{\rho^{(t)}}{2\tau}\!\sum_{i\in\mathcal{I}}\!\!\frac{N_i}{BN}\!\!\!\!\sum_{n\in\mathcal N_i^{(t)}}\!\!\!\nabla f_{s,0}({\boldsymbol\omega}_{s}^{(t)};\mathbf{x}_n),\ t=1,2,\cdots \label{eq:v_iterate}
\end{align}
where $\gamma^{(0)}=0$,  $\mathbf v_s^{(0)}=0$, and  $\rho^{(t)}$ and $\gamma^{(t)}$ satisfy \eqref{eqn:Fs0bar} and \eqref{eqn:updatew}, respectively. From \eqref{eq:omega_iterate} and \eqref{eq:v_iterate}, we can make the following remark.}
{\begin{Rem}[Connection Between Example of Algorithm~\ref{alg:uncon-sample} and Sample-based FL Algorithms via Momentum SGD\cite{9003425}] Algorithm~\ref{alg:uncon-sample} with $\bar{f}_{s,0}$ given by~\eqref{eqn:fs0bar}  can also be viewed as sample-based FL algorithm  via momentum SGD with the momentum term $\mathbf v_s^{(t)}$ and diminishing stepsize $\gamma^{(t)}$ being the update direction and stepsize, respectively. This result also reveals an analytical connection between SSCA and momentum SGD, which is established for the first time.  Furthermore, since the existing momentum SGD algorithms  \cite{9003425, QIAN1999SGDmom} with theoretical convergence guarantees all rely on constant stepsizes,  this work also enriches the results for momentum SGD. \end{Rem}}

%%The details of the analytical solution will be given in Section~\ref{sec:application}.
%\begin{Rem}[Comparison Between Example of Algorithm~\ref{alg:uncon-sample} and Sample-based FL Algorithms via SGD~\cite{mcmahan2017communication,yu2019parallel,9003425}]
%%	Notice that under the sample-based FL algorithms via SGD~\cite{mcmahan2017communication,yang2019scheduling,yu2019parallel}, each client conducts one or multiple {local} SGD updates, and the server averages the locally computed models. Thus,
%Algorithm~\ref{alg:uncon-sample} with $\bar{f}_{s,0}$ given by~\eqref{eqn:fs0bar} and the sample-based FL algorithm via SGD~\cite{mcmahan2017communication,yu2019parallel,9003425} {have the same computational complexity (in order)  and  communication load per communication round and the same level of privacy protection.} 	
%%	at each client and the server have the same order, and the computational complexities at the server have the same order (which is $\mathcal{O}(d)$); the communication loads are identical
%%	In addition, note that under the sample-based FL algorithms via SGD~\cite{mcmahan2017communication,yang2019scheduling,yu2019parallel}, each client sends its updated local model to the server.
%%	Thus, under the sample-based FL algorithms via SGD with one SGD update at each client in each communication round and under Algorithm~\ref{alg:uncon-sample} with $\bar{f}_{s,0}$ given by~\eqref{eqn:fs0bar}, the requirements for privacy-preserving are the same.
%\end{Rem}

\subsection{Sample-based Federated Learning for Constrained Optimization}
In this part, we consider the following constrained sample-based federated optimization problem:
\begin{Prob}[Constrained Sample-based Federated Optimization]\label{Prob:con-sample}
\begin{align}
\min_{\boldsymbol\omega}\ &\quad F_{s,0}(\boldsymbol\omega)\nonumber\\
\text{s.t.}
&\quad F_{s,m}(\boldsymbol\omega)\leq0,\ m=1,2,\cdots,M,\nonumber
\end{align}
where $F_{s,0}(\boldsymbol\omega)$ is given by~\eqref{eqn:Fs0}, and
\begin{align} F_{s,m}(\boldsymbol\omega)\triangleq\frac{1}{N}\sum_{n\in\mathcal{N}} f_{s,m}(\boldsymbol\omega;\mathbf{x}_n),\ m=1,2,\cdots,M.
\end{align}
\end{Prob}

To be general, $F_{s,m}(\boldsymbol\omega)$, $m=0,\cdots,M$ are not assumed to be convex in $\boldsymbol\omega$. Notice that federated optimization with nonconvex constraints has not been investigated so far.
%It is quite challenging, as the stochastic nature of a constraint function may cause infeasibility at each iteration of an ordinary stochastic iterative method~\cite{Ye}.
In the following, we propose a sample-based FL algorithm, i.e., Algorithm~\ref{alg:con-sample}, to obtain a KKT point of Problem~\ref{Prob:con-sample}, by combining the exact penalty method for SSCA in our previous work~\cite{Ye} and mini-batch techniques.

\subsubsection{Algorithm Description} {Sample convex approximations of Problem~\ref{Prob:con-sample}, obtained by directly approximating $F_{s,m}(\boldsymbol\omega),m=0,1,\cdots,M$ with the method proposed for $F_{s,0}$ in Section~\ref{subsec:uncon-sample}, may not always be feasible, leading to possibly infeasible stochastic iterates~\cite{Liu,Ye}. To ensure feasible stochastic iterates,} we first transform Problem~\ref{Prob:con-sample} to the following stochastic optimization problem whose objective function is the weighted sum of the original objective and the penalty for violating the original constraints~\cite{Ye}. We will soon see that its  sample convex approximations  are always feasible.

\begin{Prob}[Transformed Problem of Problem~\ref{Prob:con-sample}]\label{Prob:con-sample-ep}
	\begin{align}
\min_{\boldsymbol\omega,\mathbf s}
\quad&F_{s,0}(\boldsymbol\omega)+c\sum_{m=1}^M s_m\nonumber\\
\text{s.t.}
\quad&F_{s,m}(\boldsymbol\omega)\leq s_m,\ m=1,2,\cdots,M,\nonumber\\
&s_m\geq 0,\ m=1,2,\cdots,M,\nonumber
\end{align}
where $\mathbf{s}\triangleq(s_m)_{m=1,\cdots,M}$ are slack variables, and $c>0$ is a penalty parameter that trades off the original objective function and the slack penalty term.
\end{Prob}

At iteration $t$, we choose $\bar F^{(t)}_{s,0}(\boldsymbol\omega)$ given in~\eqref{eqn:Fs0bar} as an approximation function of $F_{s,0}(\boldsymbol\omega)$ and choose:
\begin{align}
&\bar F^{(t)}_{s,m}(\boldsymbol\omega)=(1-\rho^{(t)})\bar{F}^{(t-1)}_{s,m}(\boldsymbol\omega)\nonumber\\
&+\rho^{(t)}\!
\sum_{i\in\mathcal{I}}\frac{N_i}{BN}\!\!\!\sum_{n\in\mathcal N_i^{(t)}}\!\!\!\bar{f}_{s,m}(\boldsymbol\omega;{\boldsymbol\omega}_{s}^{(t)},\mathbf{x}_n),\ m=1,\cdots,M\label{eqn:Fsmbar}
\end{align}
with $\bar F_{s,m}^{(0)}(\boldsymbol\omega)=0$ as {a convex} approximation function of $F_{s,m}(\boldsymbol\omega)$, for all $m=1,\cdots,M$, where $\rho^{(t)}$ is a stepsize satisfying~\eqref{eqn:rho},
$\mathcal N^{(t)}_i$ is a randomly selected mini-batch by client $i$ at iteration $t$,
and $\bar{f}_{s,m}(\boldsymbol\omega;{\boldsymbol\omega}_{s}^{(t)},\mathbf{x}_n)$ is a convex approximation of $f_{s,m}(\boldsymbol\omega;\mathbf{x}_n)$ around ${\boldsymbol\omega}_{s}^{(t)}$ satisfying $\bar{f}_{s,m}(\boldsymbol\omega;\boldsymbol\omega,\mathbf{x})=f_{s,m}(\boldsymbol\omega;\mathbf{x})$ and Assumption~\ref{asump:fbar} for all $m=1,\cdots,M$.
%A common example of $\bar{f}_{s,m}$, $m=0,\cdots,M$ will be given later.

Note that for all $i\in\mathcal{I}$, mini-batch $\mathcal{N}'_i\subseteq\mathcal{N}_i$ with batch size $B$, {and $\boldsymbol\omega'\in \mathbb R^d$}, $\sum_{n\in\mathcal{N}'_i}\bar{f}_{s,m}(\boldsymbol\omega;\boldsymbol\omega',\mathbf{x}_n)$, $m=0,\cdots,M$ can be written as $\sum_{n\in\mathcal{N}'_i}\bar{f}_{s,m}(\boldsymbol\omega;\boldsymbol\omega',\mathbf{x}_n)
=p_{s,m}\left(\boldsymbol\omega, \mathbf q_{s,m}\left(\boldsymbol\omega',(\mathbf{x}_n)_{n\in\mathcal{N}'_i}\right)\right)$, $m=0,\cdots,M$ with $p_{s,m}:\mathbb{R}^{D_m+d}\to\mathbb{R}$ and $\mathbf q_{s,m}:\mathbb{R}^{BK+d}\to\mathbb{R}^{D_m}$.
Assume that the expressions of $\bar{f}_{s,m}$, $p_{s,m}$, $\mathbf q_{s,m}$, $m=0,\cdots,M$ are known to the server and {$I$} clients.  Each client $i\in\mathcal{I}$ computes $\mathbf q_{s,m}\left({\boldsymbol\omega}_{s}^{(t)},(\mathbf{x}_n)_{n\in\mathcal{N}^{(t)}_i}\right)$, $m=0,\cdots,M$ and {sends} them to the server. Then, the server solves the following {convex} approximate problem to obtain $\bar{\boldsymbol\omega}_{s}^{(t)}$.

\begin{Prob}[Convex Approximate Problem of Problem~\ref{Prob:con-sample-ep}]\label{Prob:con-sample-ap}
	\begin{align} (\bar{\boldsymbol\omega}_{s}^{(t)},\mathbf{s}_s^{(t)})\triangleq
\mathop{\arg\min}_{\boldsymbol\omega,\mathbf{s}}
\quad&\bar F^{(t)}_{s,0}(\boldsymbol\omega)+c\sum_{m=1}^M s_m\nonumber\\
\text{s.t.}\quad&\bar F^{(t)}_{s,m}(\boldsymbol\omega)\leq s_m,\quad m=1,2,\cdots,M,\nonumber\\
&s_m\geq 0,\quad m=1,2,\cdots,M.\nonumber
	\end{align}
\end{Prob}

Problem~\ref{Prob:con-sample-ap} is {a constrained convex problem that is always feasible} and can be readily solved {with
%conventional convex optimization methods such as
interior-point methods such as the barrier method}.\footnote{{Problem~\ref{Prob:con-sample-ap} can be efficiently solved by the barrier method, regardless of how large $c$ (which influences only the linear terms of {the} objective function) is. This is because, in each centering step of the barrier method, an unconstrained centering problem is solved by Newton's method, whose convergence rate depends only on the smallest and largest eigenvalues and Lipschitz constant of the Hessian matrix of the objective function.}}
Given $\bar{\boldsymbol\omega}_{s}^{(t)}$, the server updates ${\boldsymbol\omega}_{s}^{(t)}$ according to \eqref{eqn:updatew}.
\begin{algorithm}[t]
	\caption{Mini-batch SSCA for Problem~\ref{Prob:con-sample}}
		\begin{small}
	\begin{algorithmic}[1]
		%           \STATE \textbf{Input}: $\{\gamma^k\}$
		\STATE \textbf{initialize}: choose any ${\boldsymbol\omega}_{s}^{1}$ and $c>0$ at the server.\\
		\FOR{$t=1,2,\cdots,T-1$}
		\STATE the server sends ${\boldsymbol\omega}_{s}^{(t)}$ to all clients.
		\STATE for all $i\in\mathcal{I}$, client $i$ randomly selects a mini-batch $\mathcal{N}^{(t)}_i\subseteq\mathcal{N}_i$, computes $\mathbf q_{s,m}\left({\boldsymbol\omega}_{s}^{(t)},(\mathbf{x}_n)_{n\in\mathcal{N}^{(t)}_i}\right)$, $m=0,1,\cdots,M$, and sends them to the server.
		\STATE the server obtains $(\bar{\boldsymbol\omega}_{s}^{(t)}, \mathbf s_s^{(t)})$ by solving Problem~\ref{Prob:con-sample-ap}.
		\STATE the server  updates ${\boldsymbol\omega}_{s}^{(t+1)}$ according to \eqref{eqn:updatew}.
		\ENDFOR
		\STATE \textbf{Output}: ${\boldsymbol\omega}_{s}^T$
	\end{algorithmic}\label{alg:con-sample}
		\end{small}
\end{algorithm}
The detailed procedure is summarized in Algorithm~\ref{alg:con-sample}. The convergence of Algorithm~\ref{alg:con-sample} is summarized below. Consider a sequence $\{c_j\}$. For all $j$, let $({\boldsymbol\omega}_{s,j}^\star,\mathbf s_{s,j}^\star)$ denote a limit point of $\{({\boldsymbol\omega}_{s}^{(t)},\mathbf{s}_s^{(t)})\}$ generated by Algorithm~\ref{alg:con-sample} with $c=c_j$.
\begin{Thm}[Convergence of Algorithm~\ref{alg:con-sample}]\label{thm:con-sample}
	Suppose that $f_{s,m}$, $m=0,\cdots,M$ satisfy Assumption~\ref{asump:f}, $\bar{f}_{s,0}$ satisfies Assumption~\ref{asump:fbar}, $\bar{f}_{s,m}$ satisfies $\bar{f}_{s,m}(\boldsymbol\omega;\boldsymbol\omega,\mathbf{x})=f_{s,m}(\boldsymbol\omega;\mathbf{x})$ and Assumption~\ref{asump:fbar} for all $m=1,\cdots,M$, {the sequence $\{{\boldsymbol\omega}_{s}^{(t)}\}$ generated by Algorithm~\ref{alg:con-sample} with $c=c_j$ is bounded for all $j$,} and the sequence $\{c_j\}$ satisfies $0<c_j<c_{j+1}$ and $\lim_{j\to\infty}c_j=\infty$. Then, the following statements hold.
	i) For all $j$, if $\mathbf s_{s,j}^\star=\mathbf0$, then ${\boldsymbol\omega}_{s,j}^\star$ is a KKT point of Problem~\ref{Prob:con-sample} almost surely;
	%Problem~\ref{Prob:con-sample-ep};
	ii) A limit point of $\{({\boldsymbol\omega}_{s,j}^\star,\mathbf s_{s,j}^\star)\}$, denoted by $\{({\boldsymbol\omega}_{s,\infty}^\star,\mathbf s_{s,\infty}^\star)\}$, satisfies that $\mathbf s_{s,\infty}^\star=\mathbf{0}$, and ${\boldsymbol\omega}_{s,\infty}^\star$ is a KKT point of Problem~\ref{Prob:con-sample} almost surely.
	%\item If $\mathbf{x}^{(0)}$ is feasible for Problem~\ref{prob:gen}, then $\mathbf{x}^*$ is a KKT point of Problem~\ref{prob:gen} and $\mathbf s^*=\mathbf0$.
\end{Thm}

\begin{IEEEproof}{Please refer to Appendix B.}
\end{IEEEproof}

%In practice, we can choose a sequence $\{c_j\}$ which satisfies that $0<c_j<c_{j+1}$, $\lim_{j\to\infty}c_j=\infty$, and $c_1$ is large and repeat Algorithm~\ref{alg:con-sample} with $c=c_j$ until $\left\Vert\mathbf s_{s,j}^\star\right\Vert_2$ is sufficiently small.

\subsubsection{Security Analysis}
%We establish the security of Algorithm~\ref{alg:con-sample}.
If for all $i\in\mathcal{I}$, mini-batch $\mathcal{N}'_i\subseteq\mathcal{N}_i$, {and $\boldsymbol\omega'\in \mathbb R^d$},  the system of equations w.r.t. $\mathbf{z}\in\mathbb{R}^{BK}$, i.e., $\mathbf q_{s,m}\left(\boldsymbol\omega',\mathbf{z}\right)=\mathbf q_{s,m}\left(\boldsymbol\omega',(\mathbf{x}_n)_{n\in\mathcal{N}'_i}\right)$, $m=0,\cdots,M$, has an infinite (or a sufficiently large) number of solutions, then raw data $\mathbf{x}_n$, $n\in\mathcal{N}^{(t)}_i$ {can hardly} be extracted from $\mathbf q_{s,m}\left({\boldsymbol\omega}_{s}^{(t)},(\mathbf{x}_n)_{n\in\mathcal{N}^{(t)}_i}\right)$, $m=0,\cdots,M$ in Step 4 of Algorithm~\ref{alg:con-sample}. Hence, Algorithm~\ref{alg:con-sample} can {reduce privacy risk}.
Otherwise, extra privacy mechanisms need to be exploited. Note that FL {for} constrained optimization has not been studied so far, let alone privacy mechanisms for it.

\subsubsection{Algorithm Example}
We provide an example of $\bar{f}_{s,m}$, $m=0,\cdots,M$ with $\bar{f}_{s,0}$ satisfying Assumption~\ref{asump:fbar} and $\bar{f}_{s,m}$ satisfying $\bar{f}_{s,m}(\boldsymbol\omega;\boldsymbol\omega,\mathbf{x})=f_{s,m}(\boldsymbol\omega;\mathbf{x})$ and Assumption~\ref{asump:fbar} for all $m=1,\cdots,M$.
Specifically, we can choose $\bar{f}_{s,0}$ given by \eqref{eqn:fs0bar} and choose $\bar{f}_{s,m}$, $m=1,\cdots,M$ as follows:
%and yields an analytical solution of Problem~\ref{Prob:con-sample}:
\begin{align} &\bar{f}_{s,m}(\boldsymbol\omega,{\boldsymbol\omega}_{s}^{(t)};\mathbf{x}_n)=f_{s,m}({\boldsymbol\omega}_{s}^{(t)};\mathbf{x}_n)\nonumber\\ &\!+\left(\nabla f_{s,m}({\boldsymbol\omega}_{s}^{(t)};\mathbf{x}_n)\right)^T\!\!\left(\boldsymbol\omega-{\boldsymbol\omega}_{s}^{(t)}\right)+\tau\left\Vert{\boldsymbol\omega-{\boldsymbol\omega}_{s}^{(t)}}\right\Vert_2^2, \label{eqn:fsmbar}
\end{align}
where $\tau>0$ can be any constant. {Then, $\sum_{n\in\mathcal{N}^{(t)}_i}\nabla f_{s,0}({\boldsymbol\omega}_{s}^{(t)};\mathbf{x}_n)$ can be viewed as $\mathbf q_{s,0}\left({\boldsymbol\omega}_{s}^{(t)},(\mathbf{x}_n)_{n\in\mathcal{N}^{(t)}_i}\right)$ (implying $D_0=d$), and {$\left(\sum_{n\in\mathcal{N}_i^{(t)}} f_{s,m}({\boldsymbol\omega}_{s}^{(t)};\mathbf{x}_n),\sum_{n\in\mathcal{N}^{(t)}_i}\nabla f_{s,m}({\boldsymbol\omega}_{s}^{(t)};\mathbf{x}_n)\right)$} can be viewed as $\mathbf q_{s,m}\left({\boldsymbol\omega}_{s}^{(t)},(\mathbf{x}_n)_{n\in\mathcal{N}^{(t)}_i}\right)$ (implying $D_m={1+}d$), {for all $m=1,\cdots,M$.}}
Recall that with $f_{s,0}$ given in \eqref{eqn:fs0bar}, $\bar{F}^{(t)}_{s,0}(\boldsymbol\omega)$ is given in \eqref{eqn:Fs0bar_eg}. In addition, for all $m=1,\cdots,M$, substituting \eqref{eqn:fsmbar} into \eqref{eqn:Fsmbar}, $\bar{F}^{(t)}_{s,m}(\boldsymbol\omega)$ can be rewritten as:
\begin{align*}
	\bar{F}^{(t)}_{s,m}(\boldsymbol\omega)
=\hat f^{(t)}_{s,m,0}+\left(\hat{\mathbf f}^{(t)}_{s,m,1}\right)^T\boldsymbol\omega+\tau\left\Vert{\boldsymbol\omega}\right\Vert_2^2,m=1,\cdots,M,%\label{eqn:Fsmbar_eg}
\end{align*}
where $\hat f^{(t)}_{s,m,0}$ and $\hat{\mathbf f}^{(t)}_{s,m,1}\in\mathbb{R}^d$ are given by:
\begin{align}
	&\hat f^{(t)}_{s,m,0}=(1-\rho^{(t)})\hat f^{(t-1)}_{s,m,0}+\rho^{(t)}\sum_{i\in\mathcal{I}}\frac{N_i}{BN}\!\!\!\sum_{n\in\mathcal N_i^{(t)}}\!\!\Big(f_{s,m}({\boldsymbol\omega}_{s}^{(t)};\mathbf{x}_n)\nonumber\\
	&-\left(\nabla f_{s,m}({\boldsymbol\omega}_{s}^{(t)};\mathbf{x}_n)\right)^T\!\!{\boldsymbol\omega}_{s}^{(t)}+\tau\left\Vert{{\boldsymbol\omega}_{s}^{(t)}}\right\Vert_2^2\Big),m=1,\cdots,M,\nonumber\\
	&\hat{\mathbf f}^{(t)}_{s,m,1}=(1-\rho^{(t)})\hat{\mathbf f}^{(t-1)}_{s,m,1}\nonumber\\ &+\rho^{(t)}\sum_{i\in\mathcal{I}}\frac{N_i}{BN}\!\!\sum_{n\in\mathcal N_i^{(t)}}\!\!\Big(\nabla f_{s,m}({\boldsymbol\omega}_{s}^{(t)};\mathbf{x}_n)-2\tau{\boldsymbol\omega}_{s}^{(t)}\Big),m=1,\cdots,M,\nonumber
\end{align}
with $\hat f^{(0)}_{s,m,0}=0$ and $\hat{\mathbf f}^{(0)}_{s,m,1}=\mathbf 0$.
Apparently, Problem~\ref{Prob:con-sample-ap} with $\bar{f}_{s,0}$ given by~\eqref{eqn:fs0bar} and $\bar{f}_{s,m}$, $m=1,\cdots,M$ given by~\eqref{eqn:fsmbar} is a convex quadratically constrained quadratic programming and can be solved using an {interior-point} method.

{Therefore, Step 4 and Step 5 of Algorithm~\ref{alg:con-sample} with $\bar{f}_{s,0}$ given by~\eqref{eqn:fs0bar} and $\bar{f}_{s,m}$, $m=1,\cdots,M$ given by~\eqref{eqn:fsmbar} (i.e., an example of Algorithm~\ref{alg:con-sample})  are given below.}
{In Step 4, each client $i\in \mathcal I$ computes $\sum_{n\in\mathcal{N}^{(t)}_i}\nabla f_{s,0}({\boldsymbol\omega}_{s}^{(t)};\mathbf{x}_n)\in \mathbb R^d$ and $\left(\sum_{n\in\mathcal{N}_i^{(t)}} f_{s,m}({\boldsymbol\omega}_{s}^{(t)};\mathbf{x}_n),\sum_{n\in\mathcal{N}^{(t)}_i}\nabla f_{s,m}({\boldsymbol\omega}_{s}^{(t)};\mathbf{x}_n)\right)\in \mathbb R^{1+d},m=1,\cdots,M$ and sends the $d$-dimensional vector and $M$ $(1+d)$-dimensional vectors to the server.}
{In Step 5, the server calculates $(\bar{\boldsymbol\omega}_{s}^{(t)}, \mathbf s_s^{(t)})$ using an {interior-point} method.}
If for all $i\in\mathcal{I}$, $\mathcal{N}'_i\subseteq\mathcal{N}_i$, {and $\boldsymbol\omega'\in \mathbb R^d$},  the system of equations w.r.t. $(\mathbf{z}_n)_{n=1\cdots,B}$ with $\mathbf{z}_n\in\mathbb{R}^{K}$, i.e., $\sum_{n=1}^B f_{s,m}({\boldsymbol\omega}';\mathbf{z}_n)=\sum_{n\in\mathcal{N}'_i} f_{s,m}({\boldsymbol\omega}';\mathbf{x}_n)$, $m=1,\cdots,M$ and $\sum_{n=1}^B\nabla f_{s,m}({\boldsymbol\omega}';\mathbf{z}_n)=\sum_{n\in\mathcal{N}'_i}\nabla f_{s,m}({\boldsymbol\omega}';\mathbf{x}_n)$, $m=0,\cdots,M$, has an infinite (or a sufficiently large) number of solutions, then {the example of Algorithm~\ref{alg:con-sample}} can {reduce privacy risk}.

\section{Feature-based Federated Learning}\label{sec:feature}
In this section, we
%investigate two types of {feature}-based federated optimization, namely, unconstrained feature-based federated optimization and constrained feature-based federated optimization.
%For each optimization problem, we propose an FL algorithm using mini-batch SSCA.
{propose FL algorithms for unconstrained and constrained feature-based federated optimization problems, respectively, using mini-batch SSCA. In feature-based FL, the batch size $B$ satisfies $B\leq N$.}

\subsection{Feature-based Federated Learning for Unconstrained Optimization}\label{subsec:uncon-feature}
In this part, we consider the following unconstrained feature-based federated optimization problem:
\begin{Prob}[Unconstrained Feature-based Federated Optimization]\label{Prob:uncon-feature}
	\begin{align}
		&\min_{\boldsymbol\omega} \quad F_{f,0}(\boldsymbol\omega)\nonumber
	\end{align}
	where $F_{f,0}(\boldsymbol\omega)$ is given by~\eqref{eqn:Ff0}.
\end{Prob}

{In \cite{hardy2017private}, SGD is utilized to obtain a  {stationary} point of Problem~\ref{Prob:uncon-feature} only with $I=2$ and $F_{f,0}(\boldsymbol\omega)$ being the cross-entropy function.
In the following, we propose a  feature-based FL algorithm, i.e., Algorithm~\ref{alg:uncon-feature}, to obtain a {stationary} point of Problem~\ref{Prob:uncon-feature}  using mini-batch SSCA, which empirically achieves a higher convergence speed than SGD.}

\subsubsection{Algorithm Description}
\begin{algorithm}[t]
	\caption{Mini-batch SSCA for Problem~\ref{Prob:uncon-feature}}
	\begin{small}
		\begin{algorithmic}[1]
			%           \STATE \textbf{Input}: $\{\gamma^k\}$
			\STATE \textbf{initialize}: choose any ${\boldsymbol\omega}_{f}^{1}$ at the server.\\
			\FOR{$t=1,2,\cdots,T-1$}
			\STATE the server randomly selects a mini-batch with the index set denoted by $\mathcal{N}^{(t)}\subset\mathcal{N}$ and sends $\mathcal{N}^{(t)}$ and $({\boldsymbol\omega}_0^{(t)},{\boldsymbol\omega}_i^{(t)})$ to client $i$ for all $i\in\mathcal{I}$.
			\STATE for all $i\in\mathcal{I}$, client $i$ computes $\mathbf{h}_{0,i}({\boldsymbol\omega}^{(t)}_i,\mathbf{x}_{n,i})$, $n\in\mathcal{N}^{(t)}$ and sends them to the other clients.
						\STATE the client with the highest computation speed (or any client) computes $\mathbf q_{f,0,0}\left({\boldsymbol\omega}^{(t)}_0,\left(\mathbf{h}_{0,i}({\boldsymbol\omega}^{(t)}_i,\mathbf{x}_{n,i})\right)_{n\in\mathcal{N}^{(t)},i\in\mathcal{I}}\right)$ and sends it to the server.
			\STATE for all $i\in\mathcal{I}$, client $i$ computes $\mathbf q_{f,0,i}\!\!\left(\!{\boldsymbol\omega}^{(t)}_0\!,{\boldsymbol\omega}^{(t)}_i\!,\!{(\mathbf{x}_{n,i})_{n\in\mathcal{N}^{(t)}}},\!\left(\mathbf{h}_{0,j}({\boldsymbol\omega}^{(t)}_j,\mathbf{x}_{n,j})\!\right)_{n\in\mathcal{N}^{(t)},j\in\mathcal{I}}\right)$ and sends it to the server.
			\STATE the server obtains $\bar{\boldsymbol\omega}_{f}^{(t)}$ by solving Problem~\ref{Prob:uncon-feature-ap}.
			\STATE the server updates ${\boldsymbol\omega}_{f}^{(t+1)}$ according to \eqref{eqn:updatew-f}.
			\ENDFOR
			\STATE \textbf{Output}: ${\boldsymbol\omega}_{f}^T$
		\end{algorithmic}\label{alg:uncon-feature}
	\end{small}
\end{algorithm}

At iteration $t$, we choose:
\begin{align}
	\bar F^{(t)}_{f,0}(\boldsymbol\omega)\!=\!(1-\rho^{(t)})\bar{F}^{(t-1)}_{f,0}(\boldsymbol\omega)\!+\!\rho^{(t)}\frac{1}{B}\!\!\sum_{n\in\mathcal N^{(t)}}\!\!\bar f_{f,0}(\boldsymbol\omega;{\boldsymbol\omega}_{f}^{(t)}\!,\mathbf{x}_n)\label{eqn:Ff0bar}
\end{align}
with $\bar F_{f,0}^{(0)}(\boldsymbol\omega)=0$ as {a convex} approximation function of $F_{f,0}(\boldsymbol\omega)$,
where $\rho^{(t)}$ is a stepsize satisfying~\eqref{eqn:rho},
$\mathcal N^{(t)}\in\mathcal{N}$ is a randomly selected mini-batch by the server at iteration $t$,   and
$\bar{f}_{f,0}(\boldsymbol\omega;{\boldsymbol\omega}_{f}^{(t)},\mathbf{x}_n)$ is a convex approximation of $f_{f,0}(\boldsymbol\omega;\mathbf{x}_n)$ around ${\boldsymbol\omega}_{f}^{(t)}$ satisfying Assumption~\ref{asump:fbar}.
%A common example of $\bar{f}_{f,0}$ will be given later.

\begin{figure*}
	\begin{align} &\sum_{n\in\mathcal{N}'}\!\!\bar{f}_{f,0}(\boldsymbol\omega;\boldsymbol\omega',\mathbf{x}_n)
\!=\!p_{f,0}\bigg(\!\boldsymbol\omega,
		\mathbf q_{f,0,0}\!\left(\!\boldsymbol\omega'_0,\left(\mathbf{h}_{0,i}(\boldsymbol\omega'_i,\mathbf{x}_{n,i})\right)_{n\in\mathcal{N}',i\in\mathcal{I}}\!\right)\!\!,
\!\left(\!\mathbf q_{f,0,i}\!\left(\!\boldsymbol\omega'_0,\boldsymbol\omega'_i, \!{(\mathbf{x}_{n,i})_{n\in\mathcal{N}'}},\!\left(\mathbf{h}_{0,j}(\boldsymbol\omega'_j,\mathbf{x}_{n,j})\!\right)_{n\in\mathcal{N}',j\in\mathcal{I}}\!\right)\!\!\right)_{i\in\mathcal{I}}\!\!\bigg)\label{eqn:ff0bar-composition}
	\end{align}
	\normalsize \hrulefill
\end{figure*}
{Suppose} that for any mini-batch $\mathcal{N}'\subseteq\mathcal{N}$ with batch size $B$, $\sum_{n\in\mathcal{N}'}\bar{f}_{f,0}(\boldsymbol\omega;\boldsymbol\omega',\mathbf{x}_n)$, {a function of  $\boldsymbol\omega$ with parameters jointly determined by $\boldsymbol\omega'$ and $\mathbf{x}_n,n\in\mathcal{N}'_i$,}  can be written as~\eqref{eqn:ff0bar-composition}, as shown at the top of {the next} page,
with $p_{f,0}:\mathbb{R}^{d+{\sum_{i=1}^I E_{0,i}}}\to\mathbb{R}$, {$\mathbf q_{f,0,0}:\mathbb{R}^{d_0+H_0BI}\to\mathbb{R}^{E_{0,0}}$,}
 and $\mathbf q_{f,0,i}:\mathbb{R}^{d_0+d_i+{K_iB+H_0BI}}\to\mathbb{R}^{{E_{0,i}}}$, $i\in\mathcal{I}$, {for some positive integers $E_i,i=0,1,\cdots,I$.}\footnote{{{This} assumption is met by commonly used loss functions such as those in ~\cite{yang2019federated,hardy2017private,yang2019parallel} and Section~\ref{sec:application}.}}
Assume that the expressions of $\bar{f}_{f,0}$, $p_{f,0}$, {$\mathbf q_{f,0,0}$}, $\mathbf q_{f,0,i},i\in\mathcal{I}$, and $\mathbf{h}_{0,i},i\in\mathcal{I}$ are known to the server and {$I$} clients.
Each client $i\in\mathcal{I}$ computes $\mathbf{h}_{0,i}({\boldsymbol\omega}^{(t)}_i,\mathbf{x}_{n,i})$, $n\in\mathcal{N}^{(t)}$ and sends them to the other clients.
{The client with the highest computation speed (or any client) computes $\mathbf q_{f,0,0}\left({\boldsymbol\omega}^{(t)}_0,\left(\mathbf{h}_{0,i}({\boldsymbol\omega}^{(t)}_i,\mathbf{x}_{n,i})\right)_{n\in\mathcal{N}^{(t)},i\in\mathcal{I}}\right)$ based on $\mathbf{h}_{0,i}({\boldsymbol\omega}^{(t)}_i,\mathbf{x}_{n,i})$, $i\in\mathcal{I}$, $n\in\mathcal{N}^{(t)}$ and sends it to the server.
Moreover, each client $i\in\mathcal{I}$ computes
$\mathbf q_{f,0,i}\!\left(\!{\boldsymbol\omega}^{(t)}_0\!,{\boldsymbol\omega}^{(t)}_i\!,{(\mathbf{x}_{n,i})_{n\in\mathcal{N}^{(t)}}}, \left(\mathbf{h}_{0,j}({\boldsymbol\omega}^{(t)}_j,\mathbf{x}_{n,j})\right)_{n\in\mathcal{N}^{(t)},j\in\mathcal{I}}\right)$ and sends it to the server.}\footnote{{The information collection mechanism in Algorithm~\ref{alg:uncon-feature} can be viewed as an extension of that in the feature-based FL algorithm via SGD~\cite{hardy2017private}.}}
Then, the server solves the following convex approximate problem to obtain $\bar{\boldsymbol\omega}_{f}^{(t)}$.
\begin{Prob}[Convex Approximate Problem of Problem~\ref{Prob:uncon-feature}]\label{Prob:uncon-feature-ap}
	\begin{align} &\bar{\boldsymbol\omega}_{f}^{(t)}\triangleq
\mathop{\arg\min}_{\boldsymbol\omega}
\ \bar F_{f,0}^{(t)}(\boldsymbol\omega)\nonumber
	\end{align}
\end{Prob}

{Like} Problem~\ref{Prob:uncon-sample-ap},  Problem~\ref{Prob:uncon-feature-ap} is {an unconstrained convex problem and can be readily solved.}
Given $\bar{\boldsymbol\omega}_{f}^{(t)}$, the server updates ${\boldsymbol\omega}_{f}^{(t)}$ according to:
{\begin{align} &{\boldsymbol\omega}_{f}^{(t+1)}=(1-\gamma^{(t)}){\boldsymbol\omega}_{f}^{(t)}+\gamma^{(t)}\bar{\boldsymbol\omega}_{f}^{(t)},\ t=1,2,\cdots \label{eqn:updatew-f}
\end{align}
where $\gamma^{(t)}$ is a stepsize satisfying \eqref{eqn:gamma}.} The detailed procedure is summarized in Algorithm~\ref{alg:uncon-feature}. The convergence of Algorithm~\ref{alg:uncon-feature} is summarized below.
\begin{Thm}[Convergence of Algorithm~\ref{alg:uncon-feature}]\label{thm:uncon-feature}
	Suppose that $f_{f,0}$ satisfies Assumption~\ref{asump:f}, $\bar f_{f,0}$ satisfies Assumption~\ref{asump:fbar}, and the sequence $\{{\boldsymbol\omega}_{f}^{(t)}\}$ generated by Algorithm~\ref{alg:uncon-feature} is bounded almost surely. Then, every limit point of $\{{\boldsymbol\omega}_{f}^{(t)}\}$ is a {stationary} point of Problem~\ref{Prob:uncon-feature} almost surely.
\end{Thm}

\begin{IEEEproof}{Please refer to Appendix A.}
\end{IEEEproof}

\subsubsection{Security Analysis}
%We establish the security of Algorithm~\ref{alg:uncon-feature}.
Suppose 1) for all $i\in\mathcal{I}$, mini-batch $\mathcal{N}'\subseteq\mathcal{N}$, {and $\boldsymbol\omega_i'\in \mathbb R^{d_i}$},  the system of equations w.r.t. {$(\boldsymbol\theta,(\mathbf{z}_{n})_{n=1,\cdots,B})\in \mathbb R^{d_i+BK_i}$ with $\boldsymbol\theta\in\mathbb R^{d_i}$ and $\mathbf{z}_{n}\in\mathbb{R}^{K_i}$, i.e., $\mathbf{h}_{0,i}(\boldsymbol\theta,\mathbf{z}_{n})=\mathbf{h}_{0,i}(\boldsymbol\omega'_i,\mathbf{x}_{n,i})$}, $n\in\mathcal{N}'$, has an infinite (or a sufficiently large) number of solutions;
2) for any mini-batch $\mathcal{N}'\subseteq\mathcal{N}$ {and $\boldsymbol\omega'\in \mathbb R^d$}, the system of equations w.r.t. $\left(\mathbf{z}_{n,i}\right)_{n=1,\cdots,B,i\in\mathcal{I}}\in \mathbb R^{BK}$ with $\mathbf{z}_{n,i}\in\mathbb{R}^{K_i}$, i.e.,
$\mathbf q_{f,0,0}\left(\boldsymbol\omega'_0,\left(\mathbf{h}_{0,i}(\boldsymbol\omega'_i,\mathbf{z}_{n,i})\right)_{n=1,\cdots,B,i\in\mathcal{I}}\right)=\mathbf q_{f,0,0}\left(\boldsymbol\omega'_0,\left(\mathbf{h}_{0,i}(\boldsymbol\omega'_i,\mathbf{x}_{n,i})\right)_{n\in\mathcal{N}',i\in\mathcal{I}}\right)$,
$\mathbf q_{f,0,i}\left(\boldsymbol\omega'_0,\boldsymbol\omega'_i, {(\mathbf{z}_{n,i})_{n\in\mathcal{N}'}},\left(\mathbf{h}_{0,j}(\boldsymbol\omega'_j,\mathbf{z}_{n,j})\right)_{n=1,\cdots,B,j\in\mathcal{I}}\right)=\mathbf q_{f,0,i}\left(\boldsymbol\omega'_0,\boldsymbol\omega'_i, {(\mathbf{x}_{n,i})_{n\in\mathcal{N}'}}, \left(\mathbf{h}_{0,j}(\boldsymbol\omega'_j,\mathbf{x}_{n,j})\right)_{n\in\mathcal{N}',j\in\mathcal{I}}\right)$, $i\in\mathcal{I}$, has an infinite (or a sufficiently large) number of solutions. In {that} case, raw data $\mathbf{x}_n$, $n\in\mathcal{N}^{(t)}$ {can hardly be} extracted by {any} client from $\mathbf{h}_{0,i}({\boldsymbol\omega}^{(t)}_i,\mathbf{x}_{n,i})$, $n\in\mathcal{N}^{(t)}$, $i\in\mathcal{I}$ or by the server from $\mathbf q_{f,0,0}\left({\boldsymbol\omega}^{(t)}_0,\left(\mathbf{h}_{0,i}({\boldsymbol\omega}^{(t)}_i,\mathbf{x}_{n,i})\right)_{n\in\mathcal{N}^{(t)},i\in\mathcal{I}}\right)$,
$\mathbf q_{f,0,i}\!\left(\!{\boldsymbol\omega}^{(t)}_0\!,{\boldsymbol\omega}^{(t)}_i\!,{(\mathbf{x}_{n,i})_{n\in\mathcal{N}^{(t)}}}, \left(\mathbf{h}_{0,j}({\boldsymbol\omega}^{(t)}_j,\mathbf{x}_{n,j})\right)_{n\in\mathcal{N}^{(t)},j\in\mathcal{I}}\!\right)$, $i\in\mathcal{I}$  in Steps 4-6 of Algorithm~\ref{alg:uncon-feature}, and hence Algorithm~\ref{alg:uncon-feature} can {reduce privacy risk}.
%\footnote{{For the extension of the feature-based FL algorithm via SGD without extra privacy mechanisms~\cite{hardy2017private} to the general case with  $I>2$ and $F_{f,0}(\boldsymbol\omega)$ given in \eqref{eqn:Ff0}, the corresponding  systems of equations w.r.t.
%{$(\boldsymbol\theta,(\mathbf{z}_{n})_{n=1,\cdots,B})\in \mathbb R^{d_i+BK_i}$} and $(\mathbf{z}_{n})_{n=1,\cdots,B}\in\mathbb R^{
%BK}$ are   {$\mathbf{h}_{0,i}(\boldsymbol\theta,\mathbf{z}_{n})=\mathbf{h}_{0,i}(\boldsymbol\omega'_i,\mathbf{x}_{n,i})$}, $n\in\mathcal{N}'$ and $\sum_{n=1}^B\nabla_{\boldsymbol\omega} f_{f,0}({\boldsymbol\omega}';\mathbf{z}_{n})=\sum_{n\in\mathcal{N}'}\nabla_{\boldsymbol\omega} f_{f,0}({\boldsymbol\omega}';\mathbf{x}_n)$, respectively.}\label{ft:4}}
{However,} if the two assumptions mentioned above are not satisfied, extra privacy mechanisms are required. For instance, if $\bar{\boldsymbol\omega}_{f}^{(t)}$ is linear in $q_{f,0,0}$ and $q_{f,0,i}$, $i\in\mathcal{I}$, then homomorphic encryption~\cite{hardy2017private} can be applied.

\subsubsection{Algorithm Example}
We provide an example of $\bar{f}_{f,0}$ which satisfies Assumption~\ref{asump:fbar} and yields an analytical solution of Problem~\ref{Prob:uncon-feature-ap}:
\begin{align}
	\bar f_{f\!,0}(\boldsymbol\omega;{\boldsymbol\omega}_{f}^{(t)}\!,\mathbf{x}_n)\!=&\!\left(\nabla_{\boldsymbol\omega} f_{f\!,0}({\boldsymbol\omega}_{f}^{(t)}\!;\mathbf{x}_n)\!\right)^T\!\!\left(\boldsymbol\omega\!-\!{\boldsymbol\omega}_{f}^{(t)}\!\right)\nonumber\\ &+\tau\left\Vert{\boldsymbol\omega-{\boldsymbol\omega}_{f}^{(t)}}\right\Vert_2^2,\ n\in\mathcal N^{(t)}, \label{eqn:ff0bar}
\end{align}
where $\tau>0$ can be any constant. By the chain rule, we have:
\begin{align}
&\nabla_{\boldsymbol\omega_0}f_{f,0}({\boldsymbol\omega}_{f}^{(t)};\mathbf{x}_n)=\nabla_{\boldsymbol\omega_0} g_0\!\left(\boldsymbol\omega_0^{(t)},\left(\mathbf{h}_{0,i}({\boldsymbol\omega}^{(t)}_i,\mathbf{x}_{n,i})\right)_{i\in\mathcal{I}}\right),\nonumber\\
&\hspace{160pt} n\in\mathcal N^{(t)}, \label{eqn:chainrule1}\\
&\nabla_{\boldsymbol\omega_i}f_{f,0}({\boldsymbol\omega}_{f}^{(t)};\mathbf{x}_n)\nonumber\\
=&\nabla_{\mathbf{h}_{0,i}} g_0\!\left(\boldsymbol\omega_0^{(t)},\left(\mathbf{h}_{0,i}({\boldsymbol\omega}^{(t)}_i,\mathbf{x}_{n,i})\right)_{i\in\mathcal{I}}\right)^{T}\!\frac{\partial\mathbf{h}_{0,i}({\boldsymbol\omega}^{(t)}_i,\mathbf{x}_{n,i})}{\partial\boldsymbol\omega_i},\nonumber\\
&\hspace{140pt}\ n\in\mathcal N^{(t)}, i\in\mathcal{I}. \label{eqn:chainrule2}
\end{align}
Substituting \eqref{eqn:chainrule1} and \eqref{eqn:chainrule2} into \eqref{eqn:ff0bar}, we know that $\sum_{n\in\mathcal{N}^{(t)}}\nabla_{\boldsymbol\omega_0} f_{f,0}({\boldsymbol\omega}_{f}^{(t)};\mathbf{x}_n)$ and $\sum_{n\in\mathcal{N}^{(t)}}\nabla_{\boldsymbol\omega_i} f_{f,0}({\boldsymbol\omega}_{f}^{(t)};\mathbf{x}_n),i\in\mathcal I$
can be viewed as $\mathbf q_{f,0,0}\left({\boldsymbol\omega}^{(t)}_0,\left(\mathbf{h}_{0,i}({\boldsymbol\omega}^{(t)}_i,\mathbf{x}_{n,i})\right)_{n\in\mathcal{N}^{(t)},i\in\mathcal{I}}\right)$ {(implying $E_{0,0}=d_0$)} and $\mathbf q_{f,0,i}\!\left(\!{\boldsymbol\omega}^{(t)}_0\!,{\boldsymbol\omega}^{(t)}_i\!, {(\mathbf{x}_{n,i})_{n\in\mathcal{N}^{(t)}}}, \left(\mathbf{h}_{0,j}({\boldsymbol\omega}^{(t)}_j,\mathbf{x}_{n,j})\right)_{n\in\mathcal{N}^{(t)},j\in\mathcal{I}}\right)$ {(implying $E_{0,i}=d_i$), $i\in\mathcal I$}, respectively. Besides, substituting \eqref{eqn:ff0bar} into \eqref{eqn:Ff0bar}, $\bar{F}^{(t)}_{f,0}(\boldsymbol\omega)$ can be rewritten as:
\begin{align}
	\bar{F}^{(t)}_{f,0}(\boldsymbol\omega)=&\left(\hat{\mathbf f}^{(t)}_{f,0,1}\right)^T\boldsymbol\omega+\tau\left\Vert{\boldsymbol\omega}\right\Vert_2^2,\label{eqn:Ff0bar_eg}
\end{align}
where $\hat{\mathbf f}^{(t)}_{f,0,1}\in\mathbb{R}^d$ is given by:
\begin{align}
\!\!\!\!\!\!\hat{\mathbf f}^{(t)}_{f,0,1}\!\!=\!(1\!-\!\rho^{(t)}\!)\hat{\mathbf f}^{(t-1)}_{f,0,1}\!\!+\!\frac{\rho^{(t)}}{B}\!\!\!\!\!\sum_{n\in\mathcal N^{(t)}}\!\!\!\!\!\Big(\!\nabla f_{f,0}({\boldsymbol\omega}_{f}^{(t)};\mathbf{x}_n)\!-\!2\tau{\boldsymbol\omega}_{f}^{(t)}\!\Big)\!\!\label{eqn:feature-accu-hat-f}
\end{align}
with $\hat{\mathbf f}^{(0)}_{f,0,1}=\mathbf 0$.
{Similar} to Problem~\ref{Prob:uncon-sample-ap} with $\bar{f}_{s,0}$ given by~\eqref{eqn:fs0bar}, Problem~\ref{Prob:uncon-feature-ap} with $\bar{f}_{f,0}$ given by~\eqref{eqn:ff0bar} is an unconstrained convex quadratic programming w.r.t. $\boldsymbol\omega$ and hence has the following analytical solution:
\begin{align}
	\bar{\boldsymbol\omega}_{f}^{(t)}=-\frac{1}{2\tau}\hat{\mathbf f}^{(t)}_{f,0,1}.\label{eqn:omegafbar}
\end{align}
%The computational complexity for calculating $\bar{\boldsymbol\omega}_{f}^{(t)}$ in~\eqref{eqn:omegafbar} is $\mathcal O(d)$.

{Therefore, Steps 5-7 of Algorithm~\ref{alg:uncon-feature} with $\bar{f}_{f,0}$ given by~\eqref{eqn:ff0bar} (i.e., an example of Algorithm~\ref{alg:uncon-feature}) are given below.}
{In Step 5, the client with the highest computation speed (or any client) computes $\sum_{n\in\mathcal{N}^{(t)}}\nabla_{\boldsymbol\omega_0} f_{f,0}({\boldsymbol\omega}_{f}^{(t)};\mathbf{x}_n)\in \mathbb R^{d_0}$ and sends the $d_0$-dimensional vector to the server.  In Step 6, each client $i\in \mathcal I$ computes $\sum_{n\in\mathcal{N}^{(t)}}\nabla_{\boldsymbol\omega_i} f_{f,0}({\boldsymbol\omega}_{f}^{(t)};\mathbf{x}_n)\in \mathbb R^{d_i}$ and sends the $d_i$-dimensional vector to the server.
In Step 7, the server calculates $\bar{\boldsymbol\omega}_{f}^{(t)}$ according to \eqref{eqn:omegafbar}.}
Suppose that for all $i\in\mathcal{I}$, mini-batch $\mathcal{N}'\subseteq\mathcal{N}$, {and $\boldsymbol\omega_i'\in \mathbb R^{d_i}$},  the system of equations w.r.t.  {$(\boldsymbol\theta,(\mathbf{z}_{n})_{n=1,\cdots,B})\in \mathbb R^{d_i+BK_i}$ with $\boldsymbol\theta\in\mathbb R^{d_i}$ and $\mathbf{z}_{n}\in\mathbb{R}^{K_i}$, i.e., $\mathbf{h}_{0,i}(\boldsymbol\theta,\mathbf{z}_{n})=\mathbf{h}_{0,i}(\boldsymbol\omega'_i,\mathbf{x}_{n,i})$}, $n\in\mathcal{N}'$, has an infinite (or a sufficiently large) number of solutions, and for any $\mathcal{N}'\subseteq\mathcal{N}$ {and $\boldsymbol\omega'\in \mathbb R^{d}$}, the system of equations w.r.t. $(\mathbf{z}_n)_{n=1\cdots,B}$ with $\mathbf{z}_n\in\mathbb{R}^{K}$, i.e., $\sum_{n=1}^B\nabla f_{f,0}(\boldsymbol\omega';\mathbf{z}_n)=\sum_{n\in\mathcal{N}'}\nabla f_{f,0}(\boldsymbol\omega';\mathbf{x}_n)$, has an infinite (or a sufficiently large) number of solutions. In {that} case, {the example of Algorithm~\ref{alg:uncon-feature}} can {reduce privacy risk}.
%\footnote{The condition for Algorithm~\ref{alg:uncon-feature} with $\bar{f}_{f,0}$ given by~\eqref{eqn:ff0bar} to {reduce privacy risk} without extra privacy mechanisms is the same as that for the existing feature-based FL algorithm via SGD {without extra privacy mechanisms} \cite{hardy2017private}, as illustrated in Footnote~\ref{ft:4}. }
If the two assumptions are not satisfied, homomorphic encryption~\cite{HE,hardy2017private} can be applied to preserve data privacy, since $\bar{\boldsymbol\omega}_{f}^{(t)}$ is linear in $\nabla f_{f,0}({\boldsymbol\omega}_{f}^{(t)};\mathbf{x}_n)$, $n\in\mathcal{N}$, as shown in~{\eqref{eqn:feature-accu-hat-f} and}~\eqref{eqn:omegafbar}. {Similarly, {the example of Algorithm~\ref{alg:uncon-feature}}  can be viewed as a feature-based FL algorithm  via momentum SGD with  diminishing stepsize $\gamma^{(t)}$.}

\begin{Rem}[Comparison Between Example of Algorithm~\ref{alg:uncon-feature}  and Feature-based FL Algorithm via SGD\cite{hardy2017private}]
%	Notice that under the feature-based FL algorithm for $I>2$ via SGD~\cite{chen2020vafl}, each client conducts the same operation as in the above example, and the server conducts an SGD update. Thus, under
	 Algorithm~\ref{alg:uncon-feature} with $\bar{f}_{f,0}$ given by~\eqref{eqn:ff0bar} and the extension of  the feature-based FL algorithm via SGD~\cite{hardy2017private}  (without extra privacy mechanisms)  to the general case with  $I>2$ and $F_{f,0}(\boldsymbol\omega)$ given in \eqref{eqn:Ff0} {have the same order of computational complexity ($\mathcal O(B)$)  and  communication load per communication round and the same level of privacy protection (due to the same system of equations for {inferring} private data).}
\end{Rem}
{\begin{Rem}[Information Collection for Example of Algorithm~\ref{alg:uncon-feature}] When choosing $\bar{f}_{f,0}$ given by~\eqref{eqn:ff0bar}, another option for collecting information  is to let each client $i\in \mathcal I$  directly send $\mathbf{h}_{0,i}({\boldsymbol\omega}^{(t)}_i,\mathbf{x}_{n,i})$, $\nabla_{\boldsymbol\omega_i}\mathbf{h}_{0,i}({\boldsymbol\omega}^{(t)}_i,\mathbf{x}_{n,i})$, $n\in\mathcal{N}^{(t)}$  to the sever.\footnote{{This {one-step} information collection  mechanism can be viewed as an extension of that  in the feature-based FL algorithm via SGD \cite{chen2020vafl} to the case where the server maintains the global model.}} In general, it has a lower communication load but higher privacy risk  than the  information collection  mechanism in Steps 4-6
 of {the example of Algorithm~\ref{alg:uncon-feature}}, without using additional  privacy {mechanisms}.\footnote{{For the loss function given in \eqref{eqn:Fcost}, the one-step information collection mechanism exposes raw data (as $\nabla_{\boldsymbol\omega_i}\mathbf{h}_{0,i}({\boldsymbol\omega}^{(t)}_i,\mathbf{x}_{n,i})=\mathbf{x}_{n,i}$), whereas the one adopted in Algorithm~\ref{alg:uncon-feature} does not, as shown  in Section~\ref{sec:application}.}}
\end{Rem}}

\subsection{Feature-based Federated Learning for Constrained Optimization}
In this part, we consider the following constrained feature-based federated optimization problem:
\begin{Prob}[Constrained Feature-based Federated Optimization]\label{Prob:con-feature}
	\begin{align}
		\min_{\boldsymbol\omega}\quad&F_{f,0}(\boldsymbol\omega)\nonumber\\
		\text{s.t.}\quad&F_{f,m}(\boldsymbol\omega)\leq 0,\ m=1,2,\cdots,M,\nonumber
	\end{align}
	where $F_{f,0}(\boldsymbol\omega)$ is given by~\eqref{eqn:Ff0},
	and
	\begin{align} F_{f,m}(\boldsymbol\omega)\triangleq\frac{1}{N}\sum_{n\in\mathcal{N}} \underbrace{g_m\left(\boldsymbol\omega_0,\left(\mathbf{h}_{m,i}(\boldsymbol\omega_i,\mathbf{x}_{n,i})\right)_{i\in\mathcal{I}}\right)}_{\triangleq f_{f,m}(\boldsymbol\omega;\mathbf{x}_n)}.\nonumber
	\end{align}
	{Here, $f_{f,m}(\boldsymbol\omega;\mathbf{x}_n)$ is formed by composing  $g_m:\mathbb{R}^{d_0+H_mI}\to\mathbb{R}$ with  functions $\mathbf{h}_{m,i}:\mathbb R^{d_i+K_i}\to\mathbb{R}^{H_m},i\in\mathcal{I}$, for some positive integer $H_m$.}
\end{Prob}

To be general, $F_{f,m}(\boldsymbol\omega)$, $m=0,\cdots,M$ are not assumed to be convex in $\boldsymbol\omega$.
Analogously to Algorithm~\ref{alg:con-sample}, we propose a  feature-based FL algorithm, i.e., Algorithm~\ref{alg:con-feature}, to obtain a KKT point of Problem~\ref{Prob:con-feature}, by combining the exact penalty method for SSCA in our previous work~\cite{Ye} and mini-batch techniques.

\subsubsection{Algorithm Description}
\begin{algorithm}[t]
	\caption{Mini-batch SSCA for Problem~\ref{Prob:con-feature}}
		\begin{small}
	\begin{algorithmic}[1]
		%           \STATE \textbf{Input}: $\{\gamma^k\}$
		\STATE \textbf{initialize}: choose any ${\boldsymbol\omega}_{f}^{1}$ and $c>0$ at the server.\\
		\FOR{$t=1,2,\cdots,T-1$}
		\STATE the server randomly selects a mini-batch with the index set denoted by $\mathcal{N}^{(t)}\subset\mathcal{N}$ and sends $\mathcal{N}^{(t)}$ and $({\boldsymbol\omega}_0^{(t)},{\boldsymbol\omega}_i^{(t)})$ to client $i$ for all $i\in\mathcal{I}$.
		\STATE for all $i\in\mathcal{I}$, client $i$ computes $\mathbf{h}_{m,i}({\boldsymbol\omega}^{(t)}_i,\mathbf{x}_{n,i})$, $n\in\mathcal{N}^{(t)}$, $m=0,1,\cdots,M$ and sends them to the other clients.
		\STATE the client with the highest computation speed (or any client) computes $\mathbf q_{f,m,0}\left({\boldsymbol\omega}^{(t)}_0,\left(\mathbf{h}_{m,i}({\boldsymbol\omega}^{(t)}_i,\mathbf{x}_{n,i})\right)_{n\in\mathcal{N}^{(t)},i\in\mathcal{I}}\right)$, $m=0,1,\cdots,M$ and sends them to the server.
		\STATE for all $i\in\mathcal{I}$, client $i$ computes $\mathbf q_{f,m,i}\!\left(\!\!{\boldsymbol\omega}^{(t)}_0\!,\!{\boldsymbol\omega}^{(t)}_i\!,\!{(\mathbf{x}_{n,i})_{\!n\in\mathcal{N}^{(t)}}},\!\left(\!\mathbf{h}_{m,j}({\boldsymbol\omega}^{(t)}_j,\mathbf{x}_{n,j})\!\right)_{\! n\in\mathcal{N}^{(t)}\!,j\in\mathcal{I}}\!\right)$, $m=0,1,\cdots,M$ and sends them to the server.
				\STATE the server obtains $(\bar{\boldsymbol\omega}_{f}^{(t)}, \mathbf s_f^{(t)})$ by solving Problem~\ref{Prob:con-feature-ap}.
		\STATE the server  updates ${\boldsymbol\omega}_{f}^{(t+1)}$ according to~\eqref{eqn:updatew-f}.
		\ENDFOR
		\STATE \textbf{Output}: ${\boldsymbol\omega}_{f}^T$
	\end{algorithmic}\label{alg:con-feature}
		\end{small}
\end{algorithm}

{Similarly, to ensure feasible stochastic iterates, we first} transform Problem~\ref{Prob:con-feature} to the following stochastic optimization problem with a slack penalty term.
\begin{Prob}[Transformed Problem of Problem~\ref{Prob:con-feature}]\label{Prob:con-feature-ep}
	\begin{align}
		\min_{\boldsymbol\omega,\mathbf s}\quad &F_{f,0}\left(\boldsymbol\omega\right)+c\sum_{m=1}^M s_m\nonumber\\
		\text{s.t.}\quad&F_{f,m}\left(\boldsymbol\omega\right)\leq s_m,\ m=1,2,\cdots,M,\nonumber\\
		&s_m\geq 0,\ m=1,2,\cdots,M.\nonumber
	\end{align}
\end{Prob}

At iteration $t$, we choose $\bar F^{(t)}_{f,0}(\boldsymbol\omega)$ given in~\eqref{eqn:Ff0bar} as an approximation function of $F_{f,0}(\boldsymbol\omega)$ and choose:
\begin{align}
&\bar F^{(t)}_{f,m}(\boldsymbol\omega)\!=\!(1\!-\!\rho^{(t)})\bar{F}^{(t-1)}_{f,m}(\boldsymbol\omega)\!+\!\rho^{(t)}\frac{1}{B}\!\!\!\sum_{n\in\mathcal N^{(t)}}\!\!\!\!\bar f_{f,m}(\boldsymbol\omega;{\boldsymbol\omega}_{f}^{(t)},\mathbf{x}_n),\nonumber\\
&\hspace{5cm} m=1,\cdots,M\label{eqn:Ffmbar}
\end{align}
with $\bar F_{f,m}^{(0)}(\boldsymbol\omega)=0$ as {a convex} approximation function of $F_{f,m}(\boldsymbol\omega)$, for all $m=1,\cdots,M$, where $\rho^{(t)}$ is a stepsize satisfying~\eqref{eqn:rho},
$\mathcal N^{(t)}$ is a randomly selected mini-batch by the server at iteration $t$,
and $\bar f_{f,m}(\boldsymbol\omega;{\boldsymbol\omega}_{f}^{(t)},\mathbf{x}_n)$ is a convex approximation of $f_{f,m}(\boldsymbol\omega;\mathbf{x}_n)$ around ${\boldsymbol\omega}_{f}^{(t)}$ satisfying $\bar{f}_{s,m}(\boldsymbol\omega;\boldsymbol\omega,\mathbf{x})=f_{s,m}(\boldsymbol\omega;\mathbf{x})$ and Assumption~\ref{asump:fbar} for all $m=1\cdots,M$.
%A common example of $\bar{f}_{f,m}$, $m=1\cdots,M$ will be given later.

\begin{figure*}
	\begin{align} &\sum_{n\in\mathcal{N}'}\!\!\!\bar{f}_{f,m}\!(\boldsymbol\omega;\boldsymbol\omega'\!,\!\mathbf{x}_n)
\!\!=\!p_{f,m}
\!\bigg(\!\!\boldsymbol\omega,\!\mathbf q_{f,m,0}\!\left(\boldsymbol\omega'_0,\!\left(\mathbf{h}_{m,i}(\boldsymbol\omega'_i,\mathbf{x}_{n,i})\!\right)\!_{n\in\mathcal{N}',i\in\mathcal{I}}\right)\!,
		\!\left(\!\mathbf q_{f,m,i}\!\!\left(\!\boldsymbol\omega'_0,\boldsymbol\omega'_i,\!{(\mathbf{x}_{n,i})_{n\in\mathcal{N}'}}\!, \!\left(\mathbf{h}_{m,j}(\boldsymbol\omega'_j,\mathbf{x}_{n,j})\!\right)_{n\in\mathcal{N}',j\in\mathcal{I}}\!\right)\!\!\right)\!_{i\in\mathcal{I}}\!\!\bigg)\!,\nonumber\\
		&\hspace{14cm} m=0,\cdots,M\label{eqn:ffmbar-composition}
	\end{align}
	\normalsize \hrulefill
\end{figure*}
Note that for any mini-batch $\mathcal{N}'\subseteq\mathcal{N}$ with batch size $B$, $\sum_{n\in\mathcal{N}'}\bar{f}_{f,m}(\boldsymbol\omega;\boldsymbol\omega',\mathbf{x}_n)$ can be written as~\eqref{eqn:ffmbar-composition}, as shown at the top of the next page,
with $p_{f,m}:\mathbb{R}^{{d+\sum_{i=0}^I E_{m,i}}}\to\mathbb{R}$, {$\mathbf q_{f,m,0}:\mathbb{R}^{d_0+H_mBI}\to\mathbb{R}^{E_{m,0}}$}, and $\mathbf q_{f,m,i}:\mathbb{R}^{d_0+d_i+{K_iB+H_mBI}}\to\mathbb{R}^{{E_{m,i}}}$, $i\in\mathcal{I}$, {for some positive integers $E_{m,i},i=0,1,\cdots, I$.}
Assume that the expressions of $\bar{f}_{f,m}$, $p_{f,m}$, {$\mathbf q_{f,m,0}$}, $\mathbf q_{f,m,i},i\in\mathcal{I}$, and $\mathbf{h}_{m,i}$, $m=0,\cdots,M$, $i\in\mathcal{I}$ are known to the server and {$I$} clients.
Each client $i\in\mathcal{I}$ computes $\mathbf{h}_{m,i}({\boldsymbol\omega}^{(t)}_i,\mathbf{x}_{n,i})$, $n\in\mathcal{N}^{(t)}$, $m=0,\cdots,M$ and sends them to the other clients.
Based on $\mathbf{h}_{m,i}({\boldsymbol\omega}^{(t)}_i,\mathbf{x}_{n,i})$, $m=0,\cdots,M$, $n\in\mathcal{N}^{(t)}$, $i\in\mathcal{I}$,
{the client with the highest computation speed (or any client) computes $\mathbf q_{f,m,0}\left({\boldsymbol\omega}^{(t)}_0,\left(\mathbf{h}_{m,i}({\boldsymbol\omega}^{(t)}_i,\mathbf{x}_{n,i})\right)_{n\in\mathcal{N}^{(t)},i\in\mathcal{I}}\right)$, $m=0,\cdots,M$ and sends them to the server.
Moreover, each client $i\in\mathcal{I}$ computes $\mathbf q_{f,m,i}\!\!\left(\!{\boldsymbol\omega}^{(t)}_0\!,{\boldsymbol\omega}^{(t)}_i\!,{(\mathbf{x}_{n,i})_{n\in\mathcal{N}^{(t)}}}, \!\left(\mathbf{h}_{m,j}({\boldsymbol\omega}^{(t)}_j\!,\mathbf{x}_{n,j})\right)_{n\in\mathcal{N}^{(t)},j\in\mathcal{I}}\!\right)$, $m=0,\cdots,M$ and sends them to the server.}
Then, the server solves the following convex approximate problem to obtain $\bar{\boldsymbol\omega}_{f}^{(t)}$.
\begin{Prob}[Convex Approximate Problem of Problem~\ref{Prob:con-feature-ep}]\label{Prob:con-feature-ap}
	\begin{align}
		(\bar{\boldsymbol\omega}_{f}^{(t)},\mathbf{s}_f^{(t)})\triangleq\mathop{\arg\min}_{\boldsymbol\omega,\mathbf{s}}
\ &\bar F^{(t)}_{f,0}(\boldsymbol\omega)+c\sum_{m=1}^M s_m\nonumber\\
		\text{s.t.}\ &\bar F^{(t)}_{f,m}(\boldsymbol\omega)\leq s_m,\ m=1,2,\cdots,M,\nonumber\\
		&s_m\geq 0,\ m=1,2,\cdots,M.\nonumber
	\end{align}
\end{Prob}

{Like} Problem~\ref{Prob:con-sample-ap}, Problem~\ref{Prob:con-feature-ap} is {a constrained convex problem that is always feasible} and can be readily solved. Given $\bar{\boldsymbol\omega}_{f}^{(t)}$, the server updates ${\boldsymbol\omega}_{f}^{(t)}$ according to \eqref{eqn:updatew-f}.
The detailed procedure is summarized in~Algorithm~\ref{alg:con-feature}. The convergence of Algorithm~\ref{alg:con-feature} is summarized below. Consider a sequence $\{c_j\}$. For all $j$, let $({\boldsymbol\omega}_{f,j}^\star,\mathbf s_{f,j}^\star)$ denote a limit point of $\{({\boldsymbol\omega}_{f}^{(t)},\mathbf{s}_f^{(t)})\}$ generated by Algorithm~\ref{alg:con-feature} with $c=c_j$.
\begin{Thm}[Convergence of Algorithm~\ref{alg:con-feature}]\label{thm:con-feature}
	Suppose that $f_{f,m}$ satisfies Assumption~\ref{asump:f} for all $m=0,\cdots,M$, $\bar{f}_{f,0}$ satisfies Assumption~\ref{asump:fbar}, $\bar{f}_{f,m}$ satisfies $\bar{f}_{f,m}(\boldsymbol\omega;\boldsymbol\omega,\mathbf{x})=f_{f,m}(\boldsymbol\omega;\mathbf{x})$ and Assumption~\ref{asump:fbar} for all $m=1,\cdots,M$, {the sequence $\{{\boldsymbol\omega}_{f}^{(t)}\}$ generated by Algorithm~\ref{alg:con-feature} with $c=c_j$ is bounded for all $j$,} and the sequence $\{c_j\}$ satisfies $0<c_j<c_{j+1}$ and $\lim_{j\to\infty}c_j=\infty$. Then, the following statements hold.
	i) For all $j$, if $\mathbf s_{f,j}^\star=\mathbf0$, then ${\boldsymbol\omega}_{f,j}^\star$ is a KKT point of Problem~\ref{Prob:con-feature} almost surely;
	%Problem~\ref{Prob:con-feature-ep};
	ii) A limit point of $\{({\boldsymbol\omega}_{f,j}^\star,\mathbf s_{f,j}^\star)\}$, denoted by $\{({\boldsymbol\omega}_{f,\infty}^\star,\mathbf s_{f,\infty}^\star)\}$, satisfies that $\mathbf s_{f,\infty}^\star=\mathbf{0}$, and ${\boldsymbol\omega}_{f,\infty}^\star$ is a KKT point of Problem~\ref{Prob:con-feature} almost surely.
\end{Thm}

\begin{IEEEproof}{Please refer to Appendix B.}
\end{IEEEproof}

%In practice, we can choose a sequence $\{c_j\}$ which satisfies that $0<c_j<c_{j+1}$, $\lim_{j\to\infty}c_j=\infty$, and $c_1$ is large and repeat Algorithm~\ref{alg:con-feature} with $c=c_j$ until $\left\Vert\mathbf s_{f,j}^\star\right\Vert_2$ is sufficiently small.

\subsubsection{Security Analysis}
%We establish the security of Algorithm~\ref{alg:con-feature}.
Suppose 1) for all $i\in\mathcal{I}$, mini-batch $\mathcal{N}'\subseteq\mathcal{N}$, {and $\boldsymbol\omega_i'\in \mathbb R^{d_i}$}, the system of equations w.r.t. {$(\boldsymbol\theta,(\mathbf{z}_{n})_{n=1,\cdots,B})\in \mathbb R^{d_i+BK_i}$ with $\boldsymbol\theta\in\mathbb R^{d_i}$ and $\mathbf{z}_{n}\in\mathbb{R}^{K_i}$, i.e., $\mathbf{h}_{m,i}(\boldsymbol\theta,\mathbf{z}_{n})=\mathbf{h}_{m,i}(\boldsymbol\omega'_i,\mathbf{x}_{n,i})$}, $n\in\mathcal{N}'$, $m=0,\cdots,M$, has an infinite (or a sufficiently large) number of solutions;
2) for any mini-batch $\mathcal{N}'\subseteq\mathcal{N}$ {and $\boldsymbol\omega'\in \mathbb R^{d}$}, the system of equations w.r.t. $\left(\mathbf{z}_{n,i}\right)_{n=1,\cdots,B,i\in\mathcal{I}}{\in\mathbb R^{BK}}$ {with} $\mathbf{z}_{n,i}\in\mathbb{R}^{K_i}$, i.e.,
$\mathbf q_{f,m,0}\left(\boldsymbol\omega'_0,\left(\mathbf{h}_{m,i}(\boldsymbol\omega'_i,\mathbf{z}_{n,i})\right)_{n=1,\cdots,B,i\in\mathcal{I}}\right)=\mathbf q_{f,m,0}\left(\boldsymbol\omega'_0,\left(\mathbf{h}_{m,i}(\boldsymbol\omega'_i,\mathbf{x}_{n,i})\right)_{n\in\mathcal{N}',i\in\mathcal{I}}\right)$, $m=0,\cdots,M$ and
$\mathbf q_{f,m,i}\left(\boldsymbol\omega'_0,\boldsymbol\omega'_i, {(\mathbf{z}_{n,i})_{n\in\mathcal{N}'}}, \left(\mathbf{h}_{m,j}(\boldsymbol\omega'_j,\mathbf{z}_{n,j})\right)_{n=1,\cdots,B,j\in\mathcal{I}}\right)=\mathbf q_{f,m,i}\left(\boldsymbol\omega'_0,\boldsymbol\omega'_i, {(\mathbf{x}_{n,i})_{n\in\mathcal{N}'}}, \left(\mathbf{h}_{m,j}(\boldsymbol\omega'_j,\mathbf{x}_{n,j})\right)_{n\in\mathcal{N}',j\in\mathcal{I}}\right)$, $m=0,\cdots,M$, $i\in\mathcal{I}$, has an infinite (or a sufficiently large) number of solutions.
In that case, raw data $\mathbf{x}_n$, $n\in\mathcal{N}^{(t)}$
{can hardly be extracted by any client from $\mathbf{h}_{m,i}({\boldsymbol\omega}^{(t)}_i,\mathbf{x}_{n,i})$, $m=0,\cdots,M$, $n\in\mathcal{N}^{(t)}$, $i\in\mathcal{I}$ or by the server from $\mathbf q_{f,m,0}\left({\boldsymbol\omega}^{(t)}_0,\left(\mathbf{h}_{m,i}({\boldsymbol\omega}^{(t)}_i\!,\mathbf{x}_{n,i})\right)_{n\in\mathcal{N}^{(t)},i\in\mathcal{I}}\right)$, $\mathbf q_{f,m,i}\!\left(\!{\boldsymbol\omega}^{(t)}_0\!,{\boldsymbol\omega}^{(t)}_i\!,{(\mathbf{x}_{n,i})_{n\in\mathcal{N}^{(t)}}}, \!\left(\mathbf{h}_{m,j}({\boldsymbol\omega}^{(t)}_j,\mathbf{x}_{n,j})\right)_{\!n\in\mathcal{N}^{(t)},\! j\in\mathcal{I}}\!\right)$, $m=0,\cdots,M$, $i\in\mathcal{I}$}
in Steps 4-6 of Algorithm~\ref{alg:con-feature}. Hence, Algorithm~\ref{alg:con-feature} can {reduce privacy risk}. {However, extra privacy mechanisms need to be investigated if the two assumptions mentioned above are not satisfied.}

\subsubsection{Algorithm Example}
We provide an example of $\bar{f}_{f,m}$, $m=0,\cdots,M$ with $\bar{f}_{f,0}$ satisfying Assumption~\ref{asump:fbar} and $\bar{f}_{f,m}$ satisfying $\bar{f}_{f,m}(\boldsymbol\omega;\boldsymbol\omega,\mathbf{x})=f_{f,m}(\boldsymbol\omega;\mathbf{x})$ and Assumption~\ref{asump:fbar} for all $m=1,\cdots,M$.
Specifically, we can choose $\bar{f}_{f,0}$ given by \eqref{eqn:ff0bar} and choose $\bar{f}_{f,m}$, $m=1,\cdots,M$ as follows:
\begin{align}
	\bar{f}_{f\!,m}(\boldsymbol\omega;\!{\boldsymbol\omega}_{f}^{(t)}\!\!,\!\mathbf{x}_n)\!\!=\!&f_{f\!,m}({\boldsymbol\omega}_{f}^{(t)}\!;\mathbf{x}_n)\!\!+\!\!\left(\!\nabla f_{f\!,m}({\boldsymbol\omega}_{f}^{(t)}\!;\mathbf{x}_n)\!\right)^T\!\!\!\left(\!\boldsymbol\omega\!-\!{\boldsymbol\omega}_{f}^{(t)}\!\right)\nonumber\\
	&+\tau\left\Vert{\boldsymbol\omega-{\boldsymbol\omega}_{f}^{(t)}}\right\Vert_2^2,\ m=1,\cdots,M, \label{eqn:ffmbar}
\end{align}
where $\tau>0$ can be any constant. Note that $\nabla_{\boldsymbol\omega} f_{f,m}({\boldsymbol\omega}_{f}^{(t)};\mathbf{x}_n)$ can be computed according to the chain rule, similarly to~\eqref{eqn:chainrule1} and~\eqref{eqn:chainrule2}. {Thus, $\sum_{n\in\mathcal{N}^{(t)}}\nabla_{\boldsymbol\omega_0} f_{f,0}({\boldsymbol\omega}_{f}^{(t)};\mathbf{x}_n)$
can be viewed as $\mathbf q_{f,0,0}\left({\boldsymbol\omega}^{(t)}_0,\left(\mathbf{h}_{0,i}({\boldsymbol\omega}^{(t)}_i,\mathbf{x}_{n,i})\right)_{n\in\mathcal{N}^{(t)},i\in\mathcal{I}}\right)$ {(implying $E_{0,0}=d_0$)}; $\left(\sum_{n\in\mathcal{N}^{(t)}} f_{f,m}({\boldsymbol\omega}_{f}^{(t)};\mathbf{x}_n), \sum_{n\in\mathcal{N}^{(t)}}\nabla_{\boldsymbol\omega_0} f_{f,m}({\boldsymbol\omega}_{f}^{(t)};\mathbf{x}_n)\right)$ can be viewed as $\mathbf q_{f,m,0}\left({\boldsymbol\omega}^{(t)}_0,\left(\mathbf{h}_{m,i}({\boldsymbol\omega}^{(t)}_i,\mathbf{x}_{n,i})\right)_{n\in\mathcal{N}^{(t)},i\in\mathcal{I}}\right)$ (implying $E_{m,0}=1+d_0$), for all $m=1,\cdots,M$; and $\sum_{n\in\mathcal{N}^{(t)}}\nabla_{\boldsymbol\omega_i} f_{f,m}({\boldsymbol\omega}_{f}^{(t)};\mathbf{x}_n)$ can be viewed as $\mathbf q_{f,m,i}\!\left({\boldsymbol\omega}^{(t)}_0,\!{\boldsymbol\omega}^{(t)}_i,\!{(\mathbf{x}_{n,i})_{\!n\in\mathcal{N}^{(t)}}}, \!\left(\mathbf{h}_{m,j}({\boldsymbol\omega}^{(t)}_j,\mathbf{x}_{n,j})\right)_{\!n\in\mathcal{N}^{(t)},\!j\in\mathcal{I}}\right)$ (implying $E_{m,i}=d_i$), for all $m=0,\cdots,M$, $i\in\mathcal{I}$.}
Recall that $\bar{F}^{(t)}_{f,0}(\boldsymbol\omega)$ is given in \eqref{eqn:Ff0bar_eg} with $f_{f,0}$ given in \eqref{eqn:ff0bar}. In addition, for all $m=1,\cdots,M$, substituting \eqref{eqn:ffmbar} into \eqref{eqn:Ffmbar}, $\bar{F}^{(t)}_{f,m}(\boldsymbol\omega)$ can be rewritten as:
\begin{align*}
	\bar{F}^{(t)}_{f,m}(\boldsymbol\omega)=&\hat f^{(t)}_{f,m,0}+\left(\hat{\mathbf f}^{(t)}_{f,m,1}\right)^T\boldsymbol\omega+\tau\left\Vert{\boldsymbol\omega}\right\Vert_2^2,\  m=1,\cdots,M, %\label{eqn:Ffmbar_eg}
\end{align*}
where $\hat f^{(t)}_{f,m,0}$ and $\hat{\mathbf f}^{(t)}_{f,m,1}\in\mathbb{R}^d$ are given by:
\begin{align}
	&\hat f^{(t)}_{f,m,0}=(1-\rho^{(t)})\hat f^{(t-1)}_{f,m,0}+\rho^{(t)}\frac{1}{B}\sum_{n\in\mathcal N^{(t)}}\Big(f_{f,m}({\boldsymbol\omega}_{f}^{(t)};\mathbf{x}_n)\nonumber\\
	&-\left(\nabla f_{f,m}({\boldsymbol\omega}_{f}^{(t)};\mathbf{x}_n)\right)^T{\boldsymbol\omega}_{f}^{(t)}+\tau\left\Vert{{\boldsymbol\omega}_{f}^{(t)}}\right\Vert_2^2\Big),\  m=1,\cdots,M, \nonumber\\
	&\hat{\mathbf f}^{(t)}_{f,m,1}=(1-\rho^{(t)})\hat{\mathbf f}^{(t-1)}_{f,m,1}+\rho^{(t)}\frac{1}{B}\nonumber\\
	&\times\!\!\sum_{n\in\mathcal N^{(t)}}\!\!\!\Big(\nabla f_{f,m}({\boldsymbol\omega}_{f}^{(t)};\mathbf{x}_n)-2\tau{\boldsymbol\omega}_{f}^{(t)}\Big),\ m=1,\cdots,M\nonumber
\end{align}
with $\hat f^{(0)}_{f,m,0}=0$ and $\hat{\mathbf f}^{(0)}_{f,m,1}=\mathbf 0$. Problem~\ref{Prob:con-feature-ap} with $\bar{f}_{f,0}$ given by~\eqref{eqn:ff0bar} and $\bar{f}_{f,m}$, $m=1,\cdots,M$ given by \eqref{eqn:ffmbar} is a convex quadratically constrained quadratic programming and can be solved using an {interior-point} method.

{Therefore, Steps 5-7 of Algorithm~\ref{alg:con-feature} with $\bar{f}_{f,0}$ given by~\eqref{eqn:ff0bar} and $\bar{f}_{f,m}, m=1,\cdots,M$ given by~\eqref{eqn:ffmbar} (i.e., an example of Algorithm~\ref{alg:con-feature}) are given below.}
In Step 5, the client with the highest computation speed (or any client) computes $\sum_{n\in\mathcal{N}^{(t)}}\nabla_{\boldsymbol\omega_0} f_{f,0}({\boldsymbol\omega}_{f}^{(t)};\mathbf{x}_n)\in \mathbb R^{d_0}$ and $\left(\sum_{n\in\mathcal{N}^{(t)}} f_{f,m}({\boldsymbol\omega}_{f}^{(t)};\mathbf{x}_n), \sum_{n\in\mathcal{N}^{(t)}}\nabla_{\boldsymbol\omega_0} f_{f,m}({\boldsymbol\omega}_{f}^{(t)};\mathbf{x}_n)\right)\in \mathbb R^{1+d_0}$, $m=1,\cdots,M$ and sends the $d_0$-dimensional vector and $M$ $(1+d_0)$-dimensional vectors to the server. In Step 6,
{each client $i\in \mathcal I$ computes $\sum_{n\in\mathcal{N}^{(t)}}\nabla_{\boldsymbol\omega_i} f_{f,m}({\boldsymbol\omega}_{f}^{(t)};\mathbf{x}_n)$, $m=0,\cdots,M$ and sends the $(M+1)$ $d_i$-dimensional vectors to the server.
In Step 7, the server calculates $(\bar{\boldsymbol\omega}_{f}^{(t)}, \mathbf s_f^{(t)})$ using an {interior-point} method.}
Suppose that for all $i\in\mathcal{I}$, mini-batch $\mathcal{N}'\subseteq\mathcal{N}$, {and $\boldsymbol\omega_i'\in \mathbb R^{d_i}$}, the system of equations w.r.t {$(\boldsymbol\theta,(\mathbf{z}_{n})_{n=1,\cdots,B})\in \mathbb R^{d_i+BK_i}$ with $\boldsymbol\theta\in\mathbb R^{d_i}$ and $\mathbf{z}_{n}\in\mathbb{R}^{K_i}$, i.e., $\mathbf{h}_{m,i}(\boldsymbol\theta,\mathbf{z}_{n})=\mathbf{h}_{m,i}(\boldsymbol\omega'_i,\mathbf{x}_{n,i})$}, $n\in\mathcal{N}'$, $m=0,\cdots,M$, has an infinite (or a sufficiently large) number of solutions;
and for all $\mathcal{N}'\subseteq\mathcal{N}$ {and $\boldsymbol\omega'\in \mathbb R^{d}$}, the system of equations w.r.t. $(\mathbf{z}_n)_{n=1,\cdots,B}$ {with $\mathbf{z}_n\in\mathbb{R}^{K}$}, i.e., $\sum_{n=1}^B f_{f,m}(\boldsymbol\omega';\mathbf{z}_n)=\sum_{n\in\mathcal{N}'} f_{f,m}(\boldsymbol\omega';\mathbf{x}_n)$, $m=1,\cdots,M$ and $\sum_{n=1}^B\nabla f_{f,m}(\boldsymbol\omega';\mathbf{z}_n)=\sum_{n\in\mathcal{N}'}\nabla f_{f,m}(\boldsymbol\omega';\mathbf{x}_n)$, $m=0,\cdots,M$, has an infinite (or a sufficiently large) number of solutions.
In that case, {the example of Algorithm~\ref{alg:con-feature}} can {reduce privacy risk}.

\section{Application Examples}\label{sec:application}
In this section, we customize the proposed algorithmic frameworks to some applications and provide detailed solutions for the specific problems.
{The server and  $I$ clients collaboratively solve  an $L$-class classification problem with a dataset of $N$ samples using FL. Denote $\mathcal P\triangleq\{1,\cdots, P\}$ and $\mathcal L\triangleq\{1,\cdots, L\}$.  The $n$-th sample is represented by $\mathbf x_n\triangleq (\mathbf z_n,\mathbf y_n)\in\mathbb R^K$,  where $K=P+L$, and $\mathbf z_n\triangleq(z_{n,p})_{p\in\mathcal P}\in \mathbb R^P$ and $\mathbf y_n\triangleq(y_{n,l})_{l\in\mathcal L}\in \{0,1\}^L$ represent the $P$ features and label of the $n$-th sample, respectively. In feature-based FL, $\mathcal P$ is partitioned into $I$ subsets, denoted by $\mathcal P_i, i\in \mathcal I$, and for each sample $n\in \mathcal N$,  client $i$ maintains {the} $P_i$ features  $\mathbf z_{n,i}\triangleq(z_{n,p})_{p\in\mathcal P_i}\in\mathbb R^{P_i}$ and the label $\mathbf y_n$. {Note that $P=\sum_{i\in\mathcal I}P_i$.} Thus, the $i$-th subvector for the $n$-th sample is given by $\mathbf x_{n,i}\triangleq(\mathbf z_{n,i},\mathbf y_{n})$.}

Consider a {two}-layer neural network, including an input layer composed of $P$ cells, a hidden layer composed of $J$ cells, and an output layer composed of $L$ cells. Denote $\mathcal J\triangleq\{1,\cdots, J\}$. The model parameters are represented by $\boldsymbol\omega\triangleq \left(({\omega}_{0,l,j})_{l\in\mathcal{L},j\in\mathcal{J}},({\omega}_{1,j,p})_{j\in\mathcal{J},p\in\mathcal{P}}\right)\in\mathbb R^d$, where $d=J(P+L)$. For feature-based FL,  $\boldsymbol\omega$ is also expressed as $\boldsymbol\omega=(\boldsymbol\omega_0,(\boldsymbol\omega_i)_{i\in\mathcal I})$, where $\boldsymbol\omega_0\triangleq({\omega}_{0,l,j})_{l\in\mathcal{L},j\in\mathcal{J}}$ and $\boldsymbol\omega_i\triangleq({\omega}_{1,j,p})_{j\in\mathcal{J},p\in\mathcal P_i}$, {$i\in\mathcal I$}.
We use the swish activation function $S(z)={z}/{(1+\exp(-z))}$~\cite{swish} for the hidden layer and the softmax activation function for the output layer. {Note that $S'(z)=\frac{1}{1+\exp(-z)}\left(1+\frac{z\exp(-z)}{1+\exp(-z)}\right)$.}
We consider the cross-entropy loss function. Thus, the resulting loss function for sample-based and feature-based FL {is} given by:
\begin{align}
	F(\boldsymbol\omega)\triangleq
	-\frac{1}{N}\sum_{n\in\mathcal{N}}\sum_{l\in\mathcal{L}} y_{n,l}\log\left(Q_l(\boldsymbol\omega;\mathbf{x}_n)\right),
	\label{eqn:Fcost}
\end{align}
where
	\begin{align}
		&Q_l(\boldsymbol\omega;\mathbf{x}_n)\triangleq\frac{\exp(\sum_{j\in\mathcal{J}}{\omega}_{{0},l,j} S(\sum_{{p\in\mathcal P}}{\omega}_{1,j,{p}}{z}_{n,{p}}))}{\sum_{h=1}^L \exp(\sum_{j\in\mathcal{J}}{\omega}_{{0},h,j} S(\sum_{{p\in\mathcal P}}{\omega}_{1,j,{p}}{z}_{n,{p}}))}.\nonumber
		%&\hspace{6cm} l\in\mathcal{L}.\label{eqn:Ql}
	\end{align}

{
For ease of exposition, in the rest of this section, we denote:
	\begin{align}
		&{\bar A}_{a,l,j}^{(t)}{\triangleq}
		\begin{cases} \sum_{i\in\mathcal{I}}\frac{N_i}{BN}\sum_{n\in\mathcal N_i^{(t)}}\bar a_{a,n,l,j},\ a=s\\
			\frac{1}{B}\sum_{n\in\mathcal N^{(t)}}\bar a_{a,n,l,j}, \ a=f
		\end{cases},
\label{eqn:Abar}\\
&{\bar B}_{a,j,p}^{(t)}{\triangleq}
		\begin{cases} \sum_{i\in\mathcal{I}}\frac{N_i}{BN}\sum_{n\in\mathcal N_i^{(t)}}\bar b_{a,n,j,p},\ a=s\\
			\frac{1}{B}\sum_{n\in\mathcal N^{(t)}}\bar b_{a,n,j,p},\ a=f
		\end{cases},
\label{eqn:Bbar}\\
&{\bar C}_{a}^{(t)}\!{\triangleq}\!
		\begin{cases}
\sum_{i\in\mathcal{I}}\frac{N_i}{BN}\sum_{n\in\mathcal N_i^{(t)}}\bar c_{a,n}
+\tau\left\Vert{{\boldsymbol\omega}_{a}^{(t)}}\right\Vert_2^2,\ a=s\!\!\!\!\!\!\!\!\\
			\frac{1}{B}\sum_{n\in\mathcal N^{(t)}}\bar c_{a,n}
+\tau\left\Vert{{\boldsymbol\omega}_{a}^{(t)}}\right\Vert_2^2,\ a=f
		\end{cases},\label{eqn:Cbar}
	\end{align}
where
\begin{align*}
\bar a_{a,n,l,j}\!\!\triangleq&(Q_l({\boldsymbol\omega}_{a}^{(t)};\mathbf{x}_n)\!\!-\!\!y_{n,l})S(\sum_{p'=1}^P\omega^{(t)}_{a,1,j,p'}x_{n,p'}),\\
\bar b_{a,n,j,p}\!\!\triangleq\!\!&\sum_{l\in\mathcal{L}}\!(\!Q_l(\!{\boldsymbol\omega}_{a}^{(t)}\!;\mathbf{x}_n\!)\!\!-\!\!y_{n,l}\!)S'\!(\!\sum_{p'=1}^P\!\!\omega^{(t)}_{a,1,j,p'}x_{n,p'}\!)\omega^{(t)}_{a,0,l,j}x_{n,p},\\
\bar c_{a,n}\!\!\triangleq\!\!&\sum_{l\in\mathcal{L}}y_{n,l}
\log(Q_l({\boldsymbol\omega}_{a}^{(t)};\mathbf{x}_n)).
\end{align*}
}

\subsection{Unconstrained Federated Optimization}\label{sec:classification}
For $a=s,f$, one unconstrained federated optimization formulation for the $L$-class classification problem is to minimize the weighted sum of the loss function $F(\boldsymbol\omega)$ in~\eqref{eqn:Fcost} {and} the $\ell_2$-norm regularization term $\left\Vert\boldsymbol\omega\right\Vert^2_2$:
\begin{align}
	\min_{\boldsymbol\omega}\quad&F_{a,0}(\boldsymbol\omega)\triangleq F(\boldsymbol\omega)+\lambda\left\Vert\boldsymbol\omega\right\Vert^2_2\label{prob:class-uncon}
\end{align}
where $\lambda>0$ is the regularization parameter that trades off the cost and model sparsity. Obviously, $F(\boldsymbol\omega)+\lambda\left\Vert\boldsymbol\omega\right\Vert^2_2$ satisfies the additional restrictions on the structure of $F_{f,0}(\boldsymbol\omega)$.
We can {view $-\sum_{l\in\mathcal{L}} y_{n,l}\log\left(Q_l(\boldsymbol\omega;\mathbf{x}_n)\right)$ as $f_{a,0}(\boldsymbol\omega;\mathbf{x}_n)$},
apply Algorithm~\ref{alg:uncon-sample} with $\bar{f}_{s,0}(\boldsymbol\omega;{\boldsymbol\omega}_{s}^{(t)},\mathbf{x}_n)$ given by \eqref{eqn:fs0bar} to solve the problem in~\eqref{prob:class-uncon} for $a=s$,
and apply Algorithm~\ref{alg:uncon-feature} with $\bar{f}_{f,0}(\boldsymbol\omega;{\boldsymbol\omega}_{f}^{(t)},\mathbf{x}_n)$ given by \eqref{eqn:ff0bar} to solve the problem in~\eqref{prob:class-uncon} for $a=f$.

First, we present the details of Step 4 in Algorithm~\ref{alg:uncon-sample} and the details of Steps 4-6 in Algorithm~\ref{alg:uncon-feature}.
%for solving the unconstrained federated optimization problem in~\eqref{prob:class-uncon}.
In Step 4 of Algorithm~\ref{alg:uncon-sample}, each client $i$ computes {$((\sum_{n\in\mathcal N_i^{(t)}}\bar a_{s,n,l,j})_{l\in\mathcal{L},j\in\mathcal{J}},(\sum_{n\in\mathcal N_i^{(t)}}\bar b_{s,n,j,p})_{j\in\mathcal{J},p\in\mathcal{P}})$} and sends {it} to the server.
In Steps 4-6 of Algorithm~\ref{alg:uncon-feature}, each client $i$ computes {$(\omega^{(t)}_{f,1,j,p}x_{n,p})_{j\in\mathcal{J},p\in\mathcal{P}_i}$ $n\in\mathcal N^{(t)}$}
and sends them to the other clients;
based on {$(\omega^{(t)}_{f,1,j,p}x_{n,p})_{j\in\mathcal{J},p\in\mathcal{P}_i}$, $n\in\mathcal N^{(t)}$},
the client with the highest computation speed (or any client) computes
{$(\sum_{n\in\mathcal N^{(t)}}\bar a_{f,n,l,j})_{j\in\mathcal{J},l\in\mathcal{L}}$} and sends {it} to the server;
each client $i$ computes
{$(\sum_{n\in\mathcal N^{(t)}}\bar b_{f,n,j,p})_{j\in\mathcal{J},p\in\mathcal{P}_i}$}
and sends {it} to the server.

Next, we present the details of Step 5 in Algorithm~\ref{alg:uncon-sample} and the details of Step 7 in Algorithm~\ref{alg:uncon-feature}. For $a=s,f$, the  convex approximate problem is given by:
\begin{align}
	\min_{\boldsymbol\omega}\quad&\bar F_{a,0}^{(t)}(\boldsymbol\omega)=\bar F_{a}^{(t)}(\boldsymbol\omega)+2\lambda(\boldsymbol\beta^{(t)})^T\boldsymbol\omega\label{prob:class-uncon-ap}
\end{align}
where
$\bar F_{a}^{(t)}(\boldsymbol\omega)$ is given by
\begin{align}
\bar F_{a}^{(t)}(\boldsymbol\omega)
\!=\!\sum_{l\in\mathcal{L}}\!\sum_{j\in\mathcal{J}} {{A}_{a,l,j}^{(t)}}{\omega}_{0,l,j}\!+\!\sum_{j\in\mathcal{J}}\!\sum_{p\in\mathcal{P}}{{B}_{a,j,p}^{(t)}}{\omega}_{1,j,p} \!+\!\tau\left\Vert{\boldsymbol\omega}\right\Vert_2^2,\label{eqn:fbar-app}
\end{align}
and $\boldsymbol\beta^{(t)}\in\mathbb{R}^d$, ${{A}_{a,l,j}^{(t)}}\in\mathbb{R}$, and ${{B}_{a,j,p}^{(t)}}\in\mathbb{R}$ are updated according to:
\begin{align} &\boldsymbol\beta^{(t)}=(1-\rho^{(t)})\boldsymbol\beta^{(t-1)}+\rho^{(t)}{\boldsymbol\omega}_{a}^{(t)},\\ &{{A}_{a,l,j}^{(t)}}=(1-\rho^{(t)}){{A}_{a,l,j}^{(t-1)}}+\rho^{(t)}\left({\bar A}_{a,l,j}^{(t)}-2\tau\omega^{(t)}_{a,0,l,j}\right), \label{eqn:A}\\
&{{B}_{a,j,p}^{(t)}}\!=(1-\rho^{(t)}){{B}_{a,j,p}^{(t-1)}}+\rho^{(t)}\left({\bar B}_{a,j,p}^{(t)}\!-2\tau\omega^{(t)}_{a,1,j,p}\!\right),\label{eqn:B}
\end{align}
respectively, with ${\boldsymbol{\beta}}^{(0)}=\mathbf 0$ and ${{A}_{a,l,j}^{(0)}}={{B}_{a,j,p}^{(0)}}=0$.
Here, ${\bar A}_{a,l,j}^{(t)}$ and ${\bar B}_{a,j,p}^{(t)}$ are given by~\eqref{eqn:Abar} and~\eqref{eqn:Bbar} respectively.
By~\eqref{eqn:omegasbar} for $a=s$ and~\eqref{eqn:omegafbar} for $a=f$, the closed-form solutions of the problem in~\eqref{prob:class-uncon-ap} for $a=s,f$ are given by:
\begin{align} &\bar{\omega}_{a,0,l,j}^{(t)}=-\frac{1}{2\tau}\left({{A}_{a,l,j}^{(t)}}+2\lambda{\beta}_{2,l,j}^{(t)}\right),\ l\in\mathcal{L},\ j\in\mathcal{J},\label{eqn:omega1-uncon}\\ &\bar{\omega}_{a,1,j,p}^{(t)}=-\frac{1}{2\tau}\left({{B}_{a,j,p}^{(t)}}+2\lambda{\beta}_{1,j,p}^{(t)}\right),\  j\in\mathcal{J},\ p\in\mathcal{P}.\label{eqn:omega2-uncon}
\end{align}
Thus, in Step 5 in Algorithm~\ref{alg:uncon-sample} and Step 7 in Algorithm~\ref{alg:uncon-feature}, the server only needs to compute {$\bar{\boldsymbol\omega}_{a}^{(t)}$} according to{~\eqref{eqn:omega1-uncon} and~\eqref{eqn:omega2-uncon}}.

Theorem~\ref{thm:uncon-sample} and Theorem~\ref{thm:uncon-feature} guarantee the convergences of Algorithm~\ref{alg:uncon-sample} and Algorithm~\ref{alg:uncon-feature}, respectively, as Assumption~\ref{asump:f} and Assumption~\ref{asump:fbar} are satisfied.

\subsection{Constrained Federated Optimization}
For $a=s,f$, one constrained federated optimization formulation for the $L$-class classification problem is to minimize the $\ell_2$-norm of the network parameters $\Vert\boldsymbol\omega\Vert^2_2$ under a constraint on the loss function $F(\boldsymbol\omega)$ in~\eqref{eqn:Fcost}:
\begin{align}
	\min_{\boldsymbol\omega}\quad& F_{a,0}(\boldsymbol\omega)\triangleq\left\Vert\boldsymbol\omega\right\Vert^2_2\label{prob:class-con}\\
	\text{s.t.}\quad &F_{a,1}(\boldsymbol\omega)\triangleq F(\boldsymbol\omega)-U\leq0,\nonumber
\end{align}
where $U$ represents the limit on the cost.  {We can view $0$ and $-\sum_{l\in\mathcal{L}} y_{n,l}\log\left(Q_l(\boldsymbol\omega;\mathbf{x}_n)\right)$ as $f_{a,0}(\boldsymbol\omega;\mathbf{x}_n)$ and $f_{a,1}(\boldsymbol\omega;\mathbf{x}_n)$, respectively.}
Then, we can apply Algorithm~\ref{alg:con-sample} with $\bar{f}_{s,0}(\boldsymbol\omega;{\boldsymbol\omega}_{s}^{(t)},\mathbf{x}_n)$ given by \eqref{eqn:fs0bar} and $\bar{f}_{s,1}(\boldsymbol\omega;{\boldsymbol\omega}_{s}^{(t)},\mathbf{x}_n)$ given by \eqref{eqn:fsmbar} to solve the problem in~\eqref{prob:class-con} for $a=s$ and apply Algorithm~\ref{alg:con-feature} with $\bar{f}_{f,0}(\boldsymbol\omega;{\boldsymbol\omega}_{f}^{(t)},\mathbf{x}_n)$ given by \eqref{eqn:ff0bar} and $\bar{f}_{f,1}(\boldsymbol\omega;{\boldsymbol\omega}_{f}^{(t)},\mathbf{x}_n)$ given by \eqref{eqn:ffmbar} to solve the problem in~\eqref{prob:class-con} for $a=f$.

First, we present the details of Step 4 in Algorithm~\ref{alg:con-sample} and the details of Steps 4-6 in Algorithm~\ref{alg:con-feature}.
In Step 4 of Algorithm~\ref{alg:con-sample}, each client $i$ computes
{$((\sum_{n\in\mathcal N_i^{(t)}}\bar a_{s,n,l,j})_{l\in\mathcal{L},j\in\mathcal{J}},(\sum_{n\in\mathcal N_i^{(t)}}\bar b_{s,n,j,p})_{j\in\mathcal{J},p\in\mathcal{P}})$}
and {$\sum_{n\in\mathcal N_i^{(t)}}\bar c_{s,n}$}
and sends them to the server.
In Steps 4-6 of Algorithm~\ref{alg:con-feature}, each client $i$ computes
{$(\omega^{(t)}_{f,1,j,p}x_{n,p})_{j\in\mathcal{J},p\in\mathcal{P}_i}$, $n\in\mathcal N^{(t)}$} and sends them to the other clients;
based on {$(\omega^{(t)}_{f,1,j,p}x_{n,p})_{j\in\mathcal{J},p\in\mathcal{P}_i}$, $n\in\mathcal N^{(t)}$},
the client with the highest computation speed (or any client) computes
{$(\sum_{n\in\mathcal N^{(t)}}\bar a_{f,n,l,j})_{l\in\mathcal{L},j\in\mathcal{J}}$}
and {$\sum_{n\in\mathcal N^{(t)}}\bar c_{f,n}$} and sends them to the server;
each client $i$ computes
{$\left(\sum_{n\in\mathcal N^{(t)}}{\bar b_{f,n,j,p}}\right)_{j\in\mathcal{J},p\in\mathcal P_i}$}
and sends {it} to the server.

Next, we present the details of Step 5 in {Algorithm~\ref{alg:con-sample}} and the details of Step 7 in {Algorithm~\ref{alg:con-feature}}. For $a=s,f$,  the  convex approximate problem is given by:
\begin{align} \min_{\boldsymbol\omega,s}\quad&\left\Vert\boldsymbol\omega\right\Vert^2_2+c s\label{prob:class-con-ap}\\
	\text{s.t.}\quad &\bar F_{a}^{(t)}(\boldsymbol\omega)+{C}_{a}^{(t)}-U\leq s,\nonumber\\
	&s\geq0, \nonumber
\end{align}
where $\bar F_{a}^{(t)}(\boldsymbol\omega)$ is given by~\eqref{eqn:fbar-app} with ${{A}_{a,l,j}^{(t)}}$, ${{B}_{a,j,p}^{(t)}}$, and ${C}_{a}^{(t)}$ updated according to~{\eqref{eqn:A}, \eqref{eqn:B}}, and
\begin{align} &{C}_{a}^{(t)}=(1-\rho^{(t)}){C}_{a}^{(t-1)}+\nonumber\\
	&\rho^{(t)}\bigg({\bar C}_{a}^{(t)}
	-\sum_{l\in\mathcal{L}}\sum_{j\in\mathcal{J}} {\bar A}_{a,l,j}^{(t)}\omega^{(t)}_{a,0,l,j}-\sum_{j\in\mathcal{J}}\sum_{p\in\mathcal{P}} {\bar B}_{a,j,p}^{(t)}\omega^{(t)}_{a,1,j,p}\bigg), \label{eqn:C}
\end{align}
respectively, with ${C}_{a}^{(0)}=0$ and ${\bar C}_{a}^{(t)}$ given by~\eqref{eqn:Cbar}.
By the KKT conditions, the closed-form solutions of the problem in~\eqref{prob:class-con-ap} for $a=s,f$ are given as follows.
\begin{Lem}[Optimal Solution of Problem in~\eqref{prob:class-con-ap}]\label{lem:closedform}
	\begin{align}
&\bar{\omega}_{a,0,l,j}^{(t)}=-\frac{\nu {A}_{a,l,j}^{(t)}}{2(1+\nu\tau)},\ l\in\mathcal{L},\ j\in\mathcal{J},\label{eqn:omega2-con}\\
&\bar{\omega}_{a,1,j,p}^{(t)}=-\frac{\nu {B}_{a,j,p}^{(t)}}{2(1+\nu\tau)},\ j\in\mathcal{J},\ p\in\mathcal{P},\label{eqn:omega1-con}
	\end{align}
	where
	\begin{align}
		&\nu\!=\!
		\begin{cases} \left[\frac{1}{\tau}\left(\sqrt{\frac{b}{b+4\tau(U-{C}_{a}^{(t)})}}-1\right)\right]_0^c,\ &b+4\tau(U-{C}_{a}^{(t)})\!>\!0\\
			c,\ &b+4\tau(U-{C}_{a}^{(t)})\!\leq\!0
		\end{cases},\nonumber\\
		&b\!=\!\sum_{l\in\mathcal{L}}\sum_{j\in\mathcal{J}} ({A}_{a,l,j}^{(t)})^2+\sum_{j\in\mathcal{J}}\sum_{p\in\mathcal{P}} ({B}_{a,j,p}^{(t)})^2.\label{eqn:b}
	\end{align}
	Here, $[x]^c_0\triangleq\min\left\{\max\{x,0\},c\right\}$.
\end{Lem}
\begin{IEEEproof} {Please refer to Appendix C.}
\end{IEEEproof}

Thus, in Step 5 of Algorithm~\ref{alg:con-sample} and Step 7 of Algorithm~\ref{alg:con-feature}, the server only needs to compute {$\bar{\boldsymbol\omega}_{a}^{(t)}$} according to~\eqref{eqn:omega2-con} and~\eqref{eqn:omega1-con}.

The convergences of Algorithm~\ref{alg:con-sample} and Algorithm~\ref{alg:con-feature} are guaranteed by Theorem~\ref{thm:con-sample} and Theorem~\ref{thm:con-feature}, respectively, as Assumption~\ref{asump:f} and Assumption~\ref{asump:fbar} are satisfied.
\subsection{Comparisons of Two Formulations}
Both the unconstrained federated optimization formulation in~\eqref{prob:class-uncon} and constrained federated optimization formulation in~\eqref{prob:class-con} allow tradeoffs between the cost and model sparsity~\cite{foucart2017mathematical}. The equivalence between the two formulations is summarized in the following theorem.
%~\cite{foucart2017mathematical}
\begin{Thm}[Equivalence between Problems in~\eqref{prob:class-uncon} and~\eqref{prob:class-con}]\label{thm:compare}
	i) If $\boldsymbol{\omega}^*$ is a locally optimal solution of the problem in~\eqref{prob:class-uncon} with $\lambda>0$, then there exists $U\geq0$ such that $\boldsymbol{\omega}^*$ is a locally optimal solution of the problem in~\eqref{prob:class-con}.
	ii) If $\boldsymbol{\omega}^{\dagger}$ is a locally optimal solution of the problem in~\eqref{prob:class-con} with $U>0$, which is regular and satisfies the KKT conditions together with a corresponding Lagrange multiplier $\xi>0$, then there exists $\lambda>0$ such that $\boldsymbol{\omega}^{\dagger}$ is a stationary point of the problem in~\eqref{prob:class-uncon}. If, in addition, $\lambda$ and $\boldsymbol{\omega}^{\dagger}$ satisfy $\nabla^2 F(\boldsymbol{\omega}^{\dagger})+\lambda I\succeq 0$, then $\boldsymbol{\omega}^{\dagger}$ is a locally optimal solution of the problem in~\eqref{prob:class-uncon}.
\end{Thm}

\begin{IEEEproof}{Please refer to Appendix D.}
\end{IEEEproof}

By the above theorem, we know that the problem in~\eqref{prob:class-uncon} and the problem in~\eqref{prob:class-con} have the same locally optimal solution for certain $\lambda$ and $U$ under some conditions.
Besides, we can tradeoff between the training accuracy and model sparsity of each formulation. It is evident that with {the} constrained federated optimization formulation {in~\eqref{prob:class-con}}, one can set an explicit constraint on the training cost to control the test accuracy effectively.

\section{Numerical Results}\label{sec:simu}
In this section, we numerically evaluate the proposed examples of Algorithms~\ref{alg:uncon-sample}-\ref{alg:con-feature} using the application examples in Section~\ref{sec:application}.\footnote{Source code for the experiments is available at \cite{GitHub}.} For unconstrained federated optimization, we adopt the existing SGD-based~\cite{mcmahan2017communication,yu2019parallel,hardy2017private}  and momentum SGD-based~\cite{9003425} {FL} algorithms, called SGD and SGD-m, respectively, as the baseline algorithms for the proposed examples of Algorithm~\ref{alg:uncon-sample} and Algorithm~\ref{alg:uncon-feature}. Let $E$ denote the number of local SGD {(momentum SGD)  updates for  sample-based SGD (SGD-m)}. Note that sample-based  SGD with $B\times E=N$ becomes FedAvg\cite{mcmahan2017communication}. Feature-based SGD and SGD-m adopt the information collection mechanism used in Algorithms~\ref{alg:uncon-feature} (i.e., the extension of the one in \cite{hardy2017private}).  In each communication round, each proposed algorithm executes one iteration,  each sample-based SGD (SGD-m) executes one global iteration and $E$ local SGD (momentum SGD) updates, and each feature-based  SGD (SGD-m) executes one global iteration. Algorithm~\ref{alg:uncon-sample} (Algorithm~\ref{alg:uncon-feature}) and its baseline algorithms have the same communication load per communication round. Besides, if the value of $B$ for Algorithm~\ref{alg:uncon-sample} (Algorithm~\ref{alg:uncon-feature})  and the value of $B\times E$ for each sample-based ($B$ for each feature-based) baseline algorithm are equal, the two algorithms have the same order of computational complexity per communication round.\footnote{{The example of Algorithm~\ref{alg:uncon-sample} (Algorithm~\ref{alg:uncon-feature}) has the same level of privacy protection as its baseline algorithms, as illustrated in Section~\ref{subsec:uncon-sample} (Section~\ref{subsec:uncon-feature}).}}

{We set $\lambda=10^{-5}$ and $U=0.13$ for the unconstrained and constrained federated optimization problems in \eqref{prob:class-uncon}
 and \eqref{prob:class-con}, respectively, unless otherwise specified.} We carry our experiments on Mnist dataset.
For the training model, we set $N=60000$, $I=10$, $K=784$, $J=128$, and $L=10$.
For the proposed algorithms, we choose $T=1000$, $c=10^5$, $\rho^{(t)}=a_1/t^{\alpha}$ and $\gamma^{(t)}=a_2/t^{\alpha}$ with $a_1=0.9,0.3,0.2$, $a_2=0.5,0.3,0.3$, $\alpha=0.1,0.1,0.1$, and $\tau=0.2,0.05,0.03$ for batch sizes $B=10,100,6000$ in sample-based FL
and  $a_1=0.9,0.9,0.3$, $a_2=0.3,0.5,0.3$, $\alpha=0.3,0.1,0.1$, and $\tau=0.1,0.2,0.05$ for batch sizes $B=10,100,1000$ in feature-based FL.
For SGD, the learning rate is set as $r=\bar a/t^{\bar\alpha}$ {with $\bar a=0.3$ and $\bar\alpha=0.3$}.
{For SGD-m, the learning rate is set as $r=\bar a$ with $\bar a=0.3$ and the momentum parameter is set as $\bar\beta=0.1$.}
Note that {all} the algorithm parameters are selected using a grid search method, and all the results are given by averaging over ten runs.

%{For unconstrained sample-based FL, the proposed example for Algorithm~\ref{alg:uncon-sample} and the adopted baseline algorithms have the same communication load per communication round.
%For unconstrained feature-based FL, the proposed Algorithm~\ref{alg:uncon-feature} has the same communication load per communication round as the baseline feature-based FL algorithms via SGD at the same level of privacy protection.
%Thus, the number of (global) iterations (communication rounds) is proportional to the communication cost.}
%{Besides, for a fair comparison, we choose the same algorithm parameters $B\times E$ for sample-based FL and $B$ for feature-based FL, indicating that the same number of samples are used in each communication round, hence the computational complexities of the proposed examples for Algorithm~\ref{alg:uncon-sample} and Algorithm~\ref{alg:uncon-feature} have the same computational complexity in order as the existing SGD-based FL algorithms.
%Thus, the computation and communication costs can be measured by the number of communication rounds.}

\begin{figure}[h]
	\begin{center}
		\subfigure[\scriptsize{
			Training cost {for unconstrained federated optimization at $B, B\times E=10, 100$.}}\label{fig:SU_cost}]
		{\resizebox{4.3cm}{!}{\includegraphics{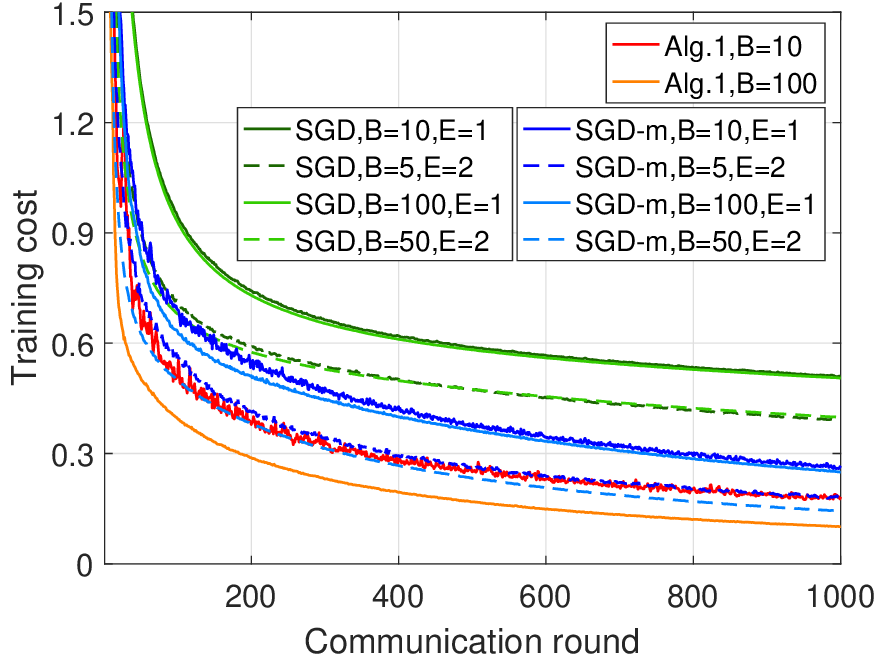}}}
		\subfigure[\scriptsize{
			Test accuracy {for unconstrained federated optimization at $B, B\times E=10, 100$.}}\label{fig:SU_acc}]
		{\resizebox{4.3cm}{!}{\includegraphics{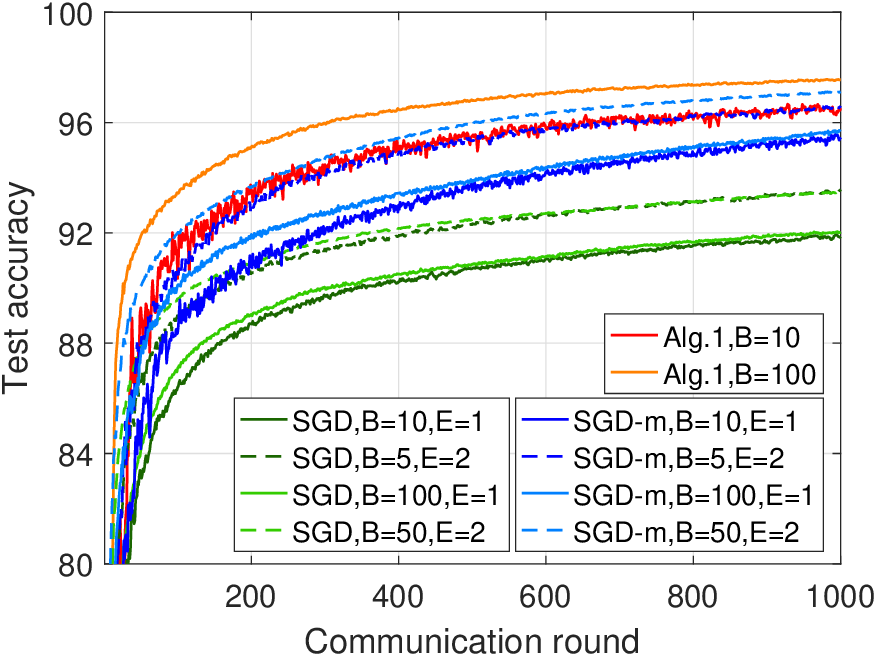}}}\\
		\subfigure[\scriptsize{
			Training cost {for unconstrained federated optimization at $B, B\times E=6000$.}}\label{fig:SU_cost_2}]
		{\resizebox{4.3cm}{!}{\includegraphics{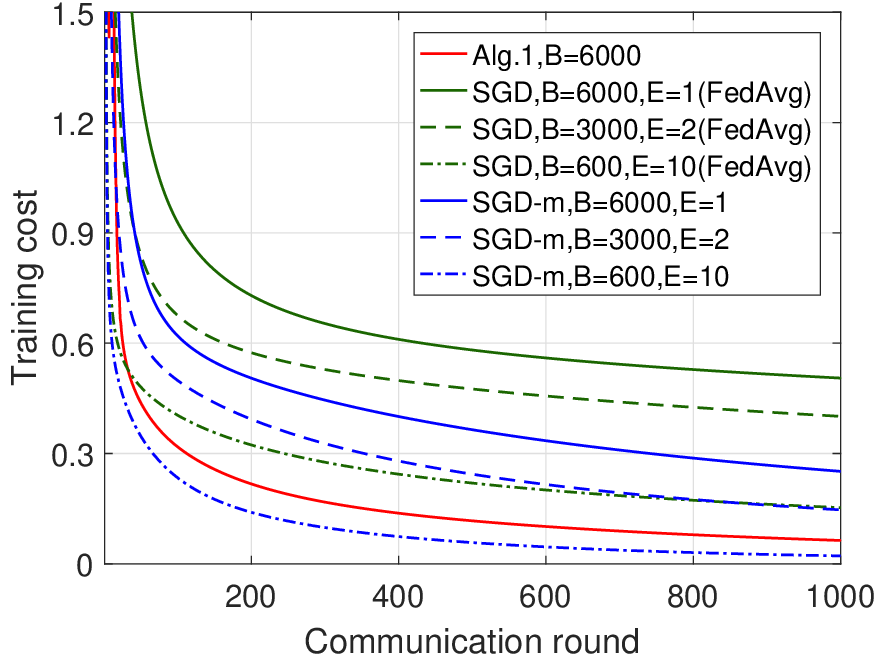}}}
		\subfigure[\scriptsize{
			Test accuracy {for unconstrained federated optimization at $B, B\times E=6000$.}}\label{fig:SU_acc_2}]
		{\resizebox{4.3cm}{!}{\includegraphics{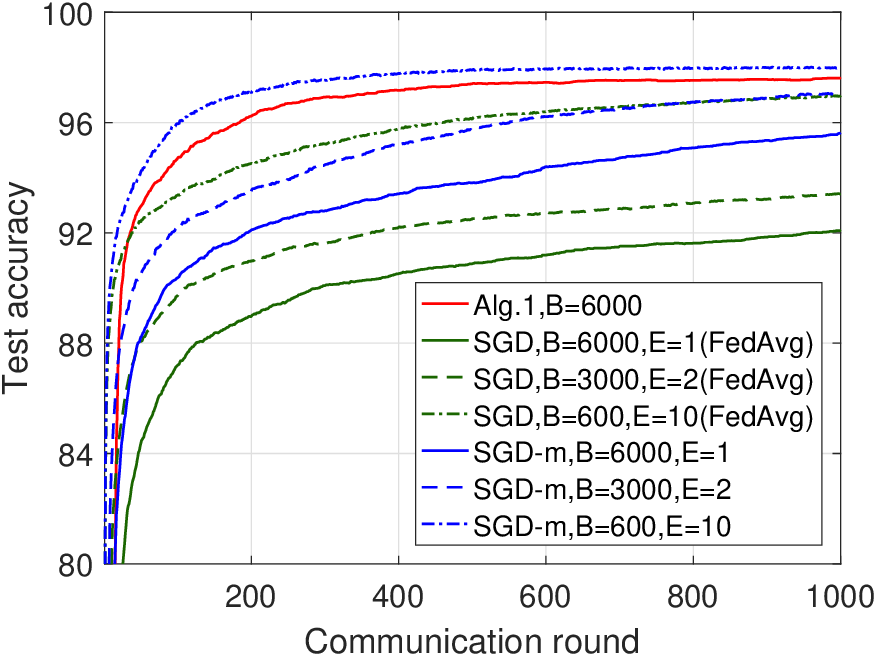}}}\\
		\subfigure[\scriptsize{
			Training cost {for constrained federated optimization at $B=10,100, 6000$.}}\label{fig:SC_cost}]
		{\resizebox{4.3cm}{!}{\includegraphics{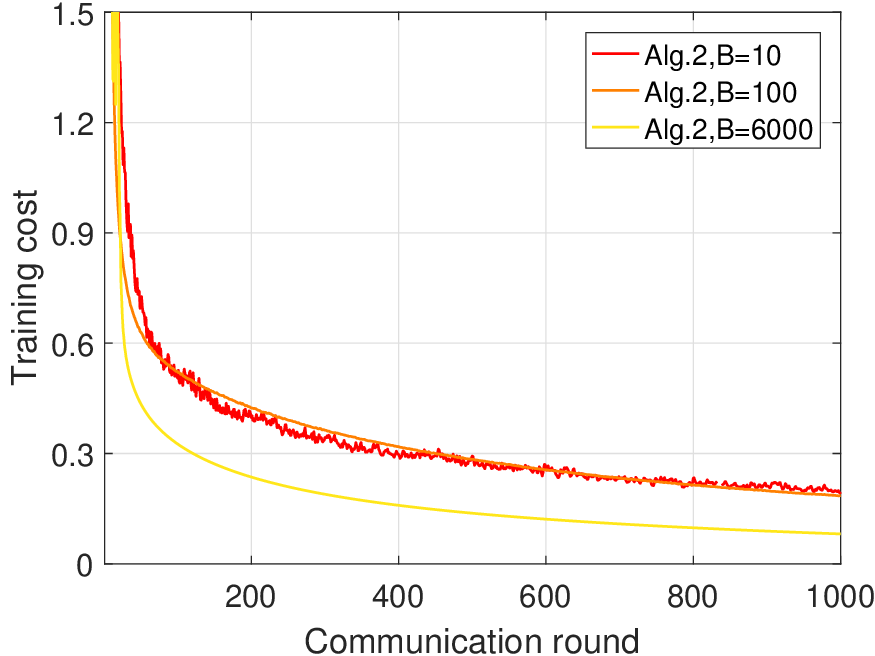}}}
		\subfigure[\scriptsize{
			Test accuracy {for constrained federated optimization at $B=10,100, 6000$.}}\label{fig:SC_acc}]
		{\resizebox{4.3cm}{!}{\includegraphics{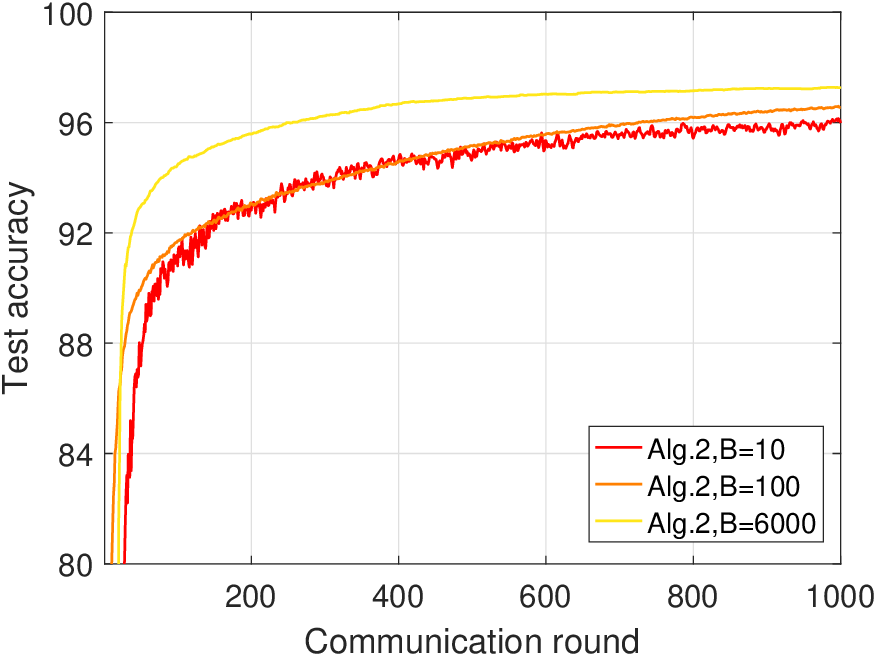}}}
	\end{center}
	\caption{\small{Training cost {$F({\boldsymbol\omega}_{s}^{(t)})$} and test accuracy {at ${\boldsymbol\omega}_{s}^{(t)}$} versus {communication round index $t$} for sample-based FL.}}
	\label{fig:Sam}
\end{figure}

\begin{figure}[h]
	\begin{center}
		\subfigure[\scriptsize{
			Training cost {for unconstrained federated optimization at $B=10,100, 1000$.}}\label{fig:FU_cost}]
		{\resizebox{4.2cm}{!}{\includegraphics{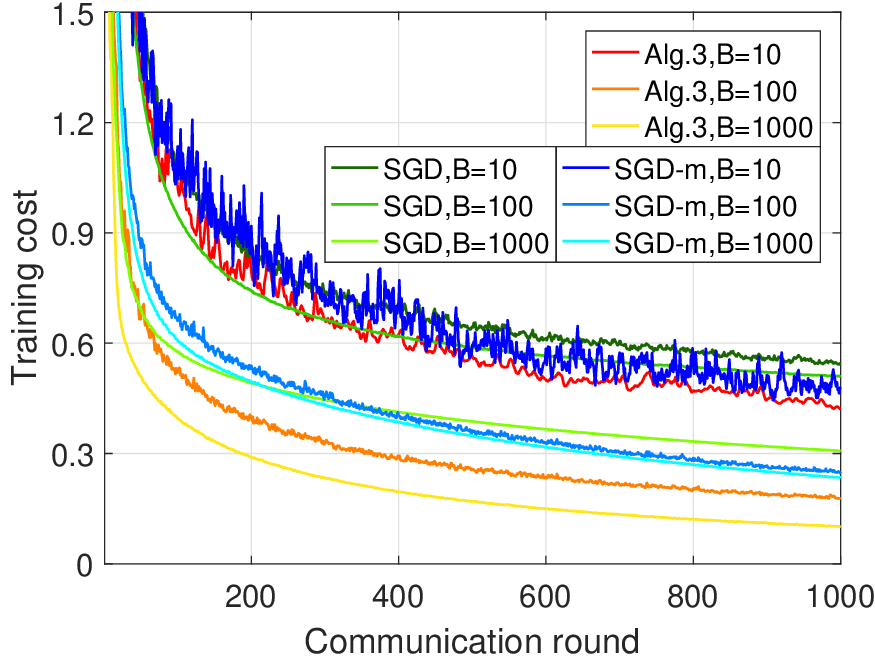}}}\quad
		\subfigure[\scriptsize{
			Test accuracy {for unconstrained federated optimization at $B=10,100, 1000$.}}\label{fig:FU_acc}]
		{\resizebox{4.2cm}{!}{\includegraphics{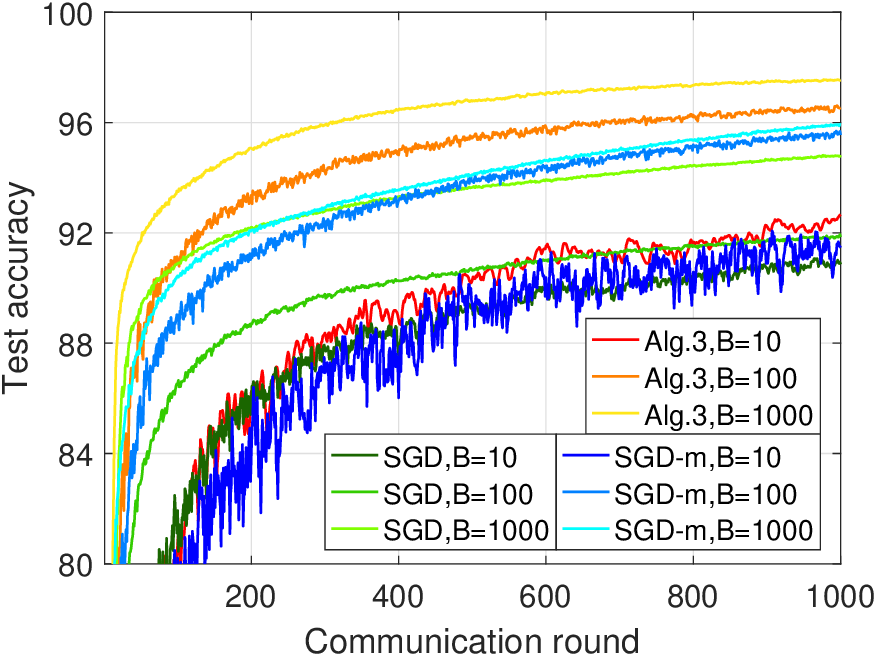}}}\\
		\subfigure[\scriptsize{
			Training cost {for constrained federated optimization at $B=10,100, 1000$.}}\label{fig:FC_cost}]
		{\resizebox{4.2cm}{!}{\includegraphics{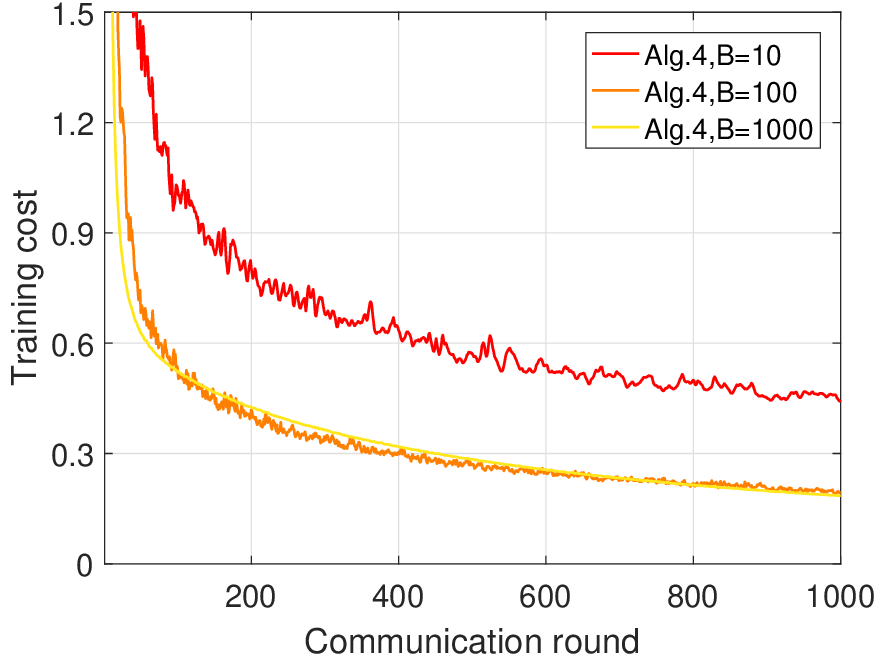}}}\quad
		\subfigure[\scriptsize{
			Test accuracy {for constrained federated optimization at $B=10,100, 1000$.}}\label{fig:FC_acc}]
		{\resizebox{4.2cm}{!}{\includegraphics{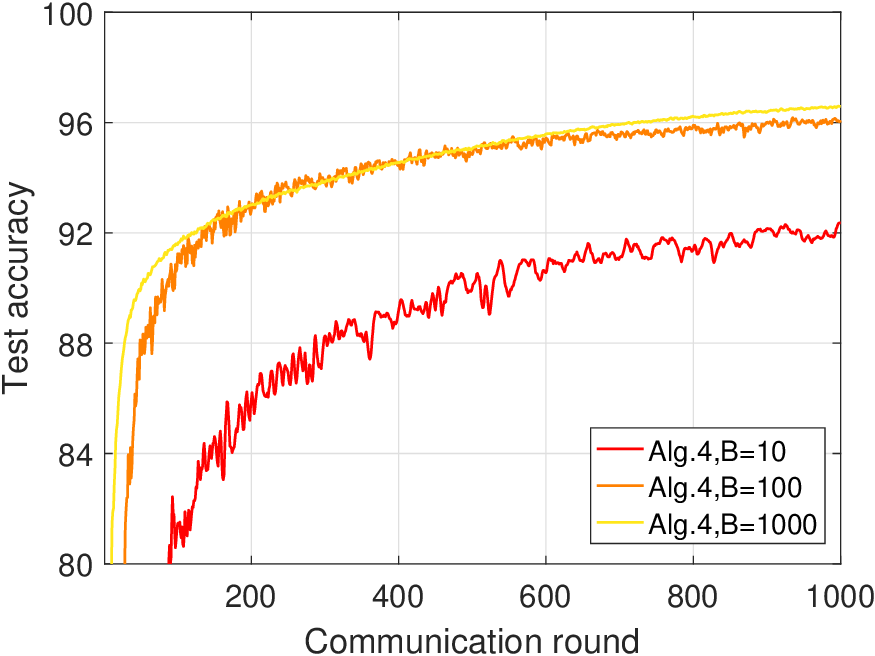}}}
	\end{center}
	\caption{\small{Training cost {$F({\boldsymbol\omega}_{f}^{(t)})$} and test accuracy {at ${\boldsymbol\omega}_{f}^{(t)}$} versus {communication round index $t$} for feature-based FL.}}
	\label{fig:Fea}
\end{figure}

\begin{figure}[h]
	\begin{center}
		\subfigure[\scriptsize{
			{Unconstrained sample-based FL.}}\label{fig:SCompComm}]
		{\resizebox{4.2cm}{!}{\includegraphics{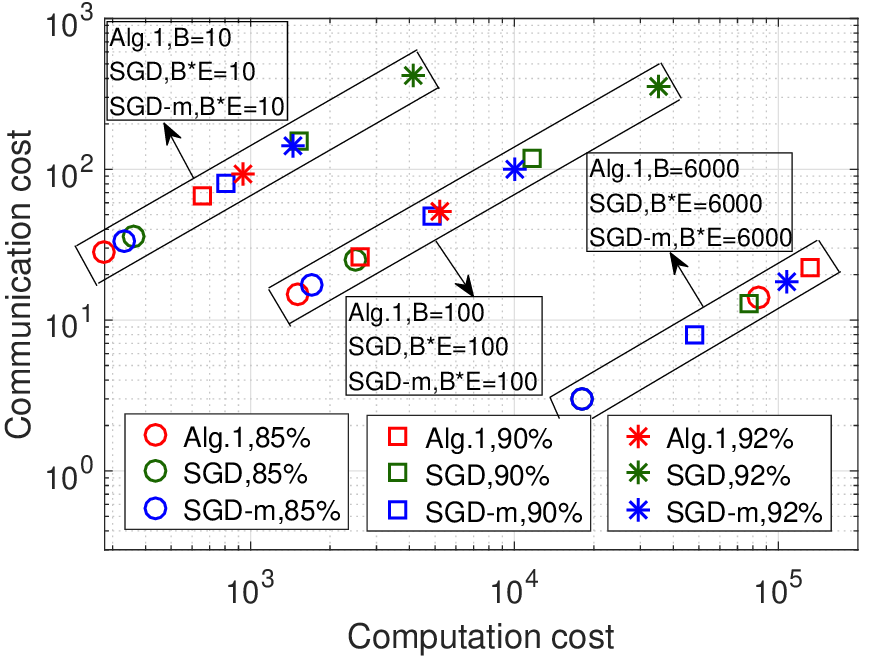}}}\quad
		\subfigure[\scriptsize{
			{Unconstrained feature-based FL.}}\label{fig:FCompComm}]
		{\resizebox{4.2cm}{!}{\includegraphics{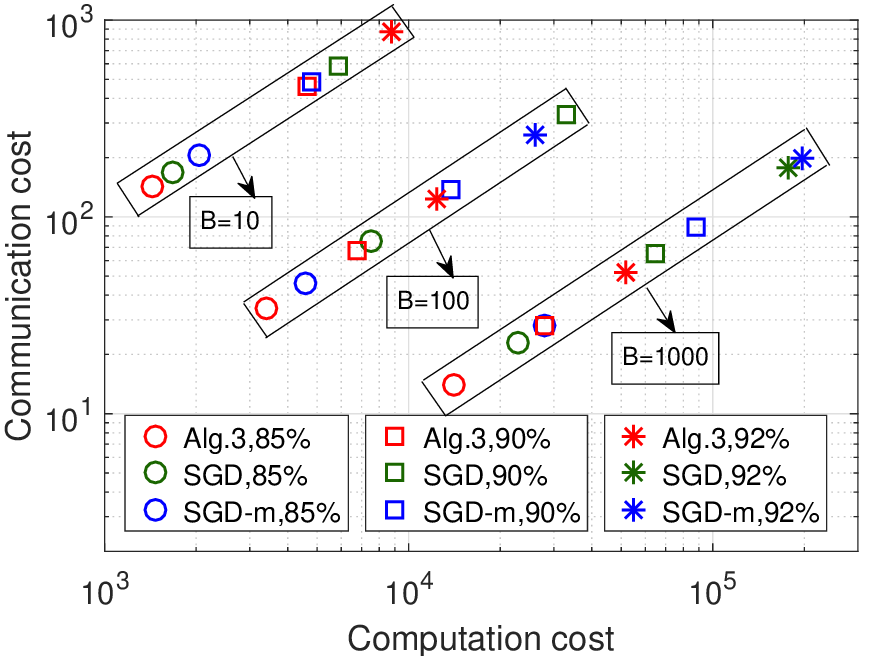}}}
	\end{center}
	\caption{\small{{Tradeoff between communication cost and computation cost for solving unconstrained federated optimization with a specific test accuracy.}}}
	\label{fig:CompComm}
\end{figure}

\begin{figure}[h]
	\begin{center}
		\subfigure[\scriptsize{
			$\ell_2$-norm  {$\Vert{\boldsymbol\omega}_{s}^{(T)}\Vert^2_2$} vs. training cost obtained by Algorithm~\ref{alg:uncon-sample}.}\label{fig:Tradeoff1}]
		{\resizebox{4.2cm}{!}{\includegraphics{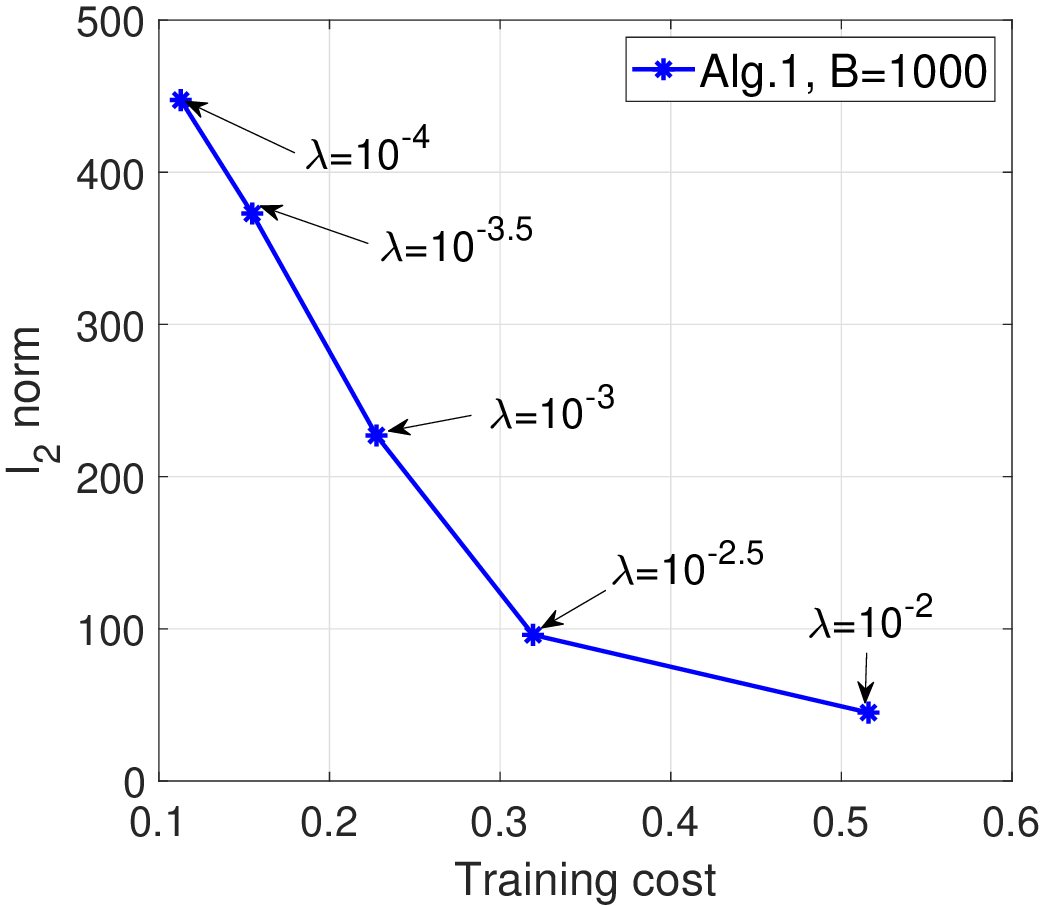}}}\quad
		\subfigure[\scriptsize{
			$\ell_2$-norm {$\Vert{\boldsymbol\omega}_{s}^{(T)}\Vert^2_2$} vs. training cost obtained by Algorithm~\ref{alg:con-sample}.}\label{fig:Tradeoff2}]
		{\resizebox{4.2cm}{!}{\includegraphics{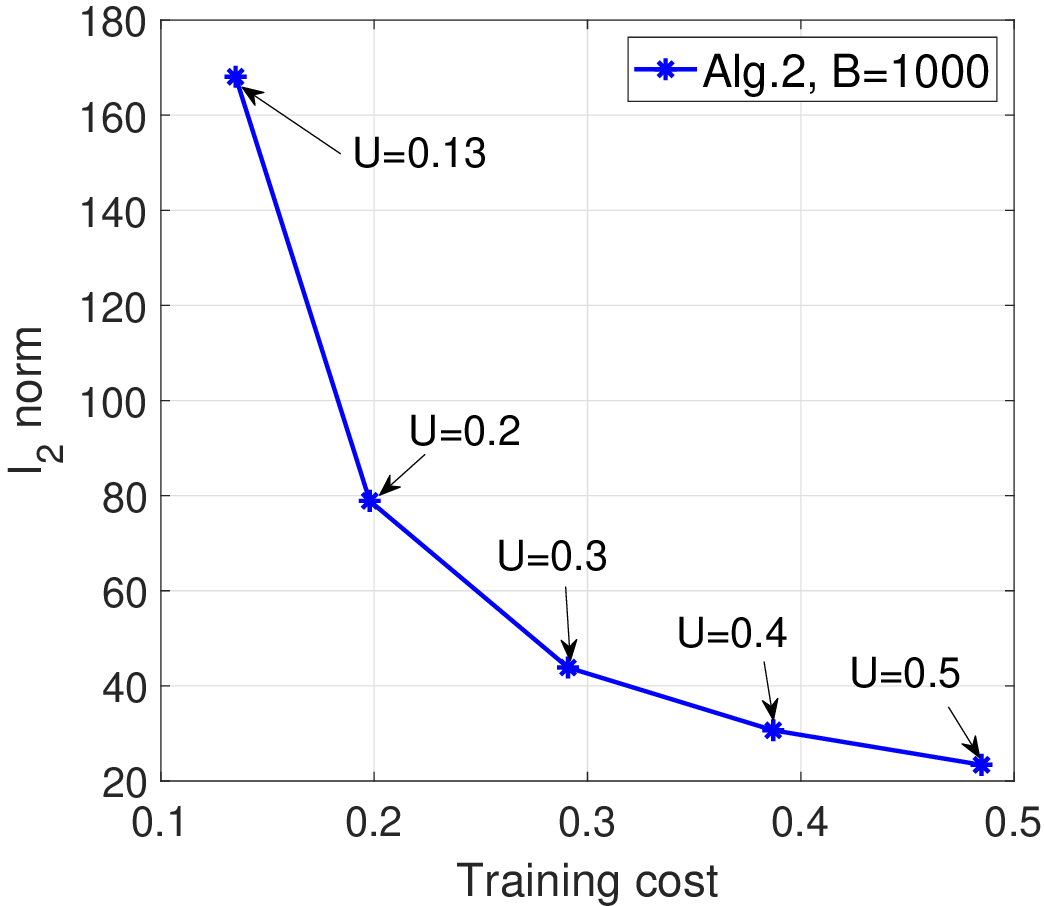}}}
	\end{center}
	\caption{\small{{Tradeoff between model sparsity and training cost {for sample-based FL} at $T=1000$.}}}
	\label{fig:Spar}
\end{figure}

%{Besides, note that an algorithm that converges faster has a lower communication cost for reaching a specific convergence performance, such as training cost and test accuracy. In each figure, if $B$ for each proposed algorithm and $B\times E$ for each sample-based ($B$ for each feature-based) baseline algorithm are equal, the two algorithms have the same order of computational complexity per communication round.}

Fig.~\ref{fig:Sam} and Fig.~\ref{fig:Fea} illustrate the training cost and test accuracy versus the {communication round} index in sample-based FL and feature-based FL, respectively. {From Fig.~\ref{fig:Sam} (a), (c), (e) and Fig.~\ref{fig:Fea} (a), (c), we can see that each proposed algorithm with larger  $B$, sample-based SGD (SGD-m) with larger  $B\times E$, and feature-based  SGD (SGD-m) with larger  $B$ converge faster at higher computation costs per communication round.
We can also observe that Algorithm~\ref{alg:uncon-sample} (Algorithm~\ref{alg:uncon-feature})  converges {faster than} all the baseline algorithms with the same order of computational complexity per communication round in most (all) cases. The only exception for Algorithm~\ref{alg:uncon-sample} is that in Fig.~\ref{fig:Sam} (c), Algorithm~\ref{alg:uncon-sample} with $B=6000$ converges slightly slower than sample-based SGD-m with $B=600$ and $E=10$.}

{Fig.~\ref{fig:CompComm} shows the tradeoff  between the communication and computation costs for solving unconstrained federated optimization. Here, the communication cost {of} each algorithm is measured by the number of communication rounds,  the computation costs {of} Algorithm~\ref{alg:uncon-sample}, Algorithm~\ref{alg:uncon-feature}, and feature-based  SGD (SGD-m) are measured by  $B$, and the communication cost {of} sample-based SGD (SGD-m) is measured by $B\times E$.
From Fig.~\ref{fig:SCompComm}, we see that the proposed algorithms achieve the best tradeoff between the communication cost and computation cost in all cases except the case where all local samples are utilized per communication round for solving for sample-based FL. Thus, Fig.~\ref{fig:SCompComm} indicates that Algorithm~\ref{alg:uncon-sample} (Algorithm~\ref{alg:uncon-feature})  achieves the lowest communication and computation costs for reaching a specific convergence performance  in most (all) cases.}

Fig.~\ref{fig:Spar} shows the tradeoff curve between the model sparsity and training cost of {each proposed algorithm for sample-based FL}. From Fig.~\ref{fig:Tradeoff2}, we see that with constrained sample-based federated optimization, one can set an explicit constraint on the training cost to control the test accuracy effectively.

\section{Conclusions}
In this paper, we investigated sample-based and feature-based federated optimization, respectively, and considered both the unconstrained problem and constrained problem for each of them. We proposed FL algorithms that converge to stationary points or KKT points using SSCA and mini-batch techniques. We also provided algorithm examples that have appealing computational complexities and communication loads per communication round and connect to FL algorithms via momentum SGD.
%We showed that data privacy could be preserved through the model aggregation mechanism and further enhanced via additional privacy mechanisms.
Numerical experiments demonstrated that the proposed {mini-batch} SSCA-based FL algorithms for unconstrained sample-based and feature-based federated optimization generally converge faster than existing FL algorithms, and the proposed mini-batch SSCA-based FL algorithms for constrained sample-based and feature-based federated optimization problems obtain models that strictly satisfy nonconvex constraints.
To the best of our knowledge, this is the first work that provides an SSCA framework for federated optimization, {highlights the value of constrained federated optimization, and establishes an analytical connection between SSCA and momentum SGD}. This paper opens up several directions for future research. An important direction is to design advanced SSCA-based FL algorithms that allow multiple local updates to reduce communication costs further. Another interesting direction is to design more privacy mechanisms for SSCA-based FL algorithms.

\section*{Appendix A: Proofs of Theorem~\ref{thm:uncon-sample} and Theorem~\ref{thm:uncon-feature}}
The proofs of Theorem~\ref{thm:uncon-sample} and Theorem~\ref{thm:uncon-feature} are identical. In the following proof, we omit the {subscripts} $s,f$ for notation simplicity.
First, we introduce the following preliminary results.
\begin{Lem}\label{lem:flim}
	%	Suppose $f_{0}$ satisfies Assumption~\ref{asump:f}, $\bar f_{0}$ satisfies Assumption~\ref{asump:fbar}, and the sequence $\{{\boldsymbol\omega}^{(t)}\}$ generated by Algorithm~\ref{alg:uncon-sample} (Algorithm~\ref{alg:uncon-feature}) is bounded.
	Let $\{{\boldsymbol\omega}^{(t)}\}$ be the sequence generated by Algorithm~\ref{alg:uncon-sample} (Algorithm~\ref{alg:uncon-feature}). Then, we have:
	\begin{align}
		%		&\lim_{t\to\infty}\vert{\bar{F}_{0}^{(t)}({\boldsymbol\omega}^{(t)})- F_{0}({\boldsymbol\omega}^{(t)})}\vert=0,\nonumber\\
		&\lim_{t\to\infty} \left\Vert{\nabla\bar{F}_{0}^{(t)}({\boldsymbol\omega}^{(t)})-\nabla F_{0}({\boldsymbol\omega}^{(t)})}\right\Vert_2=0,\nonumber\\
		&\lim_{t\to\infty}\left\vert{\bar{F}_{0}^{(t)}(\boldsymbol\omega)- G_{0}(\boldsymbol\omega;{\boldsymbol\omega}^{(t)})}\right\vert=0,\  \boldsymbol\omega\in\mathbb{R}^d,\nonumber
	\end{align}
	almost surely, where $G_{0}(\boldsymbol\omega;{\boldsymbol\omega}^{(t)})\triangleq\frac{1}{N}\sum_{n\in\mathcal{N}} \bar{f}_{0}(\boldsymbol\omega;{\boldsymbol\omega}^{(t)},\mathbf{x}_n)$.
\end{Lem}
\begin{IEEEproof}
	Lemma~\ref{lem:flim} is a consequence of~\cite[Lemma 1]{Lemma}. We only need to verify that all the technical conditions therein are satisfied. Specifically, Condition (a) of \cite[Lemma 1]{Lemma} is satisfied because $\{{\boldsymbol\omega}^{(t)}\}$ is assumed to be bounded. Condition (b) of \cite[Lemma 1]{Lemma} comes from Assumption~\ref{asump:fbar}.4. Conditions (c)-(d) of \cite[Lemma 1]{Lemma} come from the stepsize rules in \eqref{eqn:rho} and \eqref{eqn:gamma}. Condition (e) of \cite[Lemma 1]{Lemma} comes from the Lipschitz property of $F_{0}(\boldsymbol\omega)$ from Assumption~\ref{asump:f} and the stepsize rule in~\eqref{eqn:gamma}.
\end{IEEEproof}

\begin{Lem}\label{lem:et1t2}
	Let $\{{\boldsymbol\omega}^{(t)}\}$ be the sequence generated by Algorithm~\ref{alg:uncon-sample} (Algorithm~\ref{alg:uncon-feature}). Then, there exists a constant $\bar L$ such that
	\begin{align}
		\left\Vert{\bar{\boldsymbol\omega}^{(t_1)}-\bar{\boldsymbol\omega}^{(t_2)}}\right\Vert_2\leq\bar L\left\Vert{{\boldsymbol\omega}^{(t_1)}-{\boldsymbol\omega}^{(t_2)}}\right\Vert_2+e(t_1,t_2),\label{eqn:app1.0}
	\end{align}
	and $\lim_{t_1,t_2\to\infty}e(t_1,t_2)=0$ almost surely.
\end{Lem}

\begin{IEEEproof}
	It follows from Lemma~\ref{lem:flim} that
	\begin{align} \bar{F}_0^{(t)}(\boldsymbol\omega)=G_0(\boldsymbol\omega;{\boldsymbol\omega}^{(t)})+\bar{e}(t),\label{eqn:app1.l1}
	\end{align}
	where $\bar{e}(t)$ satisfies $\lim_{t\to\infty}\bar{e}(t)=0$. From Assumption~\ref{asump:fbar}.3, $G_0(\boldsymbol\omega;{\boldsymbol\omega}^{(t)})$ is Lipschitz continuous in ${\boldsymbol\omega}^{(t)}$ and thus
	\begin{align} \left\vert{G_0(\boldsymbol\omega;{\boldsymbol\omega}^{(t_1)})-G_0(\boldsymbol\omega;{\boldsymbol\omega}^{(t_2)})}\right\vert
\!\leq\! \tilde{L}\left\Vert{{\boldsymbol\omega}^{(t_1)}\!-\!{\boldsymbol\omega}^{(t_2)}}\right\Vert_2\!,\boldsymbol\omega\in\mathbb{R}^d,\label{eqn:app1.l2}
	\end{align}
	for some constant $\tilde{L}>0$. Combining~\eqref{eqn:app1.l1} and~\eqref{eqn:app1.l2}, we have:
	\begin{align} \left\vert{\bar{F}_0^{(t_1)}(\boldsymbol\omega)\!-\!\bar{F}_0^{(t_2)}(\boldsymbol\omega)}\right\vert
\!\leq\!\tilde{L} \left\Vert{{\boldsymbol\omega}^{(t_1)}\!-\!{\boldsymbol\omega}^{(t_2)}}\right\Vert_2
\!\!+\!\tilde{e}(t_1,t_2), \boldsymbol\omega\in\mathbb{R}^d, \label{eqn:app1.1}
	\end{align}
	where $\tilde{e}(t_1,t_2)$ satisfies $\lim_{t_1,t_2\to\infty}\tilde{e}(t_1,t_2)=0$.
	From Assumption~\ref{asump:fbar}.3, there exists constant $\mu>0$ such that for all $t=1,2,\cdots,\infty$, $\bar{F}_0^{(t)}(\boldsymbol\omega)$ is strongly convex with $\mu$. Due to the strong convexity of $\bar{F}_0^{(t_1)}(\boldsymbol\omega)$ and the optimality of $\bar{\boldsymbol\omega}^{(t_1)}$, we have:
	\begin{align} \bar{F}_0^{(t_1)}({\boldsymbol\omega})-\bar{F}_0^{(t_1)}(\bar{\boldsymbol\omega}^{(t_1)})\geq \frac{\mu}{2}\left\Vert{{\boldsymbol\omega}-\bar{\boldsymbol\omega}^{(t_1)}}\right\Vert_2, \ \boldsymbol\omega\in\mathbb{R}^d. \label{eqn:app1.2}
	\end{align}
	Setting $\boldsymbol\omega=\bar{\boldsymbol\omega}^{(t_2)}$ in \eqref{eqn:app1.2}, we have:
	\begin{align}
		\bar{F}_0^{(t_1)}(\bar{\boldsymbol\omega}^{(t_2)})-\bar{F}_0^{(t_1)}(\bar{\boldsymbol\omega}^{(t_1)})\geq \frac{\mu}{2} \left\Vert{\bar{\boldsymbol\omega}^{(t_2)}-\bar{\boldsymbol\omega}^{(t_1)}}\right\Vert_2. \label{eqn:app1.3}
	\end{align}
	Similarly, by the strong convexity of $\bar{F}_0^{(t_2)}(\boldsymbol\omega)$ and the optimality of $\bar{\boldsymbol\omega}^{(t_2)}$, we have:
	\begin{align}
		\bar{F}_0^{(t_2)}(\bar{\boldsymbol\omega}^{(t_1)})-\bar{F}_0^{(t_2)}(\bar{\boldsymbol\omega}^{(t_2)})\geq \frac{\mu}{2} \left\Vert{\bar{\boldsymbol\omega}^{(t_1)}-\bar{\boldsymbol\omega}^{(t_2)}}\right\Vert_2. \label{eqn:app1.4}
	\end{align}
	Thus, we have:
	\begin{align} &\left\Vert{\bar{\boldsymbol\omega}^{(t_1)}-\bar{\boldsymbol\omega}^{(t_2)}}\right\Vert_2\nonumber\\
		\!\overset{(a)}{\leq}\!&\frac{1}{\mu}\!\!\left( \left\vert{\!\bar{F}_0^{(t_1)}\!(\bar{\boldsymbol\omega}^{(t_1)}\!)\!-\!\bar{F}_0^{(t_2)}\!(\bar{\boldsymbol\omega}^{(t_1)}\!)}\!\right\vert\!+\!\left\vert{\!\bar{F}_0^{(t_1)}\!(\bar{\boldsymbol\omega}^{(t_2)}\!)\!-\!\bar{F}_0^{(t_2)}\!(\bar{\boldsymbol\omega}^{(t_2)}\!)\!}\right\vert\right)\nonumber\\
		\overset{(b)}{\leq} &\frac{2\tilde{L}}{\mu} \left\Vert{{\boldsymbol\omega}^{(t_1)}-{\boldsymbol\omega}^{(t_2)}}\right\Vert_2+\frac{2}{\mu}\tilde{e}(t_1,t_2),\label{eqn:app1.5}
	\end{align}
	where $(a)$ follows from~\eqref{eqn:app1.3} and \eqref{eqn:app1.4}, and $(b)$ follows from~\eqref{eqn:app1.1}.
	Finally, \eqref{eqn:app1.0} follows from \eqref{eqn:app1.5} immediately.
\end{IEEEproof}

\begin{Lem}\label{lem:51}
	Let $\{{\boldsymbol\omega}^{(t)}\}$ be the sequence generated by Algorithm~\ref{alg:uncon-sample} (Algorithm~\ref{alg:uncon-feature}). Then, we have:
	\begin{align}
		&F_0({\boldsymbol\omega}^{(t+1)})-F_0({\boldsymbol\omega}^{(t)})\nonumber\\
		\leq&\gamma^{(t)} \left\Vert\bar{\boldsymbol\omega}^{(t)}-{\boldsymbol\omega}^{(t)}\right\Vert_2 \left\Vert\nabla F_0({\boldsymbol\omega}^{(t)})-\nabla \bar{F}_0^{(t)}({\boldsymbol\omega}^{(t)})\right\Vert_2\nonumber\\ &-\gamma^{(t)}\left(\mu-\frac{\hat{L}}{2}\gamma^{(t)}\right)\left\Vert\bar{\boldsymbol\omega}^{(t)}-{\boldsymbol\omega}^{(t)}\right\Vert_2^2.\label{eqn:app1.7}
	\end{align}
\end{Lem}

\begin{IEEEproof}
	From Assumption~\ref{asump:fbar}.2, $\bar{F}_0^{(t)}({\boldsymbol\omega})$ is uniformly strongly convex, and thus:
	\begin{align} &{(\bar{\boldsymbol\omega}^{(t)}-{\boldsymbol\omega}^{(t)})^T\nabla \bar{F}_0^{(t)}({\boldsymbol\omega}^{(t)})}\nonumber\\
		\leq&
		-\mu\left\Vert{\bar{\boldsymbol\omega}^{(t)}-{\boldsymbol\omega}^{(t)}}\right\Vert_2+\bar{F}_0^{(t)}(\bar{\boldsymbol\omega}^{(t)})
		-\bar{F}_0^{(t)}({\boldsymbol\omega}^{(t)})\nonumber\\
		\leq&-\mu\left\Vert{\bar{\boldsymbol\omega}^{(t)}-{\boldsymbol\omega}^{(t)}}\right\Vert_2,\label{eqn:app1.6}
	\end{align}
	where the last inequality follows from the optimality of $\bar{\boldsymbol\omega}^{(t)}$.
	Suppose $\nabla F_0(\boldsymbol\omega)$ is Lipschitz continuous with constant $\hat{L}>0$, we have:
	\begin{align}	&F_0({\boldsymbol\omega}^{(t+1)})-F_0({\boldsymbol\omega}^{(t)})\nonumber\\
\leq&({\boldsymbol\omega}^{(t+1)}-{\boldsymbol\omega}^{(t)})^T \nabla F_0({\boldsymbol\omega}^{(t)})+\frac{\hat{L}}{2}\left\Vert{\boldsymbol\omega}^{(t+1)}-{\boldsymbol\omega}^{(t)}\right\Vert_2^2\nonumber\\ \leq&\gamma^{(t)}(\bar{\boldsymbol\omega}^{(t)}-{\boldsymbol\omega}^{(t)})^T \nabla F_0({\boldsymbol\omega}^{(t)})+\frac{\hat{L}}{2}(\gamma^{(t)})^2\left\Vert\bar{\boldsymbol\omega}^{(t)}-{\boldsymbol\omega}^{(t)}\right\Vert_2^2\nonumber\\	\leq&\gamma^{(t)}(\bar{\boldsymbol\omega}^{(t)}\!-\!{\boldsymbol\omega}^{(t)})^T
\!\!\left(\!\nabla F_0({\boldsymbol\omega}^{(t)})\!-\!\nabla \bar{F}_0^{(t)}\!({\boldsymbol\omega}^{(t)})\!+\!\nabla \bar{F}_0^{(t)}\!({\boldsymbol\omega}^{(t)})\!\right)\nonumber\\
		&+\frac{\hat{L}}{2}(\gamma^{(t)})^2\left\Vert\bar{\boldsymbol\omega}^{(t)}-{\boldsymbol\omega}^{(t)}\right\Vert_2^2\nonumber\\
		\leq&\gamma^{(t)} \left\Vert\bar{\boldsymbol\omega}^{(t)}-{\boldsymbol\omega}^{(t)}\right\Vert_2 \left\Vert\nabla F_0({\boldsymbol\omega}^{(t)})-\nabla \bar{F}_0^{(t)}({\boldsymbol\omega}^{(t)})\right\Vert_2\nonumber\\	&-\gamma^{(t)}\left(\mu-\frac{\hat{L}}{2}\gamma^{(t)}\right)\left\Vert\bar{\boldsymbol\omega}^{(t)}-{\boldsymbol\omega}^{(t)}\right\Vert_2^2,
	\end{align}
	where the last inequality follows form \eqref{eqn:app1.6}.
\end{IEEEproof}

%In the following, we show $\lim_{t\to\infty}\Vert\bar{\boldsymbol\omega}^{(t)}-{\boldsymbol\omega}^{(t)}\Vert=0$ by showing that $\lim\inf_{t\to\infty}\Vert\bar{\boldsymbol\omega}^{(t)}-{\boldsymbol\omega}^{(t)}\Vert=\lim\sup{t\to\infty}\Vert\bar{\boldsymbol\omega}^{(t)}-{\boldsymbol\omega}^{(t)}\Vert=0$.
Then, we show by contradiction that $\lim\inf_{t\to\infty}\left\Vert\bar{\boldsymbol\omega}^{(t)}-{\boldsymbol\omega}^{(t)}\right\Vert_2=0$ almost surely.
Suppose $\lim\inf_{t\to\infty}\left\Vert\bar{\boldsymbol\omega}^{(t)}-{\boldsymbol\omega}^{(t)}\right\Vert_2\geq\chi>0$ with a positive probability. Then we can find a realization such that $\left\Vert\bar{\boldsymbol\omega}^{(t)}-{\boldsymbol\omega}^{(t)}\right\Vert_2\geq\chi>0$ for all $t$. We focus next on such a realization. By $\left\Vert\bar{\boldsymbol\omega}^{(t)}-{\boldsymbol\omega}^{(t)}\right\Vert_2\geq\chi>0$ and Lemma~\ref{lem:51}, we have:
\begin{align}
	&F_0({\boldsymbol\omega}^{(t+1)})-F_0({\boldsymbol\omega}^{(t)})\nonumber\\
	\leq&-\gamma^{(t)}\left(\mu-\frac{\hat{L}}{2}\gamma^{(t)}-\frac{1}{\chi}
\left\Vert{\nabla F_0({\boldsymbol\omega}^{(t)})-\nabla \bar{F}_0^{(t)}({\boldsymbol\omega}^{(t)})}\right\Vert_2\right)\nonumber\\
	&\times\left\Vert\bar{\boldsymbol\omega}^{(t)}-{\boldsymbol\omega}^{(t)}\right\Vert_2^2.\label{eqn:app1.8}
\end{align}
Since $\lim_{t\to\infty}\left\Vert{\nabla\bar{F}_0^{(t)}({\boldsymbol\omega}^{(t)})-\nabla F_0({\boldsymbol\omega}^{(t)})}\right\Vert_2=0$, $\lim_{t\to\infty}\gamma^{(t)}$ and $\mu>0$, there exists a $t_0$ sufficiently large such that
\begin{align} \mu\!-\!\frac{\hat{L}}{2}\gamma^{(t)}\!-\!\frac{1}{\chi}\left\Vert{\nabla F_0({\boldsymbol\omega}^{(t)})\!-\!\nabla \bar{F}_0^{(t)}({\boldsymbol\omega}^{(t)})}\right\Vert_2\geq\bar\mu,\  \forall t\geq t_0,\label{eqn:app1.9}
\end{align}
for some $\bar\mu\in(0,\mu)$.
Therefore, it follows from \eqref{eqn:app1.8}, \eqref{eqn:app1.9} and  $\left\Vert\bar{\boldsymbol\omega}^{(t)}-{\boldsymbol\omega}^{(t)}\right\Vert_2\geq\chi$ for all $t$ that
\begin{align} F_0({\boldsymbol\omega}^{(t)})-F_0({\boldsymbol\omega}^{(t_0)})\leq-\bar\mu\chi^2\sum_{n=t_0}^{(t)}\gamma^{(t)},\label{eqn:app1.10}
\end{align}
which, in view of $\sum_{n=t_0}^{\infty}\gamma^{(t)}=\infty$, contradicts the boundedness of $\{F_0({\boldsymbol\omega}^{(t)})\}$. Therefore, it must be $\lim\inf_{t\to\infty}\left\Vert\bar{\boldsymbol\omega}^{(t)}-{\boldsymbol\omega}^{(t)}\right\Vert_2=0$ almost surely.

Next, we show by contradiction that
$\lim\sup_{t\to\infty}\left\Vert\bar{\boldsymbol\omega}^{(t)}-{\boldsymbol\omega}^{(t)}\right\Vert_2=0$ almost surely. Suppose $\lim\sup_{t\to\infty}\left\Vert\bar{\boldsymbol\omega}^{(t)}-{\boldsymbol\omega}^{(t)}\right\Vert_2>0$ with a positive probability. We focus next on a realization along with $\lim\sup_{t\to\infty}\left\Vert\bar{\boldsymbol\omega}^{(t)}-{\boldsymbol\omega}^{(t)}\right\Vert_2>0$, $\lim\inf_{t\to\infty}\left\Vert\bar{\boldsymbol\omega}^{(t)}-{\boldsymbol\omega}^{(t)}\right\Vert_2=0$, and $\lim_{t_1,t_2\to\infty}e(t_1,t_2)=0$, where $e(t_1,t_2)$ is defined in Lemma~\ref{lem:et1t2}. It follows from $\lim\sup_{t\to\infty}\left\Vert\bar{\boldsymbol\omega}^{(t)}-{\boldsymbol\omega}^{(t)}\right\Vert_2>0$ and $\lim\inf_{t\to\infty}\left\Vert\bar{\boldsymbol\omega}^{(t)}-{\boldsymbol\omega}^{(t)}\right\Vert_2=0$ that there exists a $\delta>0$ such that $\left\Vert\Delta{\boldsymbol\omega}^{(t)}\right\Vert_2\geq2\delta$ (with $\Delta{\boldsymbol\omega}^{(t)}\triangleq\bar{\boldsymbol\omega}^{(t)}-\boldsymbol\omega$) for infinitely many $t$ and also $\left\Vert\Delta{\boldsymbol\omega}^{(t)}\right\Vert_2\leq\delta$ for infinitely many $t$. Therefore, one can always find an infinite set of indices, say $\mathcal{T}$, having the following properties: for any $t\in\mathcal{T}$, we have:
\begin{align} \left\Vert\Delta{\boldsymbol\omega}^{(t)}\right\Vert_2\leq\delta,\label{eqn:app1.111}
\end{align}
and there exists an integer $i_t>t$ such that
\begin{align}
	\left\Vert\Delta{\boldsymbol\omega}^{(i_t)}\right\Vert_2\geq2\delta,\ \delta\leq\left\Vert\Delta{\boldsymbol\omega}^{(n)}\right\Vert_2\leq2\delta, \ t<n<i_t.\label{eqn:app1.11}
\end{align}
Thus, for all $t\in\mathcal{T}$, we have:
\begin{align} \delta&\leq\left\Vert\Delta{\boldsymbol\omega}^{(i_t)}\right\Vert_2-\left\Vert\Delta{\boldsymbol\omega}^{(t)}\right\Vert_2 \leq\left\Vert\Delta{\boldsymbol\omega}^{(i_t)}-\Delta{\boldsymbol\omega}^{(t)}\right\Vert_2\nonumber\\ &=\left\Vert(\bar{\boldsymbol\omega}^{(i_t)}-{\boldsymbol\omega}^{(i_t)})-(\bar{\boldsymbol\omega}^{(t)}-{\boldsymbol\omega}^{(t)})\right\Vert_2\nonumber\\ &\leq\left\Vert{\bar{\boldsymbol\omega}^{(i_t)}-\bar{\boldsymbol\omega}^{(t)}}\right\Vert_2+\left\Vert{{\boldsymbol\omega}^{(i_t)}-{\boldsymbol\omega}^{(t)}}\right\Vert_2\nonumber\\
&\overset{(a)}{\leq}(1+\bar L)\left\Vert{{\boldsymbol\omega}^{(i_t)}-{\boldsymbol\omega}^{(t)}}\right\Vert_2+e(i_t,t)\nonumber\\
&\overset{(b)}{\leq}(1+\bar L)\sum_{n=t}^{i_t-1}\gamma^{(n)}\left\Vert\Delta{\boldsymbol\omega}^{(n)}\right\Vert_2+e(i_t,t)\nonumber\\
&\leq2\delta(1+\bar L)\sum_{n=t}^{i_t-1}\gamma^{(n)}+e(i_t,t),\label{eqn:app1.12}
\end{align}
where $(a)$ is due to Lemma~\ref{lem:et1t2}, and $(b)$ is due to~\eqref{eqn:app1.111} and~\eqref{eqn:app1.11}.
By~\eqref{eqn:app1.12} and $\lim_{t\to\infty}e(i_t,t)=0$, we have:
\begin{align}
	\mathop{\lim\inf}_{\mathcal{T}\ni t\to\infty}\sum_{n=t}^{i_t-1}\gamma^{(t)}\geq\delta_1\triangleq\frac{1}{2(1+\bar L)}>0.\label{eqn:app1.13}
\end{align}
Proceeding as in~\eqref{eqn:app1.12},  for all $t\in\mathcal{T}$, we also have:
\begin{align}
	&\left\Vert\Delta{\boldsymbol\omega}^{(t+1)}\Vert-\Vert\Delta{\boldsymbol\omega}^{(t)}\right\Vert_2
	\leq\left\Vert\Delta{\boldsymbol\omega}^{(t+1)}-\Delta{\boldsymbol\omega}^{(t)}\right\Vert_2\nonumber\\
	\leq&(1+\bar L)\gamma^{(t)}\left\Vert\Delta{\boldsymbol\omega}^{(t)}\right\Vert_2+e(t,t+1),\label{eqn:app1.14}
\end{align}
which leads to
\begin{align}
	(1\!+\!(1\!+\!\bar L)\gamma^{(t)})\left\Vert\Delta{\boldsymbol\omega}^{(t)}\right\Vert_2+e(t,t+1)
	\geq\left\Vert\Delta{\boldsymbol\omega}^{(t+1)}\right\Vert_2\geq\delta,\label{eqn:app1.15}
\end{align}
where the second inequality follows from~\eqref{eqn:app1.11}. It follows from~\eqref{eqn:app1.15} and $\lim_{t\to\infty}e(t,t+1)=0$ that there exists a $\delta_2>0$ such that for a sufficiently large $t\in\mathcal{T}$,
\begin{align}
	\left\Vert\Delta{\boldsymbol\omega}^{(t)}\right\Vert_2\geq\frac{\delta-e(t,t+1)}{1+(1+\bar L)\gamma^{(t)}}\geq\delta_2>0. \label{eqn:app1.16}
\end{align}
Here after we assume w.l.o.g. that~\eqref{eqn:app1.16} holds for all $t\in\mathcal{T}$ (in fact one can always restrict $\{{\boldsymbol\omega}^{(t)}\}_{t\in\mathcal{T}}$ to a proper subsequence).
We show now that~\eqref{eqn:app1.13} is in contradiction with the convergence of $\{F_0({\boldsymbol\omega}^{(t)})\}$. By Lemma~\ref{lem:51}, for all $t\in\mathcal{T}$, we have:
\begin{align}
	&F_0({\boldsymbol\omega}^{(t+1)})-F_0({\boldsymbol\omega}^{(t)}) \!\leq\!-\gamma^{(t)}\!\left(\!\mu-\frac{\hat{L}}{2}\gamma^{(t)}\!\right)\!\left\Vert\bar{\boldsymbol\omega}^{(t)}-{\boldsymbol\omega}^{(t)}\right\Vert_2^2\nonumber\\
	&+\gamma^{(t)} \delta \left\Vert\nabla F_0({\boldsymbol\omega}^{(t)})-\nabla \bar{F}_0^{(t)}({\boldsymbol\omega}^{(t)})\right\Vert_2,\label{eqn:app1.17}
\end{align}
and for $t<n<i_t$,
\begin{small}\begin{align} &F_0({\boldsymbol\omega}^{(n+1)})-F_0({\boldsymbol\omega}^{(n)})\nonumber\\
	\leq&\!-\!\gamma^{(n)}\!\!\!\left(\!\!\mu\!-\!\frac{\hat{L}}{2}\gamma^{(n)}\!\!-\!\frac{\left\Vert\nabla F_0({\boldsymbol\omega}^{(n)})\!-\!\nabla \bar{F}_0^{(n)}\!(\!{\boldsymbol\omega}^{(n)})\right\Vert_2}{\!\left\Vert\bar{\boldsymbol\omega}^{(n)}-{\boldsymbol\omega}^{(n)}\right\Vert_2}\!\!\right)\!\!\left\Vert\bar{\boldsymbol\omega}^{(n)}\!\!\!-\!{\boldsymbol\omega}^{(n)}\!\right\Vert_2^2\nonumber\\
	\leq&\!-\!\gamma^{(n)}\!\!\!\left(\!\!\mu\!-\!\frac{\hat{L}}{2}\gamma^{(n)}\!\!-\!\frac{\!\left\Vert\nabla F_0({\boldsymbol\omega}^{(n)})\!-\!\nabla \bar{F}_0^{(n)}\!(\!{\boldsymbol\omega}^{(n)})\!\right\Vert_2}{\delta}\!\!\right)\!\!\left\Vert\bar{\boldsymbol\omega}^{(n)}\!\!\!-\!{\boldsymbol\omega}^{(n)}\!\right\Vert_2^2,\label{eqn:app1.18}
\end{align}\end{small}where the second inequality follows from~\eqref{eqn:app1.11}. Adding \eqref{eqn:app1.17} and \eqref{eqn:app1.18} over $n=t+1,\cdots,i_t-1$ and, for $t\in\mathcal{T}$ sufficiently large (so that $\mu-\frac{\hat{L}}{2}\gamma^{(t)}-\delta^{-1}\left\Vert\nabla F_0({\boldsymbol\omega}^{(t)})-\nabla \bar{F}_0^{(t)}({\boldsymbol\omega}^{(t)})\right\Vert_2\geq\hat\mu>0$ and $\left\Vert\nabla F_0({\boldsymbol\omega}^{(t)})-\nabla \bar{F}_0^{(t)}({\boldsymbol\omega}^{(t)})\right\Vert_2<\hat\mu\delta_2^2\delta^{-1}$), we have:
\begin{small}\begin{align} &F_0({\boldsymbol\omega}^{(i_t)})-F_0({\boldsymbol\omega}^{(t)})\nonumber\\
	\overset{(a)}{\leq}&\!-\!\hat\mu\!\sum\nolimits_{n=t}^{i_t-1}\!\!\gamma^{(n)}\!\left\Vert\bar{\boldsymbol\omega}^{(n)}\!\!-\!{\boldsymbol\omega}^{(n)}\!\right\Vert_2^2\!\!+\!\gamma^{(t)}\delta \left\Vert\nabla F_0({\boldsymbol\omega}^{(t)}\!)\!-\!\nabla \bar{F}_0^{(t)}({\boldsymbol\omega}^{(t)}\!)\right\Vert_2\nonumber\\
	\overset{(b)}{\leq}&\!-\!\hat\mu\delta_2^2\sum_{n=t+1}^{i_t-1}\gamma^{(n)}\!-\!\gamma^{(t)}\!\!\left(\hat\mu\delta_2^2-\delta\left\Vert\nabla F_0({\boldsymbol\omega}^{(t)})-\nabla \bar{F}_0^{(t)}({\boldsymbol\omega}^{(t)})\right\Vert_2\right)\nonumber\\
	\overset{(c)}{\leq}&\!-\!\hat\mu\delta_2^2\sum_{n=t+1}^{i_t-1}\gamma^{(n)},\label{eqn:app1.19}
\end{align}\end{small}where $(a)$ follows from $\mu-\frac{\hat{L}}{2}\gamma^{(t)}-\delta^{-1}\left\Vert\nabla F_0({\boldsymbol\omega}^{(t)})-\nabla \bar{F}_0^{(t)}({\boldsymbol\omega}^{(t)})\right\Vert_2\geq\hat\mu>0$;
$(b)$ follows from~\eqref{eqn:app1.16}; and $(c)$ follows from $\left\Vert\nabla F_0({\boldsymbol\omega}^{(t)})-\nabla \bar{F}_0^{(t)}({\boldsymbol\omega}^{(t)})\right\Vert_2<\hat\mu\delta_2^2\delta^{-1}$. Since $\{F_0({\boldsymbol\omega}^{(t)})\}$ converges, it must be $\mathop{\lim\inf}_{\mathcal{T}\ni t\to\infty}\sum_{n=t+1}^{i_t-1}\gamma^{(t)}=0$, which contradicts~\eqref{eqn:app1.13}. Therefore, it must be $\lim\sup_{t\to\infty}\left\Vert\bar{\boldsymbol\omega}^{(t)}-{\boldsymbol\omega}^{(t)}\right\Vert_2=0$ almost surely.

Finally, we show that a limit point of the sequence $\{{\boldsymbol\omega}^{(t)}\}$ generated by Algorithm~\ref{alg:uncon-sample} (Algorithm~\ref{alg:uncon-feature}), i.e., ${\boldsymbol\omega}^\star$, is a {stationary} point of Problem~\ref{Prob:uncon-sample} (Problem~\ref{Prob:uncon-feature}). It follows from first-order optimality condition for $\bar{\boldsymbol\omega}^{(t)}$ that\begin{align}
	(\boldsymbol\omega-\bar{\boldsymbol\omega}^{(t)})^T \nabla \bar{F}_0^{(t)}({\boldsymbol\omega}^{(t)})\geq0, \  \boldsymbol\omega\in\mathbb{R}^d.\label{eqn:app1.20}
\end{align}
Taking the limit of \eqref{eqn:app1.20} over the index set $\mathcal{T}$, we have:
\begin{small}\begin{align}
&\lim_{\mathcal{T}\ni t\to\infty}\!\!\!(\boldsymbol\omega-\bar{\boldsymbol\omega}^{(t)}\!)^T \nabla \bar{F}_0^{(t)}\!(\bar{\boldsymbol\omega}^{(t)}\!) \!=\!(\boldsymbol\omega-{\boldsymbol\omega}^\star\!)^T\nabla F_0({\boldsymbol\omega}^\star)\!\geq\!0,\ \boldsymbol\omega\!\in\!\mathbb{R}^d,\nonumber%\label{eqn:app1.21}
\end{align}\end{small}where the equality follows from $\lim_{t\to\infty}\left\Vert\bar{\boldsymbol\omega}^{(t)}-{\boldsymbol\omega}^{(t)}\right\Vert_2=0$ (which is due to $\lim\inf{t\to\infty}\left\Vert\bar{\boldsymbol\omega}^{(t)}-{\boldsymbol\omega}^{(t)}\right\Vert_2=0$ and $\lim\sup_{t\to\infty}\left\Vert\bar{\boldsymbol\omega}^{(t)}-{\boldsymbol\omega}^{(t)}\right\Vert_2=0$) and $\lim_{t\to\infty}\left\Vert\nabla F_0({\boldsymbol\omega}^{(t)})-\nabla \bar{F}_0^{(t)}({\boldsymbol\omega}^{(t)})\right\Vert_2=0$. This is the desired first-order optimality condition and ${\boldsymbol\omega}^\star$ is a {stationary} point of Problem~\ref{Prob:uncon-sample} (Problem~\ref{Prob:uncon-feature}).

%\section*{{Appendix B: Proofs of Lemma~\ref{lem:con-sample-ep} and Lemma~\ref{lem:con-feature-ep}}}

\section*{Appendix B: Proofs of Theorem~\ref{thm:con-sample} and Theorem~\ref{thm:con-feature}}
The proofs of Theorem~\ref{thm:con-sample} and Theorem~\ref{thm:con-feature} are identical. In the following proof, we omit the {subscripts} $s,f$ for notation simplicity.
We first introduce the following preliminary results.
{\begin{Lem}\label{lem:con-ap-ep}
%	Suppose that $f_{s,m}$, $m=0,\cdots,M$ satisfy Assumption~\ref{asump:f}, $\bar{f}_{s,0}$ satisfies Assumption~\ref{asump:fbar}, $\bar{f}_{s,m}$ satisfies $\bar{f}_{s,m}(\boldsymbol\omega;\boldsymbol\omega,\mathbf{x})=f_{s,m}(\boldsymbol\omega;\mathbf{x})$ and Assumption~\ref{asump:fbar} for all $m=1,\cdots,M$, the sequence $\{{\boldsymbol\omega}_{s}^{(t)}\}$ generated by Algorithm~\ref{alg:con-sample} with $c=c_j$ is bounded for all $j$, and the sequence $\{c_j\}$ satisfies $0<c_j<c_{j+1}$ and $\lim_{j\to\infty}c_j=\infty$.
	Let	$({\boldsymbol\omega}_{j}^\star,\mathbf s_{j}^\star)$ denote a KKT point of Problem~\ref{Prob:con-sample-ep} (Problem~\ref{Prob:con-feature-ep}) with $c=c_j$ and let  $({\boldsymbol\omega}_{\infty}^\star,\mathbf s_{\infty}^\star)$
 denote a limit point of $\{({\boldsymbol\omega}_{j}^\star,\mathbf s_{j}^\star)\}$.  Then, the following statements hold.
	i) For all $j$, if $\mathbf s_{s,j}^\star=\mathbf0$, then ${\boldsymbol\omega}_{s,j}^\star$ is a KKT point of  Problem~\ref{Prob:con-sample} (Problem~\ref{Prob:con-feature});
	ii) $\mathbf s_{\infty}^\star=\mathbf{0}$, and ${\boldsymbol\omega}_{\infty}^\star$ is a KKT point of Problem~\ref{Prob:con-sample} (Problem~\ref{Prob:con-feature}).
\end{Lem}}
{
\begin{IEEEproof} i) The KKT conditions   of Problem~\ref{Prob:con-sample-ep} (Problem~\ref{Prob:con-feature-ep}) with $c=c_j$ are given by:
\begin{subequations}\label{eqn:KKT1}
\begin{align}
&F_m({\boldsymbol\omega}^\star_j)\leq s_{m,j}^\star,\ s_{m,j}^\star\geq 0, \ m=1,2,\cdots,M,\\
&\lambda_m(F_m({\boldsymbol\omega}^\star_j)\!-\!s_{m,j}^\star)\!\!=\!0, \mu_m s_{m,j}^\star\!=\!0,\ m=1,\cdots,M,\\ &\nabla_{\boldsymbol\omega}F_0({\boldsymbol\omega}^\star_j)+\sum_{m=1}^M\lambda_m\nabla_{\boldsymbol\omega}F_m({\boldsymbol\omega}^\star_j)=0,\\
&c_j-\lambda_m-\mu_m=0,\ m=1,\cdots,M.
\end{align}
\end{subequations}
On the other hand,  the KKT conditions   of Problem~\ref{Prob:con-sample} (Problem~\ref{Prob:con-feature}) are given by:
\begin{subequations}\label{eqn:KKT2}
\begin{align}
&F_m({\boldsymbol\omega}^\star_j)\leq 0,\ m=1,2,\cdots,M,\\
&\lambda_m(F_m({\boldsymbol\omega}^\star_j)\!-\!0)\!=\!0,\ m=1,\cdots,M,\\	&\nabla_{\boldsymbol\omega}F_0({\boldsymbol\omega}^\star_j)+\sum_{m=1}^M\lambda_m\nabla_{\boldsymbol\omega}F_m({\boldsymbol\omega}^\star_j)=0.
\end{align}
\end{subequations}
As \eqref{eqn:KKT1}   with $\mathbf s_{s,j}^\star=\mathbf0$ implies \eqref{eqn:KKT2}, we can show the first statement. ii) Construct a convex approximation of Problem~\ref{Prob:con-sample-ep} (Problem~\ref{Prob:con-feature-ep}) with $c=c_j$  around $({\boldsymbol\omega}_{j}^\star,\mathbf s_{j}^\star)${, which satisfies the assumptions in Theorem~\ref{thm:con-sample} and Theorem~\ref{thm:con-feature}}. It is clear that $({\boldsymbol\omega}_{j}^\star,\mathbf s_{j}^\star)$ is an optimal solution of the approximate problem for $c=c_j$. Following the proof of~[37, {Theorem 1}], we can show the second statement.\end{IEEEproof}}

\begin{Lem}\label{lem:fmlim}
	%	Suppose that $f_{m}$ satisfies Assumption~\ref{asump:f} for all $m=0,\cdots,M$, $\bar{f}_{0}$ satisfies Assumption~\ref{asump:fbar}, $\bar{f}_{m}$ satisfies $\bar{f}_{m}(\boldsymbol\omega;\boldsymbol\omega,\mathbf{x})=f_{m}(\boldsymbol\omega;\mathbf{x})$ and Assumption~\ref{asump:fbar} for all $m=1,\cdots,M$, and
{Let $\{{\boldsymbol\omega}^{(t)}\}$ be the sequence generated by Algorithm~\ref{alg:con-sample} (Algorithm~\ref{alg:con-feature}).} Then, we have:
\begin{align} &\lim_{t\to\infty}\left\vert{\bar{F}_{m}^{(t)}({\boldsymbol\omega}^{(t)})- F_{m}({\boldsymbol\omega}^{(t)})}\right\vert=0,\ m=1,\cdots,M, \nonumber\\ &\lim_{t\to\infty}\left\Vert{\nabla\bar{F}_{m}^{(t)}({\boldsymbol\omega}^{(t)})-\nabla F_{m}({\boldsymbol\omega}^{(t)})}\right\Vert_2=0,\ m=0,\cdots,M,\nonumber\\
&\lim_{t\to\infty}\left\vert{\bar{F}_{m}^{(t)}(\boldsymbol\omega)- G_{m}(\boldsymbol\omega;{\boldsymbol\omega}^{(t)})}\right\vert=0,\  {\boldsymbol\omega\in\mathbb{R}^d},\ m=0,\cdots,M\nonumber
\end{align}
	almost surely, where $G_{m}(\boldsymbol\omega;{\boldsymbol\omega}^{(t)})\triangleq\frac{1}{N}\sum_{n\in\mathcal{N}} g_m(\boldsymbol\omega;{\boldsymbol\omega}^{(t)},\mathbf{x}_n)$.
\end{Lem}
\begin{IEEEproof}
	Lemma~\ref{lem:fmlim} is a consequence  of~\cite[Lemma 1]{Lemma}.
	We just need to verify that all the technical conditions therein are satisfied. Specifically, Condition (a) of \cite[Lemma 1]{Lemma} is satisfied because {$\{{\boldsymbol\omega}^{(t)}\}$ is assumed to be bounded.} Condition (b) of \cite[Lemma 1]{Lemma} comes from Assumption~\ref{asump:fbar}.4. Conditions (c)-(d) of \cite[Lemma 1]{Lemma} come from the stepsize rules in \eqref{eqn:rho} and \eqref{eqn:gamma}. Condition (e) of \cite[Lemma 1]{Lemma} comes from the Lipschitz property of $F(\boldsymbol\omega)$ from Assumption~\ref{asump:f}.2 and the stepsize rule in~\eqref{eqn:gamma}.
\end{IEEEproof}

%For notation simplicity, we omit the index $a=s,f$ in the rest of the proof without loss of generality.
\begin{Lem}\label{lem:fmconv}
	%	Suppose that $f_{m}$ satisfies Assumption~\ref{asump:f} for all $m=0,\cdots,M$, $\bar{f}_{0}$ satisfies Assumption~\ref{asump:fbar}, $\bar{f}_{m}$ satisfies $\bar{f}_{m}(\boldsymbol\omega;\boldsymbol\omega,\mathbf{x})=f_{m}(\boldsymbol\omega;\mathbf{x})$ and Assumption~\ref{asump:fbar} for all $m=1,\cdots,M$, and $\boldsymbol\Omega$ is compact.
	Consider a subsequence $\{{\boldsymbol\omega}^{(t_l)}\}_{l=1}^{\infty}$ generated by Algorithm~\ref{alg:con-sample} (Algorithm~\ref{alg:con-feature}) with $c=c_j$ converging to a limit point ${\boldsymbol\omega}^\star_j$. There exist uniformly continuous functions $\tilde F_m(\boldsymbol\omega)$, $m=0,\cdots,M$ such that
\begin{align}
\lim_{l\to\infty}\bar{F}_m^{(t_l)}(\boldsymbol\omega)=\tilde F_m(\boldsymbol\omega),
\ {\boldsymbol\omega\in\mathbb{R}^d},\ m=0,\cdots,M\label{eqn:app2.l1}
\end{align}
almost surely. Moreover, we have:
\begin{align}
&{\tilde F_m({\boldsymbol\omega}^\star_j)= F_m({\boldsymbol\omega}^\star_j)},\ m=1,\cdots,M,\label{eqn:app2.l2}\\	&{\nabla\tilde F_m({\boldsymbol\omega}^\star_j)=\nabla F_m({\boldsymbol\omega}^\star_j)},\ m=0,\cdots,M.\label{eqn:app2.l3}
\end{align}
\end{Lem}
\begin{IEEEproof}
	It readily follows from Assumption~\ref{asump:fbar} that the families of functions $\{\bar{F}_m^{(t_l)}(\boldsymbol\omega)\}$ are equicontinuous. Moreover, they are bounded and defined over a compact set. Hence, the Arzela–Ascoli theorem~[36] implies that, by restricting to a subsequence, there exists uniformly continuous functions $\tilde F_m(\boldsymbol\omega)$ such that~\eqref{eqn:app2.l1} is satisfied. Finally, \eqref{eqn:app2.l2} and \eqref{eqn:app2.l3} follow immediately from \eqref{eqn:app2.l1} and Lemma~\ref{lem:fmlim}.
\end{IEEEproof}

{By Lemma~\ref{lem:con-ap-ep}, it remains to show that a limit point of $\{({{\boldsymbol\omega}^{(t)}},\mathbf{s}^{(t)})\}$ generated by Algorithm~\ref{alg:con-sample} (Algorithm~\ref{alg:con-feature}) with $c=c_j$, $({\boldsymbol\omega}_{j}^\star,\mathbf s_{j}^\star)$, is a KKT point of Problem~\ref{Prob:con-sample-ep} (Problem~\ref{Prob:con-feature-ep}).} By Assumption~\ref{asump:f}, Assumption~\ref{asump:fbar}, and Lemma~\ref{lem:fmlim}, we can show $\lim_{t\to\infty}\Vert\bar{\boldsymbol\omega}^{(t)}-{\boldsymbol\omega}^{(t)}\Vert=0$. As the proof is similar to that in Appendix A, the details are omitted for conciseness.
Consider the subsequence $\{{\boldsymbol\omega}^{(t_l)}\}_{l=1}^{\infty}$ converging to ${\boldsymbol\omega}^\star_j$.
%\begin{align}
%	(\bar{\boldsymbol\omega}^{(t_l)},\mathbf{s}^{(t_l)})\triangleq&\mathop{\arg\min}_{\boldsymbol\omega.\mathbf{s}} \bar f^{(t_l)}_0(\boldsymbol\omega)+c\sum_{m=1}^M s_m \label{prob:tj}\\
%	\text{s.t.}\quad &\bar f^{(t_l)}_m(\boldsymbol\omega)\leq s_m,\quad m=1,2,\cdots,M.\nonumber
%\end{align}
%Letting $j\to\infty$ in \eqref{prob:tj}, using
By $\lim_{t\to\infty}\left\Vert\bar{\boldsymbol\omega}^{(t)}-{\boldsymbol\omega}^{(t)}\right\Vert_2=0$ and $\lim_{l\to\infty}{\boldsymbol\omega}^{(t_l)}={\boldsymbol\omega}^\star_j$, we have $\lim_{l\to\infty}\bar{\boldsymbol\omega}^{(t_l)}={\boldsymbol\omega}^\star_j$. Then, by $\lim_{l\to\infty}\bar{\boldsymbol\omega}^{(t_l)}={\boldsymbol\omega}^\star_j$, \eqref{eqn:app2.l1}, and Problem~\ref{Prob:con-sample-ep} (Problem~\ref{Prob:con-feature-ep}) with $c=c_j$, we have:
%and the Lipschitz continuity and strong convexity of $\bar{F}_m^{(t_l)}(\boldsymbol\omega)$, $\tilde F_m(\boldsymbol\omega)$, $\forall m$,
\begin{align} ({\boldsymbol\omega}^\star_j,\mathbf{s}^\star_j)\triangleq\mathop{\arg\min}_{\boldsymbol\omega.\mathbf{s}}\ &\tilde F_0(\boldsymbol\omega)+c_j\sum_{m=1}^M s_m \label{prob:star}\\
\text{s.t.}\ &\tilde F_m(\boldsymbol\omega)\leq s_m,\ m=1,2,\cdots,M.\nonumber
\end{align}
As $({\boldsymbol\omega}^\star_j,\mathbf{s}^\star_j)$ satisfies the KKT conditions of the problem in \eqref{prob:star}, and \eqref{eqn:app2.l2} and \eqref{eqn:app2.l3} in Lemma~\ref{lem:fmconv} hold, $\{({\boldsymbol\omega}^\star_j,\mathbf{s}^\star_j)\}$ also satisfies the KKT conditions of Problem~\ref{Prob:con-sample-ep} (Problem~\ref{Prob:con-feature-ep}) with $c=c_j$, {i.e., \eqref{eqn:KKT1}, implying that it is  a KKT point of Problem~\ref{Prob:con-sample-ep} (Problem~\ref{Prob:con-feature-ep}) with $c=c_j$. Therefore, we complete the proof.}

\section*{Appendix C: Proof of Lemma~\ref{lem:closedform}}
As the problem in~\eqref{prob:class-con-ap} is convex and the Slater’s condition holds, we can solve the problem in~\eqref{prob:class-con-ap} by solving its dual problem.
The Lagrangian function of the problem in~\eqref{prob:class-con-ap} is:
\begin{align} \mathcal{L}(\boldsymbol\omega,s,\nu,\mu)&\!\!=\!\!\left\Vert\boldsymbol\omega\right\Vert^2_2
+cs+\nu\!\left(\!\bar F^{(t)}(\boldsymbol\omega)\!+\!{C}_{a}^{t}\!-\!U\!-\!s\right)\!+\!\mu(-s)\nonumber\\ &\!\!=\!\!\left\Vert\boldsymbol\omega\right\Vert^2_2
+\nu\!\left(\bar F^{(t)}(\boldsymbol\omega)\!+\!{C}_{a}^{t}\!-\!U\right)
\!+\!(c-\nu-\mu)s,\nonumber
\end{align}
where $\nu$ and $\mu$ are the Lagrange multipliers.
Thus, the Lagrange dual function is given by:
\begin{align} &g(\nu,\mu)=\inf_{\boldsymbol\omega,s\geq0}\mathcal{L}(\boldsymbol\omega,s,\nu,\mu)\nonumber\\
	=&
\begin{cases}	\inf\limits_{\boldsymbol\omega}\left(\left\Vert\boldsymbol\omega\right\Vert^2_2+\nu\left(\bar F^{(t)}(\boldsymbol\omega)+{C}_{a}^{t}-U\right)\right),\ &c-\nu-\mu\geq0,\\
-\infty,\ &c-\nu-\mu<0.
\end{cases}\nonumber
\end{align}
As $\mathcal{L}(\boldsymbol\omega,s,\nu,\mu)$ is convex w.r.t. $\boldsymbol\omega$, by taking its derivative and setting it to zero, we can obtain the optimal solution:
\begin{align}\nonumber
	&\bar{\omega}_{a,0,l,j}^{(t)}=\frac{-\nu {A}_{a,l,j}^{(t)}}{2(1+\nu\tau)},\quad
\bar{\omega}_{a,1,j,p}^{(t)}=\frac{-\nu {B}_{a,j,p}^{(t)}}{2(1+\nu\tau)},
	%j\in\mathcal{J},\ k\in\mathcal{K},\nonumber\\
%\quad l\in\mathcal{L},\ j\in\mathcal{J},\nonumber
\end{align}
%Setting ${\omega}_{a,1,j,p}=\bar{\omega}_{a,1,j,p}^{(t)}$ and ${\omega}_{a,0,l,j}=\bar{\omega}_{a,0,l,j}^{(t)}$, $l\in\mathcal{L}$, $j\in\mathcal{J}$, $k\in\mathcal{K}$ in $\Vert\boldsymbol\omega\Vert^2_2+\nu\left(\bar F(\boldsymbol\omega)^{(t)}+A_{a}^{t}-U\right)$, we have:
and the optimal value $h(\nu)=\nu\left({C}_{a}^{(t)}-U-\frac{b\nu}{4(1+\tau\nu)}\right)$,
%\begin{align}
%	h(\nu)=\nu\left({C}_{a}^{(t)}-U-\frac{b\nu}{4(1+\tau\nu)}\right),\nonumber
%\end{align}
where $b$ is given in~\eqref{eqn:b}. Therefore, the dual problem of the problem in~\eqref{prob:class-con-ap} is given by:
\begin{align}
	\max_{\nu,\mu}\quad&h(\nu)\nonumber\\
	\text{s.t.}\quad &c-\nu-\mu\geq0,\ \nu\geq0,\ \mu\geq0, \nonumber
\end{align}
which is equivalent to the following problem:
\begin{align}
	\nu^*\triangleq\mathop{\arg\max}_{\nu}\quad&h(\nu)\label{prob:minist-con-dual}\\
	\text{s.t.}\quad &0\leq\nu\leq c.\nonumber
\end{align}
As $h(\nu)$ is convex in $\nu$, and $h'(\nu)=\frac{b-\left(b+4\tau(U-{C}_{a}^{(t)})\right)(1+\nu\tau)^2}{4\tau(1+\nu\tau)^2}$, by the optimality conditions of problem in~\eqref{prob:minist-con-dual}, we have:
\begin{align}
	\nu^*\!=\!
	\begin{cases} \!\left[\frac{1}{\tau}\left(\!\sqrt{\frac{b}{b+4\tau(U-{C}_{a}^{(t)})}}-1\!\right)\right]_0^c,\ &b+4\tau(U-{C}_{a}^{(t)})\!>\!0\\
		c,\ &b+4\tau({C}_{a}^{(t)})\!\leq\!0
	\end{cases},\nonumber
\end{align}
which completes the proof.

\section*{Appendix D: Proof of Theorem~\ref{thm:compare}}
	As $\boldsymbol{\omega}^*$ is a locally optimal solution of the problem in~\eqref{prob:class-uncon}, there exists $\varepsilon>0$ such that for all $\boldsymbol\omega$ with $\left\Vert\boldsymbol\omega-\boldsymbol{\omega}^*\right\Vert_2<\varepsilon$, we have:
	\begin{align}
		F(\boldsymbol{\omega}^*)+\lambda\left\Vert\boldsymbol{\omega}^*\right\Vert^2_2\leq
		F(\boldsymbol\omega)+\lambda\left\Vert\boldsymbol\omega\right\Vert^2_2.\label{eqn:lopt1}
	\end{align}
	Set $U=F(\boldsymbol{\omega}^*)$. Then, for all $\boldsymbol\omega$ with $\left\Vert\boldsymbol\omega-\boldsymbol{\omega}^*\right\Vert_2<\varepsilon$ and $F(\boldsymbol\omega)\leq U$, $\left\Vert\boldsymbol{\omega}^*\right\Vert^2_2\overset{(a)}\leq \frac{1}{\lambda}\left(F(\boldsymbol\omega)-F(\boldsymbol{\omega}^*)\right)+\left\Vert\boldsymbol\omega\right\Vert^2_2\overset{(b)}\leq
		\left\Vert\boldsymbol\omega\right\Vert^2_2$,
%	\begin{align}
%		\left\Vert\boldsymbol{\omega}^*\right\Vert^2_2\leq
%		\frac{1}{\lambda}\left(F(\boldsymbol\omega)-F(\boldsymbol{\omega}^*)\right)+\left\Vert\boldsymbol\omega\right\Vert^2_2\leq
%		\left\Vert\boldsymbol\omega\right\Vert^2_2,\nonumber
%	\end{align}
	where $(a)$ is due to~\eqref{eqn:lopt1} and $(b)$ is due to $F(\boldsymbol\omega)\leq U=F(\boldsymbol{\omega}^*)$.
	%	After simplification, we obtain $\Vert\boldsymbol{\omega}^*\Vert^2_2\leq\Vert\boldsymbol\omega\Vert^2_2$
	%	Therefore, there exists $U=F(\boldsymbol{\omega}^*)$ such that for all $\boldsymbol\omega$ with $\Vert\boldsymbol\omega-\boldsymbol{\omega}^*\Vert<\varepsilon$ and $F(\boldsymbol\omega)\leq U$, we have $\Vert\boldsymbol{\omega}^*\Vert^2_2\leq\Vert\boldsymbol\omega\Vert^2_2$, i.e.,
	Therefore, $\boldsymbol{\omega}^*$ is a locally optimal solution of the problem in~\eqref{prob:class-con}. The first statement holds.

	As $\boldsymbol{\omega}^{\dagger}$ is a locally optimal solution of the problem in~\eqref{prob:class-con}, the  necessary KKT condition $\nabla\left\Vert\boldsymbol{\omega}^{\dagger}\right\Vert^2_2+\xi\nabla F(\boldsymbol{\omega}^{\dagger})=0$ holds.
%	\begin{align}
%		\nabla\left\Vert\boldsymbol{\omega}^{\dagger}\right\Vert^2_2+\xi\nabla F(\boldsymbol{\omega}^{\dagger})=0.\nonumber
%	\end{align}
	Set $\lambda=\frac{1}{\xi}$. Then, we have $\nabla F(\boldsymbol{\omega}^{\dagger})+\lambda\nabla\left\Vert\boldsymbol{\omega}^{\dagger}\right\Vert^2_2=0$.
	Therefore, $\boldsymbol{\omega}^{\dagger}$ is a stationary point of the problem in~\eqref{prob:class-uncon}. If, in addtion, $\lambda$ and $\boldsymbol{\omega}^{\dagger}$ satisfy $\nabla^2 F(\boldsymbol{\omega}^{\dagger})+\lambda I\succeq 0$, i.e., the  Hessian Matrix is semi-definite, then $\boldsymbol{\omega}^{\dagger}$ is a locally optimal solution of the problem in~\eqref{prob:class-uncon}. The second statement holds.

\bibliographystyle{IEEEtran}
\bibliography{IEEEabrv,fedlearning}

\end{document}